\documentclass[journal]{IEEEtran}

\usepackage{cite}
\usepackage{amsmath,amssymb,amsfonts}
\usepackage{graphicx}
\usepackage{textcomp}
\usepackage{multirow}
\usepackage{comment}
\usepackage{enumitem, array} %Added
\usepackage{booktabs} %Added
\usepackage{adjustbox}%Added
\usepackage{caption}
\usepackage{subcaption}

\usepackage{placeins}%Added
\usepackage{tabularx}%Added
\usepackage{mathtools}%Added

\usepackage[english]{babel}%Added
\usepackage[utf8]{inputenc}%Added
\usepackage{algorithm}%Added
\usepackage[noend]{algpseudocode}%Added

\usepackage{amsthm} %Added
\theoremstyle{definition}
\algnewcommand{\algorithmicand}{\textbf{ and }} %Added
\algnewcommand{\algorithmicor}{\textbf{ or }} %Added
\algnewcommand{\OR}{\algorithmicor} %Added
\algnewcommand{\AND}{\algorithmicand} %Added

\usepackage{lipsum} %Added

\graphicspath{{Images/}{Images/Uncertainty}{Images/Classification}{Images/Anomalies}{Images/General}} %Added
\def\BibTeX{{\rm B\kern-.05em{\sc i\kern-.025em b}\kern-.08em
    T\kern-.1667em\lower.7ex\hbox{E}\kern-.125emX}}
\begin{document}
% \history{Date of publication xxxx 00, 0000, date of current version xxxx 00, 0000.}
% \doi{10.1109/ACCESS.2017.  DOI}

\title{Anomaly Detection using 
Ensemble Classification and Evidence Theory
	%Application
%	 and Expert Knowledge
 }
 
\author{\IEEEauthorblockN{Fernando Ar\'{e}valo\IEEEauthorrefmark{1},
		 Tahasanul Ibrahim\IEEEauthorrefmark{4},
   Christian Alison M. Piolo, %\IEEEauthorrefmark{4},
		Andreas Schwung\IEEEauthorrefmark{4}\\
	} 
	\IEEEauthorblockA{\IEEEauthorrefmark{1}Faculty of Electrical Engineering and Information Technology\\
		Ruhr-Universit\"{a}t Bochum, Bochum 44801, Germany\\
		Email: Fernando.ArevaloNavas@ruhr-uni-bochum.de\\}
	\IEEEauthorblockA{\IEEEauthorrefmark{4}Department of Automation Technology\\
		South Westphalia University of Applied Sciences, Campus Soest, Germany\\
		%Email: \{arevalo.fernando, schwung.andreas\}@fh-swf.de
        }
 
}

% \tfootnote{This work was supported by the Open Access Publication Fund of South Westphalia University of Applied Sciences.}

%\tfootnote{This paragraph of the first footnote will contain support 
%information, including sponsor and financial support acknowledgment. For 
%example, ``This work was supported in part by the U.S. Department of 
%Commerce under Grant BS123456.''}

% \markboth
% {Ar\'{e}valo N. \headeretal: Anomaly Detection using Ensemble Classification and Evidence Theory}
% {Ar\'{e}valo N. \headeretal: Anomaly Detection using Ensemble Classification and Evidence Theory}

% \corresp{Corresponding author: Fernando Ar\'{e}valo (e-mail: Fernando.ArevaloNavas@ruhr-uni-bochum.de).}

\maketitle

\begin{abstract}
Multi-class ensemble classification remains a popular focus of investigation within the research community. The popularization of cloud services has sped up their adoption due to the ease of deploying large-scale machine-learning models. It has also drawn the attention of the industrial sector because of its ability to identify common problems in production. However, there are challenges to conform an ensemble classifier, namely a proper selection and effective training of the pool of classifiers, the definition of a proper architecture for multi-class classification, and uncertainty quantification of the ensemble classifier. 
The robustness and effectiveness of the ensemble classifier lie in the selection of the pool of classifiers, as well as in the learning process. Hence, the selection and the training procedure of the pool of classifiers play a crucial role. An (ensemble) classifier learns to detect the classes that were used during the supervised training. However, when injecting data with unknown conditions, the trained classifier will intend to predict the classes learned during the training. To this end, the uncertainty of the individual and ensemble classifier could be used to assess the learning capability.
We present a novel approach for novel detection using ensemble classification and evidence theory. A pool selection strategy is presented to build a solid ensemble classifier. We present an architecture for multi-class ensemble classification and an approach to quantify the uncertainty of the individual classifiers and the ensemble classifier. We use uncertainty for the anomaly detection approach.
Finally, we use the benchmark Tennessee Eastman to perform experiments to test the ensemble classifier's prediction and anomaly detection capabilities.
\end{abstract}

% \begin{keywords}
% \end{keywords}

% \titlepgskip=-15pt
% \maketitle

\section{Introduction}\label{section__intro}
%EC
%\PARstart{E}{nsemble}
Ensemble classification (EC) has become a popular subject of applied research in several branches of the industry, such as automotive, pharmaceutical, energy, and insurance. The EC power relies on the ability to discover patterns in data that can throw light on the optimization of the business or the discovery of anomalies in the process.
Ensemble learning in supervised classification is a common practice because it permits combining different models to achieve better performance. Bagging and Boosting are two commonly used techniques that have proved to be a good fit for fault diagnosis %\cite{Yao2016} \cite{Shah2016}. 
\cite{Wang2017} \cite{Zhang2018}.
%EC 
An alternative way to combine classifiers is through the use of information fusion. To this end, there are different strategies to achieve information fusion \cite{Khaleghi2013}, the most common ones involve the use of Fuzzy Logic \cite{Zhang2018}, Bayesian \cite{Lee2014}, %rule-based systems[CIT], 
and evidence theory (ET) \cite{LiuPan2018}. We focus on the use of ET to build the ensemble classification.  % \cite{Zeng.2017}. 
However, there are essential aspects to consider while doing ensemble classification, such as an architecture for the information fusion, the selection of (the pool of) classifiers, a solid training procedure that guarantees performing classifiers, and uncertainty quantification of the classifiers that signalizes the learning capability.
The pool selection plays a vital role in the overall performance of the EC. %\textcolor{red}{[CIT]}. 
There are different factors to consider while making the pool selection; namely, the diversity or heterogeneity of the sources \cite{ZhangLi2020}, the inclusion of expert classifiers (e.g., that can detect specific classes)\cite{JiaoGuo2022}, the data nature, and the classifier mathematical principle.
%As it was stated before, 
The use of heteregeneous sources improves the results in information fusion, in this specific case the use of diverse classifiers \cite{LahatAdali2015}.
% One prerequisite for information fusion is the use of heterogeneous sources of information (e.g., the classifiers) \cite{LahatAdali2015}.% \textcolor{red}{[CIT]}. 
The challenge here is to define an effective strategy to measure the heterogeneity or the diversity between the classifiers' combinations. An additional point to note is the expertness of a classifier, which means a classifier that can detect particularly well a specific class. Data nature and the mathematical principle are prior considerations during the selection of the pool, which in most cases involves expert knowledge.
A performing EC requires not only an effective pool selection strategy but also a measure on the certainty of the predictions, in other words, how reliable the predictions can be. A high uncertainty would signal a poor training process and a poor-performing classifier (e.g., a poor generalization). Thus, uncertainty quantification (UQ) can be used to measure the epistemic uncertainty (e.g., associated with lack of knowledge in the classifier) \cite{Senge2014}. Moreover, the UQ can be used to weigh the classifier's performance (after training), which can be useful during information fusion.
% A performing EC requires not only an effective pool selection strategy but also a solid training process that enhances generalization. To this end, uncertainty quantification (UQ) can enlighten this activity while training the pool of classifiers, because it can display the uncertainty behavior. A high uncertainty would signalize a poor training process, and subsequently a poor-performing classifier (e.g., a poor generalization). %For instance, in the case of multi-class EC, the UQ shows the associated uncertainty of a certain class, which can be considered while performing model inference. This last feature could be also exploited while performing inference on the EC.
% NN-based optimization
%\textcolor{red}{ As mentioned before, an effective training procedure increases the accuracy of the classifier, as well as the generalization capability. A strategy for the optimization of the learning process, specifically a fast convergence of the gradient (e.g., in the case of NN-based models), could play a key role to improve the ensemble performance.}
%ECET

The definition of a pool selection strategy and uncertainty quantification pave the way for a performing EC. The architecture of the EC plays a vital role in the combination of classifiers. Important considerations include the number of classes (e.g., binary or multi-class), ensemble size (e.g., the number of classifiers that form the EC), the selected classifiers (e.g., only NN-based classifiers), and the transformation of the predictions into a common framework that allows the fusion \cite{ArevaloRementeria2018}. 
%anomaly detection
While training the EC, a data subset is selected that is expected to represent the overall data. Training an EC with a specific portion of the input space %(e.g., a particular dataset from the overall data) 
assures the learning procedure on that specific portion \cite{ZhangLi2020}. %, and given the training procedure is desired to have a certain generalization degree on the EC. 
%training an ensemble classifier with a specific set of conditions transfers the classifier 
The drawback is limiting the EC with a static behavior, which means the classifier will perform predictions only based on the learned classes (e.g., in-distribution data). The challenge here is doting the ensemble classifier with an anomaly detection capability when feeding unknown conditions (e.g., out-of-distribution data).

\begin{table}[!htt]
	\centering
	\caption{List of symbols and abbreviations.}
	\begin{tabular}{l l}
		\toprule
		\textbf{Symbol} & \textbf{Description} \\ 		
		\midrule
		$DSET$          & Dempster Shafer evidence theory\\
		$AD$            & Anomaly detection\\
    	$EC$            & Ensemble classifier\\
		$w$             & Confidence weight\\
      	$p$             & Prediction\\
		$m$             & Mass function  \\
		$U$             & Uncertainty  \\
		$D^{Tr}$        & Training data\\
		$D^{Va}$        & Validation data\\
		$D^{Te}$        & Testing data\\
		$D^{B}$         & Training mini-batch data\\
		$UQ^{Va}$       & Uncertainty quantification during validation\\
		$UQ^{Te}$       & Uncertainty quantification during testing\\
		$E^{SM}$        & Expert criteria using a softmax-based approach\\
		$DV^{A2}$       & Diversity criteria alternative 2\\
		$U_{E}$         & Ensemble uncertainty\\
		$k$             & Sensitivity to zero factor\\
		$F_{D}$        & Fusion using DSET rule of combination\\
            $F_{Y}$        & Fusion using Yager rule of combination\\
		\bottomrule
	\end{tabular}
	\label{table__list_symbols} 
\end{table}

We propose a novel approach for anomaly detection using ensemble classification and evidence theory (ECET) that provides the guidelines to create performing ensemble classifiers (EC). Besides, we present an uncertainty quantification methodology to measure the uncertainty of the trained classifiers and the uncertainty of the EC during inference.  
Finally, we present an architecture for binary and multiclass EC using evidence theory that not only provides a robust classification performance but also tracks the EC uncertainty to detect anomalies while feeding unknown data (out-of-distribution data). 

The contributions of this paper are:
\begin{itemize}
\item A methodology for ensemble classification using evidence theory (ECET) that applies a fusion on the decision level for the pool of classifiers. The methodology is used for creating binary and multiclass ECs using different ensemble sizes and classifiers.
\item A strategy for pool selection that considers the criteria diversity, expert, and pre-cut. This pool selection reduces the number of possible combinations while providing a performing EC.
\item An uncertainty quantification methodology that measures the uncertainty after the training process of the pool of classifiers. Besides, it assesses the uncertainty of the EC predictions during model inference.
\item An approach for anomaly detection (AD) using ensemble classification and uncertainty quantification. The AD detects anomalies while feeding unknown data (out-of-distribution data) through an uncertainty tracking.
\end{itemize}

This paper is structured as follows: a literature review is presented in section \ref{section__related}. %A theoretical background is provided in section \ref{section:theory}. 
The ensemble classifier's architecture and methodology are presented in Section \ref{section__anomaly}. Section \ref{section__usecase} presents a use case of the proposed methodology applying the benchmark dataset Tennessee Eastman. Finally, the conclusion and future work are summarized in Section \ref{section__conclusions}.

\section{Related Work}\label{section__related}
\begin{figure}[!htt]
	\centering
		\includegraphics[width=0.45\textwidth,keepaspectratio]{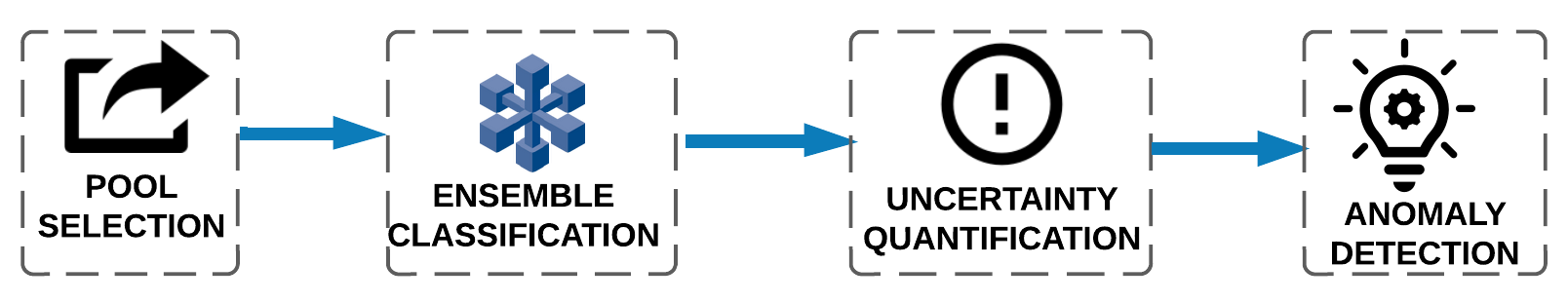}
	\caption{Main topics of ECET.}\label{fig__ECET__related}
\end{figure}

This section presents the state of the art of main topics addressed in this paper: ensemble classification, pool selection, uncertainty quantification, and anomaly detection. 
Fig. \ref{fig__ECET__related} shows the relationships between the topics of this section.
\subsection{Ensemble Classification}
Ensemble classification (EC) is a versatile approach commonly used in the research community because it provides a more robust output and benefits from the heterogeneity of its sources (e.g., classifiers) \cite{ZhangLi2020}. Defining a reliable EC architecture requires special considerations in terms of data preparation (e.g., classifiers that require standardized data), ensemble size (e.g., the number of classifiers), classifiers type (e.g., shallow classifiers, deep learning models, or hybrid) and the strategy to combine the classifiers. %\textcolor{red}{[CIT]}.
The data preparation is a usual step in a machine learning pipeline, which is an explicit requirement for some classifiers (e.g., using a standard scaler for NN-based classifiers). The EC ensemble size (or the number of classifiers involved) is usually a trial and error parameter, which means that a fixed parameter cannot be used as a general parameter for all the cases to be learned (e.g., a large number of classifiers in the EC does not guarantee a better outcome). 
In contrast, selecting a combination strategy presents multiple options for EC \cite{Khaleghi2013}. The popular strategies include the use of bagging (e.g., random forest) and boosting (e.g., AdaBoost) \cite{Wang2017} \cite{Zhang2018}, Bayesian \cite{CaoYu2020}, fuzzy \cite{Scott2017}, average \cite{ZhangLi2020}, majority voting \cite{Chen2019} \cite{GuoHe2019}, and evidence theory (ET) \cite{LiuPan2018} \cite{Sun2017}.  
ET is a preferred approach in literature because it not only provides a framework to combine different information sources (e.g., predictions of classifiers) in the form of sets of evidence but also considers the uncertainty of the information sources \cite{HuoMartinez2022}. This last feature allows allocating confidence in the ET predictions. Examples of EC using ET can be found in \cite{HuoMartinez2022} \cite{SomeroSnidaro2022} \cite{ZhouLiu2022}.  %\cite{DebaqueFlorea2019}. 
Although we stated the benefits of EC, there are essential considerations to be addressed while forming an EC, namely the pool selection, the transformation of the classifier output into a set of evidence, and the architecture, among others. 
%\cite{ZhangJiang2020}. 
% Besides, the ET is used to quantify the (epistemic) uncertainty \cite{HuoMartinez2022}. 

\subsection{Pool Selection}
The pool selection plays an essential role in EC because its strength lies in combining heterogeneous and performing classifiers. For this purpose, different criteria are proposed in literature using performance \cite{GuoHe2019}, expert area or competence sub-region \cite{JiaoGuo2022}, and diversity measurement \cite{ZhangLi2020} \cite{JanVerma2019}. A common practice for selecting the pool of (base) classifiers is using performance as a sorting criterion after the training phase because of its simplicity and effectiveness. However, the performance relies on a static value corresponding to specific data conditions (e.g., a local portion of the input space), which is an issue in case of a concept drift \cite{JiaoGuo2022}.  
Specifically, it occurs when the training data is only a local representation of the data, which incurs an underperformance when presenting data from a different part of the input space. To this end, Jiao et al. \cite{JiaoGuo2022} proposes a dynamic ensemble selection to address the concept drift by training and selecting classifiers per competence sub-regions dynamically. 
On the other hand, it is important to note that considering a $n$ number of base classifiers produces a notable amount of possible combinations and, thus, leads to the dilemma of which combination to choose. The question here is how to reduce the number of possible combinations without compromising the EC's performance and generalization.    
A requisite to form an EC relies on the heterogeneity of the information sources, or put this in different terms, in how diverse they are. The diversity between classifiers improves the generalization capability of the EC \cite{ZhangLi2020}.   
To this end, diversity measurements are proposed in literature to tackle this problem. 
Jan et al. \cite{JanVerma2019} propose a pairwise diversity measure for an incremental classifier selection to identify which classifiers improve the learning capability. The diversity measure compares two ECs using matrices with misclassified samples and uses a (customized) indicator function that quantifies the diversity. The approach discards classifiers from an EC that neither improves the accuracy nor the diversity.

\subsection{Uncertainty Quantification}
The support of a classifier relies not only on its capability to provide a condition prediction while feeding data but also on how reliable this prediction is. For this purpose, uncertainty quantification (UQ) assesses the prediction reliability (e.g., a 90\% likeliness of a prediction not only provides the associated class but also how certain the prediction can be). The uncertainty is divided into two categories: aleatoric (e.g., associated with random effects) and epistemic (e.g., or model uncertainty associated with lack of knowledge) \cite{Senge2014}. We focus on the quantification of epistemic uncertainty.
The epistemic uncertainty is quantified using different strategies, namely using entropy \cite{LiuZheng2022}, variance \cite{DongYang2021} \cite{Wang2020}, Bayesian neural networks (BNN) \cite{RavindranSantora2022}, Monte Carlo dropout \cite{PyleHughes2022}, and evidence theory \cite{HuoMartinez2022} \cite{DongYang2021}. An extensive survey of the different methods for UQ is found in \cite{Abdar2021} \cite{OsmanShirmohammadi2021}.
Using ensemble systems is a common approach to quantifying epistemic uncertainty. %\cite{Dong2021}, \cite{Wang2020}
Dong et al. \cite{DongYang2021} present a multi-expert uncertainty-aware learning (MUL) approach that compares the variance of multiple parallel dropout networks. The uncertainty is quantified using the difference between the variance of the different networks (experts). Thus, a significant difference in the variance of the experts is an indicator of high uncertainty of the prediction. Wang et al. \cite{Wang2020} uses an uncertainty-driven deep multiple instance learning to optimize the learning process by discarding noisy training samples from the positive bags. To this end, the authors calculate the mean and the standard deviation of the probabilistic ensemble prediction of a bag to determine the uncertainty. Likewise, Dong et al. stated that a larger standard deviation corresponds to higher uncertainty. 
Huo et al. \cite{HuoMartinez2022} monitors the uncertainty at the decision level using an ET-based EC.
In contrast to \cite{Wang2020} which uses the UQ during the training phase, \cite{HuoMartinez2022} and \cite{DongYang2021} use the UQ during model inference. 
A framework that monitors the uncertainty in the complete EC lifecycle, namely from the training phase until the inference, could not only improve the EC performance but also add the feature of anomaly detection while performing inference.

\subsection{Anomaly Detection}% using Ensemble Classification}
Anomaly detection has been addressed using different approaches namely multivariate statistical methods (e.g., Hoteling, Mahalanobis distance) \cite{SunS2021} \cite{NguyenSood2022}, variational autoencoders (VAE) \cite{HeJin2021} \cite{GuoJi2020}, Bayesian neural networks (BNN) \cite{WuJin2021} \cite{WangJere2022}, ensemble learning \cite{HuoMartinez2022} \cite{YangWang2019}, among others. 
An extensive survey of the different methods for anomaly detection is found in \cite{VelasquezPerez2022}.
%A EC using ET  Specifically, the EC could detect anomalies using a strategy based on the UQ.
The use of the Mahalanobis distance and Chi-square distribution is commonly preceded by a feature reduction on the data using principal component analysis (PCA) \cite{SunS2021} \cite{SunWang2019} and canonical variate analysis (CVA) \cite{CaoYu2020}.
Sun et al. \cite{SunS2021} propose using a confidence region to diagnose heart diseases using a Gaussian mixture model and the Mahalanobis distance. 
Nguyen et al. \cite{NguyenSood2022} use the Mahalanobis distance and Chi-square distribution for IoT node authentication by setting a cut-off value in the Chi-square distribution to identify non-legitimate nodes. Cao et al. \cite{CaoYu2020} propose using clustering, a sub-region CVA, and Hoteling to identify faults.
He et al. \cite{HeJin2021} proposes a DL-based framework for diagnosis and fault detection in bearings. The authors implemented a VAE, in which the latent space of the encoder is used together with a novelty threshold (based on the Bhattacharyya distance) to identify unseen faults.  
The anomaly detection is related to the out-of-distribution data (OOD), in which the test data distribution presents dissimilarities to the training data (in-distribution data) \cite{PyleHughes2022} \cite{UgrenovicVankeirsbilck2020}. An (ensemble) classifier should be aware of anomalies, which implies a strategy that considers monitoring the prediction confidence. The prediction confidence is addressed by the approaches, namely, BNN and EC using ET.
Wu al. \cite{WuJin2021} proposes a method using dropout-based Bayesian deep learning to detect unexpected faults of high-speed train bogie. The approach outputs a diagnosis result and an uncertainty indicator of the detected class. 
Ensemble learning using ET is a popular approach because it not only provides a robust classification but also can detect anomalies by monitoring the uncertainty \cite{HuoMartinez2022}. The last feature addresses the issue of EC that cannot detect unknown classes (as a result of having classifiers trained with a fixed dataset \cite{ZhangLi2020}).

%Closing paragraph
Our approach differentiates from the current contributions in literature, in which we present a holistic methodology for ensemble classification using evidence theory (ECET) that not only provides a robust classification output but also assesses the likeliness of the output by quantifying the uncertainty. First, we present an uncertainty quantification (UQ) approach using evidence theory that measures not only the uncertainty after the training phase of the (individual) classifiers but also the EC uncertainty during inference. Second, we propose a pool selection strategy that allows the creation of an EC by choosing the pool of (base) classifiers using key criteria such as performance, diversity, and expert area. Third, we propose an EC architecture using evidence theory that provides binary and multiclass ECs with a robust prediction (with respect to the base classifiers) with a likeliness degree through the use of UQ.
Fourth, we propose a novel anomaly detection procedure that performs uncertainty tracking on the EC while performing inference on unknown data (out-of-distribution data).
We demonstrated the approach's robustness by applying the Tennessee Eastman benchmark.    

\section{ECET: Anomaly Detection using Ensemble Classification and Evidence Theory}\label{section__anomaly}

This research proposes a data-based model approach to address anomaly detection using Ensemble Classification and Evidence Theory (ECET). The main topics considered in this section are: a theoretical background, uncertainty quantification, %an optimized training of ANN classifiers using adaptive learning rate, ANN training using early stopping, 
criteria selection of classifiers, the ensemble classifier using evidence theory (ECET), and anomaly detection using ECET. 

\begin{figure*}[!htt]
	\centering
		\includegraphics[width=0.6\textwidth,keepaspectratio]{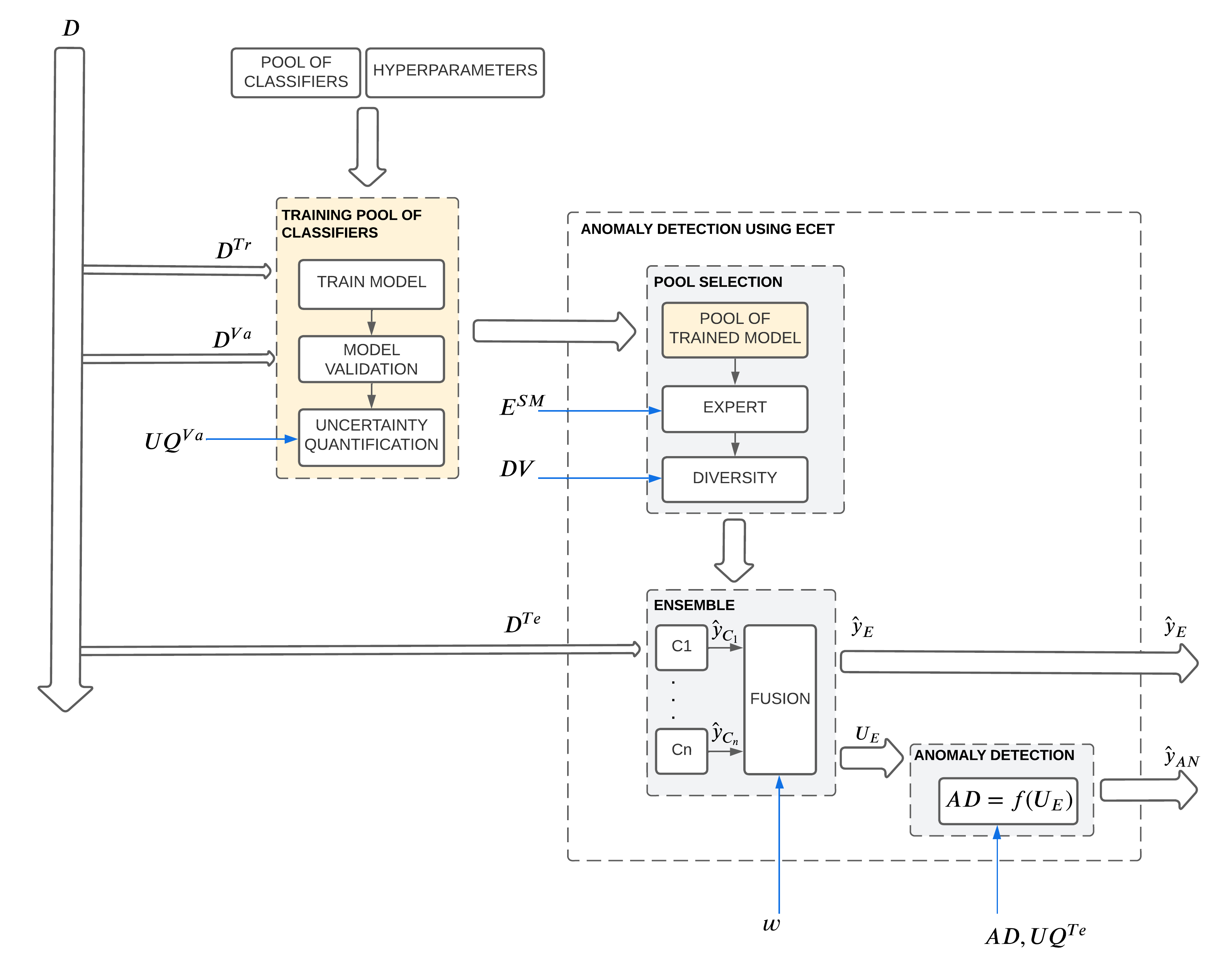}
	\caption{Overview of anomaly detection using ECET.}\label{fig__ecet__overview}
\end{figure*}

This approach consists of two main blocks: the training process and the inference model for classification and anomaly detection (see Fig. \ref{fig__ecet__overview}). 

The overall system is detailed as follows:

\begin{itemize}
    \item The prior step is the \textbf{training of the pool of classifiers}, and comprises three main activities namely the \textit{model training}, \textit{model validation} and \textit{uncertainty quantification}. The main goal of this block is to provide a pool of trained models. The inputs to this block are the training data $D^{Tr}$ and validation data $D^{Va}$, a pool of classifiers, and a list of hyperparameters for the classifiers. The main contributions of this block, represented with blue arrows, are:
    \begin{itemize}
        \item the overall uncertainty quantification $UQ^{Va}$ of each model per trained class.
    \end{itemize}
    \item The second step is the \textbf{anomaly detection using ECET}. This step is the model deployment, and it consists of three main blocks, namely the \textit{pool selection}, the \textit{ensemble} (classifier), and the \textit{anomaly detection}. The main goals of this block are the classification of known conditions and the detection of anomalies in the data. The inputs to this block are the pool of trained models and the testing data $D^{Te}$. The main contributions of this block are:
    \begin{itemize}
        \item a methodology for the ensemble classifier using evidence theory (ECET).
        \item a strategy for anomaly detection using uncertainty quantification $UQ^{Te}$.
    \end{itemize}
\end{itemize}

\subsection{Theoretical Background}
This section presents the theoretical basics for the development of the sections \ref{approach__uncert_quantif}, %\ref{approach__optimized_training}, \ref{approach__EC}, \ref{approach__AD}.
In the first place, the basics of evidence theory are presented. Secondly, the evidential treatment of supervised classification presents the application of the ET for uncertainty quantification, ensemble classification, and anomaly detection. %At last, the theory behind the gradient is described.

\subsubsection{Evidence Theory}
The evidence theory (ET) serves as a framework to model epistemic uncertainty. Important to mention is that the ET allows to combine different information sources. The effectiveness of the combination or fusion relies upon using heterogeneous sources of information. 
Formally, a frame of discernment $\Theta$ is defined as \cite{Shafer1976}: $\Theta = \{A,B\}$, where A and B are focal elements. The power set is $2^{\theta}$ represented as: $2^{\theta} = \{ \phi, \{A\},\{B\}, \Theta \} \}$. A mass function is defined using: m: $2^{\theta} \rightarrow [0,1]$. The mass function fulfills the conditions: $m(\phi) = 0$, and $\sum_{A \subseteq \Theta} m(A) = 1$. The focal elements are mutually exclusive, thus: $A \cap B = \phi$.

The \textit{Dempster-Shafer Rule of Combination} (DSRC) allows to combine two sources of information, specifically two mass functions, using the equation:
\begin{equation} \label{dempster__1}
\begin{split}
	m_{DS}(A) & = m_{1}(B) \bigotimes m_{2}(C) \\
	& = \frac{\sum_{B \cap C \neq \phi} m_{1}(B)\ \times\ m_{2}(C)}{1 - \sum_{B \cap C =\phi} m_{1}(B)\ \times\ m_{2}(C)} 
\end{split}
\end{equation}
where $m_{1}(B)$ and $m_{2}(C)$ are the mass functions of each information source, $m_{DS}(A)$ is the resulting mass function after the DSET fusion. 

The amount of conflicting evidence $b_{k}$ is represented by:
\begin{equation} \label{dempster__2}
\begin{split}
b_{k} = \sum_{B \cap C=\phi} m_{1}(B) \times m_{2}(C)
\end{split}
\end{equation}
Important to note is that the uncertainty represented by the term $b_{k}$ is split by each combined focal element of $m_{DS}(A)$.

The \textit{Yager rule of combination} (YRC), likewise the DSRC, allows to combine two mass functions using \cite{Yager1987}:
\begin{equation} \label{yager__1}
\begin{split}
m_{Y}(A) = \sum_{B \cap C \neq \phi}^{}m_{1}(B)\times m_{2}(C)
\end{split}
\end{equation}
where $m_{Y}(A)$ is the resulting mass function after the Yager fusion.
The mass function of the focal element $\Theta$ is represented by:
\begin{equation} \label{yager__2}
\begin{split}
m_{Y}(\theta) = q(\theta) + q(\phi)
\end{split}
\end{equation}
where $q(\theta)$ represents the evidence of the focal element $\Theta$, and $q(\phi)$ is the conflicting evidence.
Thus, $q(\phi)$ is defined as:
\begin{equation} \label{yager__3}
\begin{split}
q(\phi) = \sum_{B\bigcap C=\phi}^{}m_{1}(B)\times m_{2}(C)
\end{split}
\end{equation}
In contrast to DSRC, YRC only performs a fusion over the known elements, the intersection of elements that results in $\phi$ are considered separately in the term $q(\phi)$. The latest means that the conflicting evidence $q(\phi)$ is assigned solely to the focal element $\Theta$ of the combined mass function $m_{Y}(\theta)$.

The following equation is used while performing a fusion of more than two information sources: 
\begin{equation} \label{dempster__3}
\begin{split}
m(A) = \bigg( \Big( m_{1}(B) \bigotimes m_{2}(C) \Big) ... \bigotimes m_{n}(Z) \bigg)
\end{split}
\end{equation}
where $m_{n}(Z)$ is the mass function of the $n_{th}$ information source, and $n \in \mathbf{N}$. 

\subsubsection{Evidential Treatment of Supervised Classification}\label{approach__evidential_treatment}

The prediction of a classifier is represented as either a unique label or an array with the probabilities for all the possible labels. Thus, the classifier prediction is: 
    \begin{equation} \label{pred2evid__1}
	\begin{split}
	p = L_{1}
	\end{split}
	\end{equation}    
    \begin{equation} \label{pred2evid__2}
	\begin{split}
	p = [L_{1}, L_{2},..., L_{n} ]
	\end{split}
	\end{equation}    

In the case of a unique label, a transformation is required to consider all the possible labels in the mass function. $\Theta$ includes all the possible cases: 

    \begin{equation} \label{pred2evid__3}
	\begin{split}
	\Theta = \{L_{1}, L_{2},..., L_{n} \}
	\end{split}
	\end{equation}  

The elements of the power set for two labels are represented as:

     \begin{equation} \label{pred2evid__4}
	\begin{split}
	2^{\Theta} = \{\phi, \{L_{1}\}, \{L_{2}\}, \{L_{1},L_{2}\} \}
	\end{split}
	\end{equation}     

The last term corresponds to the uncertainty since it corresponds to both cases. This term will be referred to as $U$. 
Thus, the mass function is represented as:
 \begin{equation} \label{pred2evid__5}
  	\begin{split}
  	     m_{p} = [L_{1} \quad L_{2} \quad
  		... \quad 
  		L_{n}
  		\quad U] 
  	\end{split}
  \end{equation}  
where,
    \begin{equation} \label{pred2evid__6}
	\begin{split}
	U = \{L_{1},L_{2},..., L_{n}\}
	\end{split}
	\end{equation}  
	
Since only one label will be active at a time, a strategy is necessary to fill up the mass function. This strategy considers which label is active by assigning a nearly one and a nearly zero for the rest of the inactive labels:
      \begin{equation} \label{pred2evid__7}
  	\begin{split}
  	    m_{p} = [C_{A}
  		\quad C_{R_{1}} \quad C_{R_{2}} \quad
  		... \quad 
  		C_{R_{n-1}}
  		\quad U]
  	\end{split}
  \end{equation}    
  
Cheng et al. \cite{Cheng1988} applied a sensitivity to zero approach, which approximates the zero and one value to a nearly-zero value and nearly-one value, respectively. This approach enhances information fusion since all the evidence will be considered, even if these values are small. This approach is relevant when using DSET and its orthogonal multiplication and was applied to a set of evidence in \cite{ArevaloNguyen2017}:
  \begin{equation} \label{pred2evid__8}
  	\begin{split}
  		k = 1-10^{-F}
  	\end{split}
  \end{equation}
In \cite{ArevaloRementeria2018}, a methodology was presented to transform a prediction into a set of evidence:
  \begin{equation} \label{pred2evid__9}
  	\begin{split}
  		C_{A} = k*w_{C_{A}}
  	\end{split}
  \end{equation}
   
  \begin{equation} \label{pred2evid__10}
  	\begin{split}
  		C_{R_{m}} = \frac{1-k}{n-1}*w_{C_{R_{m}}}
  	\end{split}
  \end{equation}
   
  \begin{equation} \label{pred2evid__11}
  	\begin{split}
  		U = 1- C_{A}-\sum_{m=1}^{n-1} C_{R_{m}}\\
  	\end{split}
  \end{equation}
   
 Having the prediction as a row vector $\mathbf{p}$ (e.g., a common scenario for NN-based classifiers with a softmax layer), where $\mathbf{p} \in \{L_{1}, L_{2}, L_{3} ... L_{n} \}$, and $n \in \mathbb{N}$. Thus, a prediction $\mathbf{p}$ can be represented as a row vector of size $1 \times n$:
    \begin{equation} \label{pred2evid__12}
  	\begin{split}
  		\mathbf{p} = [p_{L_{1}},p_{L_{2}},...,p_{L_{N}}]\\
  	\end{split}
  \end{equation}
 
 The prediction row vector $\mathbf{p}$ has an associated confidence weight row vector $\mathbf{w_{p}}$ with the size $1xn$: 
    \begin{equation} \label{pred2evid__13}
  	\begin{split}
  		 \mathbf{w_{p}} = [w_{p_{L_{1}}},w_{p_{L_{2}}},...,w_{p_{L_{N}}}]\\
  	\end{split}
  \end{equation}                 
 
 The row vector prediction $\mathbf{p}$ is transformed into a row vector evidence $\mathbf{e}$ of size $1 \times n+1$ as follows:
    \begin{equation} \label{pred2evid__14}
  	\begin{split}
  		m[i,j] = (\mathbf{p} \quad o \quad \mathbf{w_{p}})_{ij}\\
  	\end{split}
  \end{equation}
 
where $i = 1$ and $j \neq n+1 $, this operation is also denoted as the Hadamard product of the row vectors $\mathbf{p}$ and $\mathbf{w_{p}}$. The last term of the row vector $\mathbf{e}$ defines the uncertainty of the prediction $\mathbf{p}$:
    \begin{equation} \label{pred2evid__15}
  	\begin{split}
  	    m[i,j] = 1 - \mathbf{p} \cdot \mathbf{w_{p}}\\
  	\end{split}
  \end{equation}
 
 where $i = 1$ and $j = n+1$, and $\mathbf{p} \cdot \mathbf{w_{p}}$ is the dot product between the row vectors $\mathbf{p}$ and $\mathbf{w_{p}}$. The uncertainty can also be represented as:
  \begin{equation} \label{pred2evid__16}
  	\begin{split}
  		U = 1- \sum_{i=1}^{N} p_{i} \cdot w_{p_{i}}\\
  	\end{split}
  \end{equation}

Thus, the evidence $\mathbf{e}$ can be represented as the row vector:
      \begin{equation} \label{pred2evid__17}
      \begin{split}
    	m_{p} = [p_{1}\cdot w_{p_{1}} \quad    p_{2}\cdot w_{p_{2}} \quad
  		... \quad 
  		p_{n}\cdot w_{p_{n}}
  		\quad U]      
      \end{split}
  	  \end{equation}

\subsection{Ensemble Classifier and Evidence Theory}\label{approach__EC}
\begin{figure*}[!htt]
	\centering
		\includegraphics[width=0.99\textwidth,keepaspectratio]{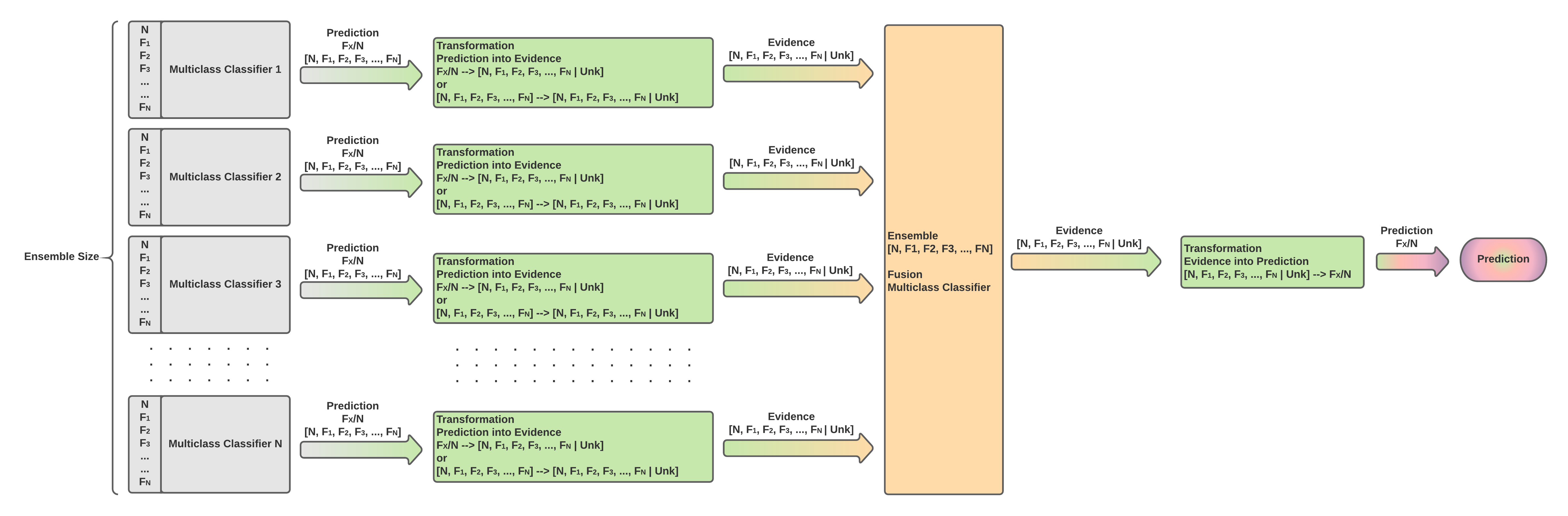}
	\caption{Architecture for EC. 
	%\textcolor{red}{CHANGE: remove N, and change subindex}
	}\label{fig__ECET_Multiclass__multi}
\end{figure*}

The ECET provides the guidelines for the creation of performing binary and multiclass ensemble classifier (EC). A preliminary step involves the selection of the pool of classifiers (detailed in the following subsection), the transformation of the classifiers' predictions into sets of evidence (detailed in section \ref{approach__evidential_treatment}), and the definition of an EC architecture.
The robustness of the EC lies in the use of (heterogeneous) multiple. In this context, heterogeneity is understood as the diversity between classifiers (e.g., different classification principles, or trained in different portions of the input space). It is important to note, the role that the architecture plays in the performance of the EC, specifically the ensemble size (e.g., the number of classifiers), the pool of classifiers (e.g., only NN-based classifiers), the use of confidence weights for the fusion (e.g., calculated from the validation data after the training process). We propose a supervised EC architecture using evidence theory. Fig. \ref{fig__ECET_Multiclass__multi} displays the main components of the architecture, namely the number of classifiers, the transformation of prediction to a set of evidence, the information fusion, and the transformation of the set of evidence to a prediction.   

Formally, the EC consists of a $N$ number of classifiers trained with a  training dataset $D^{Tr}$ with $o_{Tr}$ number of observations and $f_{Tr}$ number of features, where $o_{Tr}, f_{Tr}, N \in \mathbf{N}$. The confidence weights of the classifiers are calculated using a performance metric and the validation dataset $D^{Va}$ with $o_{Va}$ number of observations and $f_{Va}$ number of features, where $o_{Va}, f_{Va} \in \mathbf{N}$. These weights are used during the information fusion to weigh the set of evidence of each classifier.

The framework of discernment $\Theta$ is described as: $\Theta = \{F_{1},...,F_{M}\}$, where $M$ is the number of classes or faults, and $M \in \mathbf{N}$. Thus, the prediction of a classifier $\hat{y_{i}}$ takes a values in $\Theta$, while feeding the testing dataset $D^{Te}$ with $o_{Te}$ number of observations and $f_{Te}$ number of features, where $o_{Te}, f_{Te} \in \mathbf{N}$. The prediction $\hat{y_{i}}$, then, is transformed into a set of evidence $m_{i}$ using Eq. (\ref{pred2evid__1})-(\ref{pred2evid__17}). The set of evidence $m_{i}$ is a row vector $1 \times M+1$, which includes in its last element the overall uncertainty. %Algorithm \ref{algorithm_approach__1} shows the steps to transform a prediction into a set of evidence.  

% \begin{algorithm}[!ht]
% \caption{\textcolor{red}{Prediction to Evidence} }\label{algorithm_approach__1}
% \begin{algorithmic}[3]
% \Procedure{pred2Evid}{}
% \State $N_{P}\gets calculating\_pool\_size()$
% \If{$U_{EN_{D}}<Tr_{D_{Min}} \&\& U_{EN_{Y}} < Tr_{Y_{Min}}$ }
%     \State Train model 
%     \State Validate model and obtaining performance $P_{M_{i}}$ \Comment by Eq. (??)
% \Else
        
% \EndIf
% \State calculate diversity $D$ of the pool by Eq.(??)
% \State \textbf{return $S$, $w$} 
% \EndProcedure
% \end{algorithmic}
% \end{algorithm}

The next step is the combination or fusion of the set of evidence of each classifier; refer to Algorithm \ref{algorithm_approach__1}. For this purpose, we apply the rules of combination DSRC and YRC using the Eq. (\ref{dempster__1})-(\ref{dempster__3}) to obtain the ensemble set of evidence $F_{D_{i}}, F_{Y_{i}}$, using DSRC and YRC respectively. Important to mention is that while effectuating the information fusion, the ensemble uncertainties $U_{D_{i}}, U_{Y_{i}}$ are obtained.     

\begin{algorithm}[!htt]
\caption{Ensemble Classifier using Evidence Theory}\label{algorithm_approach__1}
\begin{algorithmic}[2]
\Procedure{Ensemble Classifier}{}
\State initialize selection of classifiers
\State initialize $N_{SEL_{EN}}$
    \For{$i=1$ to $N_{SEL_{EN}}$}
        \State $\hat{y_{i}} \gets Mo_{i}(D^{Te})$
        \State $m_{i} \gets pred2Evid(\hat{y_{i}},w^{Mo}_{i})$ \Comment by Eq.(\ref{pred2evid__1})-(\ref{pred2evid__17})
        \If{i = 1}
            \State $F_{D_{i-1}} = F_{Y_{i-1}} = 0$
            \State $m_{D_{i-1}} = m_{Y_{i-1}} = m_{i}$
            \State $U_{D_{i-1}} = U_{Y_{i-1}} = 0$
        \Else 
            \State $F_{D_{i}}$, $U_{D_{i}}=m_{i} \otimes m_{D_{i-1}}$ \Comment by Eq.(\ref{dempster__1})-(\ref{dempster__3}) 
            \State $F_{Y_{i}}$, $U_{Y_{i}}=m_{i} \otimes m_{Y_{i-1}}$ \Comment by Eq.(\ref{yager__1})-(\ref{yager__3})
            \State $m_{D_{i-1}}=F_{D_{i}}$
            \State $m_{Y_{i-1}}=F_{Y_{i}}$
        \EndIf
    \EndFor
    \State $\hat{y_{EN}} \gets conv2Label(F_{D_{i}})$ \Comment by Eq.(\ref{argmax})
    \State $U_{D} \gets U_{D_{i}}$ 
    \State $U_{Y} \gets U_{Y_{i}}$ 
\State \textbf{return $\hat{y_{EN}}$, $U_{D}$, $U_{Y}$} 
\EndProcedure
\end{algorithmic}
\end{algorithm}	

The last step consists of the transformation of the set of evidence of EC $F_{D_{i}}$ into an ensemble prediction $\hat{y_{EN}}$ using an \textit{argmax} function: %$\hat{y_{EN}} = \operatorname*{arg\,max}_\Theta F_{D_{i}}$, 
  \begin{equation} \label{argmax}
  	\begin{split}
  		%\hat{y_{EN}} = argmax f(x) = 1- \sum_{i=1}^{N} p_{i}*w_{p_{i}}\\
  		\hat{y_{EN}} = \operatorname*{arg\,max}_\Theta F_{D_{i}}
  	\end{split}
  \end{equation}
%   \begin{equation} \label{argmax}
%   	\begin{split}
%   		%\hat{y_{EN}} = argmax f(x) = 1- \sum_{i=1}^{N} p_{i}*w_{p_{i}}\\
%   		\hat{y_{EN}} = \operatorname*{arg\,max}_\Theta F_{D_{i}}
%   	\end{split}
%   \end{equation}
where $\hat{y_{EN}} \in \Theta$.   

% \begin{algorithm}[!htt]
% \caption{\textcolor{red}{Conversion to Label}}\label{algorithm_approach__3}
% \begin{algorithmic}[3]
% \Procedure{conv2Label}{}
% \State $N_{P}\gets calculating\_pool\_size()$
% \If{$U_{EN_{D}}<Tr_{D_{Min}} \&\& U_{EN_{Y}} < Tr_{Y_{Min}}$ }
%     \State Train model 
%     \State Validate model and obtaining performance $P_{M_{i}}$ \Comment by Eq. (??)
% \Else
        
% \EndIf
% \State calculate diversity $D$ of the pool by Eq.(??)
% \State \textbf{return $S$, $w$} 
% \EndProcedure
% \end{algorithmic}
% \end{algorithm}

\subsection{Criteria for the pool selection of classifiers}\label{criteria__pool__selection}
\begin{figure*}[!ht]
	\centering
		\includegraphics[width=0.8\textwidth,keepaspectratio]{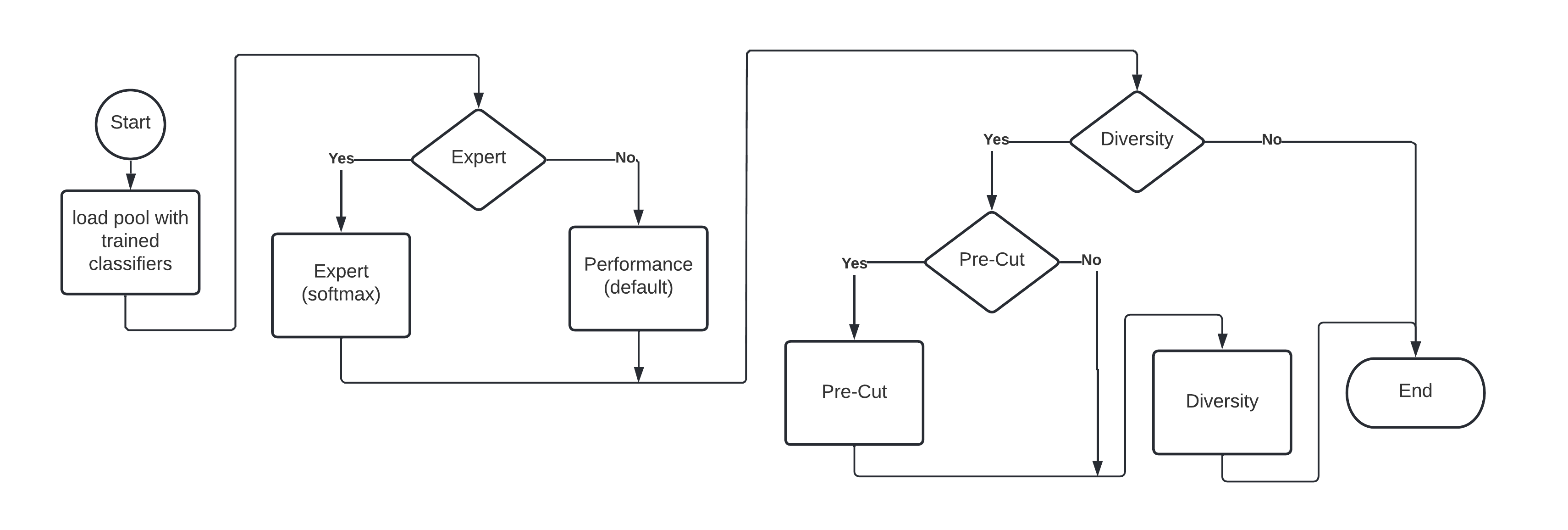}
	\caption{Ensemble selection.}\label{fig__ensemble__selection}
\end{figure*}

The criteria selection of classifiers to build the pool for the ensemble classifier plays a vital role. There are many considerations when selecting a combination of classifiers for the ensemble. We propose the following criteria: performance, expert area, and diversity. Besides, we use an extra parameter pre-cut that affects only the diversity. Fig. \ref{fig__ensemble__selection} shows the ensemble selection procedure.

\subsubsection{Performance}
A common practice for classifiers selection is sorting out the classifiers according to a performance metric. To this end, we propose using the F1-score as a performance metric because it is not sensitive to unbalanced classes. Besides, the \textit{F1-score} (F1) combines the metrics \textit{precision} (PR) and \textit{recall} (RE) for its calculation. RE is also known as fault detection rate (FDR). A detailed description, of how F1, PR and RE are calculated, is presented in \cite{Panda2021}. 

% The \textit{precision (PR)} is the ratio between the \textit{true positives (TP)} and the sum of \textit{true positives (TP)} and \textit{false negatives (FN)}:

% \begin{equation} \label{eq26}
%   \begin{split}
%   	PR = \frac{TP}{TP+FN}
%   \end{split}
% \end{equation}

% The \textit{recall (RE)} is the ratio between the \textit{true positives (TP)} and the sum of \textit{true positives (TP)} and \textit{false positives (FP)}:

% \begin{equation} \label{eq27}
%   \begin{split}
%   	RE = \frac{TP}{TP+FP}
%   \end{split}
% \end{equation}

% Finally, the \textit{F1-score (F1)} is calculated as follows:

% \begin{equation} \label{eq28}
%   \begin{split}
%   	F1 = 2*\frac{PR*RE}{PR+RE}
%   \end{split}
% \end{equation}

\subsubsection{Expert Area}
The motivation of an expert area lies in the ability of certain classifiers to excel in learning specific classes \cite{JiaoGuo2022}. This ability is an important feature, because an "expert" classifier could improve the ensemble in the specific class where the expert classifier excels. We propose using a \textit{softmax}-based strategy that identifies the expertness of classifiers. Lastly, the pool is sorted out using the expert area criteria.

The \textit{softmax} function $\sigma$ of a vector $\mathbf{z}$ fulfills $\sigma: \mathbf{R}^{K} \rightarrow (0,1)^{K}$, for $K \in \mathbf{N}$. Thus, the \textit{softmax} function $\mathbf{z}_{i}$ is defined using:   
\begin{equation} \label{expert__1}
   \begin{split}
   	\sigma(\mathbf{z})_{i} = \frac{e^{z_{i}}}{\sum_{j=1}^{K} e^{z_{j}}}
   \end{split}
\end{equation}
where $i,j \in \mathbf{N}$.

The expert area of an ensemble classifier $EXP_{EN}$ extends Eq. (\ref{expert__1}) to measure the degree of expertness of a classifier $Mo_{i}$ with regard to a class $F_{i}$. Thus, the equation is defined as:
\begin{equation} \label{expert__2}
   \begin{split}
   	\sigma_{EN}(\mathbf{z})_{i,k} = \frac{e^{z_{i}^{k}}}{\sum_{j=1}^{K} e^{z_{j}^{k}}}
   \end{split}
\end{equation}
where $z_{i}^{k}$ represents the prediction of the $i_{th}$ classifier for the class $k$, $K$ is the number of classifiers in the pool, and $i,j,k \in \mathbf{N}$.

As stated above, an expert classifier could detect a class with high performance while having a bad performance with the rest of the classes. For this reason, the results of the softmax function $\sigma_{EN}$ apply a selection of the classifiers with respect to a class. In order to increase the numerical separation between the elements of the $k_{th}$ class, each element of $\sigma_{EN_{msk}}(\mathbf{z})$ is masked using:
% \begin{equation} \label{expert__3}
%      \sigma_{EN}(\mathbf{z})_{i,k} = \begin{cases}
%          VA_{Max} ,\quad\quad \text{if}\ Dif > TH_{Max} \\
%          VA_{Min}=1/N_{C} ,\quad\quad otherwise  \\
%          \end{cases}
% \end{equation}
\begin{equation} \label{expert__3}
     \sigma_{EN_{msk}}(\mathbf{z})_{i,k} = \begin{cases}
         VA_{Max}/|S_{\sigma_{EN}(\mathbf{z})_{i,k}}| ,\quad \text{if}\ \\
         \quad\quad\quad S_{\sigma_{EN}(\mathbf{z})_{i,k}} \subset max(\sigma_{EN}(\mathbf{z})_{i,k}) \\
         VA_{Min} ,\quad\quad otherwise  \\
         \end{cases}
\end{equation}

where $S_{\sigma_{EN}(\mathbf{z})_{i,k}}$ represents the subset of maximum values of the $\sigma_{EN}(\mathbf{z})_{i,k}$ evaluated in a specific class $k$, and $|S_{\sigma_{EN}(\mathbf{z})_{i,k}}|$ is the cardinality of the subset. $VA_{Max}$ and $VA_{Min}$ are the upper and lower values assigned to the $i_{th}$ element of the $k$ class, respectively. Where $VA_{Max}>0, VA_{Min} \geq 0$ and  $\in \mathbf{R}$, and $VA_{Max} \gg VA_{Min}$. The latest considerations for $VA_{Max}$ and $VA_{Min}$ play a key role in the numerical separation and, therefore, for the assignment of expertness to a classifier. For this reason, it is recommendable at least a separation of 10 during the application of the expert area to the pool selection (e.g., $\frac{VA_{Max}}{VA_{Min}} \approx 10$).

The next step is summing the results per fault on each classifier using the following equation:
\begin{equation} \label{expert__5}
   \begin{split}
   	\sigma_{EN_{msk-sm}}(\mathbf{z})_{i} = \sum_{k=1}^{N_{C}} \sigma_{EN_{msk}}(\mathbf{z})_{i,k}
   	\end{split}
\end{equation}
where $N_{C}$ represents the number of classes.
The last step is applying a softmax to $\sigma_{EN_{msk-sm}}(\mathbf{z})$ using Eq. (\ref{expert__2}):%, in order to obtain $\sigma_{EN_{msk-sm}}(\mathbf{z})$:
\begin{equation} \label{expert__4}
   \begin{split}
   	EXP_{{EN}_{i}} = \frac{e^{\sigma_{EN_{msk-sm}}(\mathbf{z})_{i}}}{\sum_{j=1}^{K} e^{\sigma_{EN_{msk-sm}}(\mathbf{z})_{i}}}
   \end{split}
\end{equation}
where $EXP_{{EN}_{i}}$ represents the expertness of the $i_{th}$ classifier.

% The next step is applying a softmax to $\sigma_{EN_{msk}}(\mathbf{z})$ using Eq. \ref{expert__2}, in order to obtain $\sigma_{EN_{msk-sm}}(\mathbf{z})$:
% \begin{equation} \label{expert__4}
%    \begin{split}
%    	\sigma_{EN_{msk-sm}}(\mathbf{z})_{i,k} = \frac{e^{\sigma_{EN_{msk}}(\mathbf{z})_{i,k}}}{\sum_{j=1}^{K} e^{\sigma_{EN_{msk}}(\mathbf{z})_{i,k}}}
%    \end{split}
% \end{equation}
% Finally, the expertness of each classifier $EXP_{{EN}_{i}}$ can be calculated using the equation:
% \begin{equation} \label{expert__5}
%    \begin{split}
%    	EXP_{{EN}_{i}} = \sum_{k=1}^{N_{C}} \sigma_{EN_{msk-sm}}(\mathbf{z})_{i,k}
%    	\end{split}
% \end{equation}
% where $EXP_{{EN}_{i}}$ represents the expertness of the $i_{th}$ classifier, and $N_{C}$ represents the number of classes.

\subsubsection{Diversity}
The strength of ensemble classification derives from the heterogeneous nature of its information sources. For this purpose, it is necessary to have a grounded strategy that considers the diversity between their sources. There are different methods in Literature that aboard this topic, specifically the diversity measurement between the predictions of two classifiers, namely the disagreement measure, Kohavi-Wolpert variance, generalized diversity, and misclassification diversity \cite{Chen2019} \cite{Tang2006} \cite{Yanzhao2021} \cite{JanVerma2019}.
We concentrate on the following criteria: data nature, classification principle, and measuring diversity. The authors use the first two criteria considering the classifiers' nature according to the available Literature. The last criterion has been explored in Literature using different diversity measurement methods.

\textbf{Data nature} considers whether the data is linear or non-linear separable. Some classifiers can handle only linear separable data (e.g.,  perceptron, linear support vector machine, and linear regression). Other classifiers, such as the support vector machines (SVM), can perform a data transformation, where the non-linear data is transformed into several dimensions using the kernel trick \cite{Richards2013}. After this transformation, the data can be linearly separable, with the inconvenience of an increase in computing power.

\textbf{The classification (mathematical) principle} has an essential role in this approach because each of the considered classifiers handles the input space differently. Common supervised classification principles are based on how the methods identify the regions in the input space (an extensive review on this topic can be found in \cite{Richards2013}):
\begin{itemize}

    \item \textit{Maximum Likelihood Classification}: A popular method is the naive Bayes classifier, which uses conditional probability with the assumption that the input features are mutually independent: 
   $P(A|B) = \frac{P(B|A) \dot P(A)}{P(B)}$
   where $P(A|B)$ represents the posterior (or the prediction), $P(A)$ is the prior (assigned with expert domain knowledge), $P(B|A)$ is the likelihood, and $P(B)$ is the evidence.

    \item \textit{Spectral measurement space}: To this category belongs the methods k nearest neighbor (KNN) and support vector machines (SVM). KNN parts from the premise that an instance of the input space is likely to belong to K neighbors instances of a certain class. This method commonly uses the euclidean distance to determine the membership of the instance. 
    Support vector machines proved to be powerful to separate the classes using (high-dimensional) hyperplanes: $\mathbf{w}\mathbf{x}-b = 1$. As mentioned above, one strength of the method relies on the use of the kernel trick to perform a data transformation, which makes the data linearly separable.      
    
    \item \textit{Committees of Classifiers}: This method implies training several algorithms of the same type (e.g., decision tree) in parallel and performing a majority voting on the classifier predictions to obtain the class labels. The committees of classifiers include the approaches using boosting (e.g., AdaBoost, XGBoost) and bagging (e.g., Random Forest). 

    \item \textit{Networks of Classifiers or layered classifiers}: The decision trees, committees of classifiers, and artificial neural networks or neural network-based (NN) models belong to this category.  
    Decision trees use a decision-like model in which the attributes are tested using decision nodes, and their outcomes generate new branches. The leaf nodes contain the class labels. 
    The basic unit of an NN-based model is the perceptron. The perceptron is represented using: $\hat{\mathbf{y}} = f(\mathbf{w}{^T}\mathbf{x}+b)$, where $x$ represents the inputs, $w$ are the weights, $\hat{\mathbf{y}}$ is the predicted output, and $b$ are the biases. The basic NN-based model that can handle non-linearities is the multilayer perceptron, which uses (several) layers of neurons and a non-linear output layer. From this basis, other architectures have been proposed that consider stacked layers such as convolutional NN (CNN).  
    \end{itemize}
    
% input space (graphic TSNE/PCA) and selecting random times a random sample of classifiers
Chen et al. \cite{Chen2019} propose a diversity measurement for an ensemble classifier. The first step is the definition of a \textbf{diversity measurement} $Div(i,j)$ between two classifiers:
    
    \begin{equation} \label{approach__criteria__div__1}
       \begin{split}
       	Div(i,j) = \frac{N_{diff}}{N}
       \end{split}
    \end{equation}

where the subindex $i$ corresponds to the $i_{th}$ predicted class of the first classifier, subindex $j$ is the $j_{th}$ predicted class of the second classifier, $N$ is the number of samples, and $N_{diff}$ represents the number of misclassifications respect to the ground truth by the two classifiers. It is important to note that both classifiers must classify the sample wrongly under test. 
The second step is the definition of a diversity measurement of the ensemble classifier (EC), which considers all the individual classifiers that form the EC:
    \begin{equation} \label{approach__criteria__div__2}
       \begin{split}
       	DIV_{EN} & = \frac{1}{K} \sum_{i=1}^{D_{V}} \sum_{j=1}^{D_{V}} Div(i,j)\\
       \end{split}
    \end{equation}
where $i \neq j$, and $K$ is the number of classifiers in the pool. 

A variation of \cite{Chen2019} considers the diversity measurement between two classifiers $Div(i,j)$, where $N_{diff}$ counts the number of misclassifications when at least one classifier is wrong with respect to the ground truth, also known as the disagreement measure \cite{KunchevaWhitaker2003}.

\subsubsection{Pool Selection}\label{approach__pool_selection}

Now in this approach, we consider an ensemble classifier of size m, which means that the predictions of m classifiers will be combined using information fusion. Considering a pool of $N_{P}$ classifiers and an ensemble size $N_{ES}$, it generates a $N_{C}$ number of combinations using the equation:
    \begin{equation} \label{approach__criteria__1}
       \begin{split}
       	C(N_{P},N_{ES}) = \frac{N_{P}!}{(N_{ES}!(N_{P}-N_{ES})!)}
       \end{split}
    \end{equation}
which might be a trivial question when having $N_{C} = C(5,3) = 10$ combinations, but in the case of a pool size of $N_{P} = 10$ classifiers and an ensemble size of $N_{ES} = 5$: $N_{C} = C(10,5) = 252$ combinations.
This situation brings a new challenge since each combination is a potential experiment that needs to be performed. Considering all the ensemble sizes $N_{ES} = 2..10$ and a pool size of $N_{P} = 10$, that generates 1013 combinations or experiments.
 
The present approach reduces the number of combinations for a pair ($N_{P}$,$N_{ES}$) to ten possible experiments (e.g., $N_{C} = C(10,5) = 10$), see Table \ref{table__results__pool_selection__combination}.

\begin{table}[!ht]
\centering
\caption{Parameters Combinations for Pool Selection}
\begin{tabular}{c|c|c|c|c}
\hline
\multicolumn{1}{l|}{\textbf{Combination}} & \multicolumn{1}{l|}{\textbf{Exp}} & \multicolumn{1}{l|}{\textbf{Div}} & \multicolumn{1}{l|}{\textbf{Ver}} & \multicolumn{1}{l}{\textbf{PC}} \\ \hline
Co1 & False & False & False & False \\ %\hline
Co2 & False & True & False & False \\ %\hline
Co3 & False & True & False & True \\ %\hline
Co4 & False & True & True & False \\ %\hline
Co5 & False & True & True & True \\ \hline
Co6 & True & False & False & False \\ %\hline
Co7 & True & True & False & False \\ %\hline
Co8 & True & True & False & True \\ %\hline
Co9 & True & True & True & False \\ %\hline
Co10 & True & True & True & True \\ \hline
\end{tabular}\label{table__results__pool_selection__combination}
\end{table}

The parameters under consideration are the expert criteria $Exp$, the diversity $Div$, the version of diversity $Ver$, and the pre-cut $PC$. 

The pre-cut is a parameter that only affects the diversity because it reduces the size of the pool:
    \begin{equation} \label{approach__criteria__2}
       \begin{split}
       N_{P} = N_{ES} + 1
       \end{split}
    \end{equation}
Having defined the different criteria for the pool selection, it is possible to summarize the strategy for the ensemble selection using a Pseudo-Code, see Algorithm \ref{algorithm_approach__4}.

\begin{algorithm}[!ht]
\caption{Selection of Classifiers}\label{algorithm_approach__4}
\begin{algorithmic}[1]
\Procedure{Selection of Classifiers}{}
\State initialize pool of classifiers
\State initialize hyperparameters for classifiers
\State $N_{P}\gets calculating\_pool\_size()$
    \For{$i=1$ to $N_{P}$}
        \State $\hat{y_{i}} \gets Mo_{i}(D^{Tr})$ %\Comment ANN? use Eq. (\ref{adaptive__1})-(\ref{adaptive__3})
        \State $\mathbf{P_{Mo_{i}}}$, $w_{Mo_{i}} \gets VA(y_{i},\hat{y}_{i}) $ \Comment Using F1-score%By Eq. (\ref{eq28})
        \State $U_{Mo_{i}} \gets UQ(y_{i},\hat{y}_{i})$ \Comment By Eq. (\ref{uncertainty__quant__1})-(\ref{uncertainty__quant__3})
        % \State $w_{Mo_{i}} \gets f(U_{Mo_{i}})$ \Comment By Eq.(??)
    \EndFor
\State $P_{EN} \gets EX(P_{Mo_{i}})$ \Comment By Eq.(\ref{expert__1})-(\ref{expert__5})
\State $P_{EN} \gets DV(\mathbf{y},\hat{\mathbf{y}}, P_{EN}, PC)$ \Comment By Eq.(\ref{approach__criteria__div__1})-(\ref{approach__criteria__div__2})
% \State $\textbf{Pool_{EN}} \gets SE(P,D,E)$ \Comment By Eq.(??)
\State \textbf{return $P_{EN}$, $w_{Mo}$} 
\EndProcedure
\end{algorithmic}
\end{algorithm}

\subsection{Uncertainty Quantification}\label{approach__uncert_quantif}

The uncertainty quantification (UQ) provides a glimpse of the learning capability of the (ensemble) classifier. The UQ can be performed during training, validation, and testing. After training, the UQ is used to assess the learning process using validation data (e.g., EC performance). The UQ can assess anomaly detection while feeding unknown data using an uncertainty tracking strategy during testing.
We propose three ways to quantify the uncertainty: using a performance metric and the rules of combination DSET and Yager.  

The UQ using a performance metric $U_{Q_{P}}$ is represented by:
\begin{equation} \label{uncertainty__quant__1}
   \begin{split}
   	UQ_{P}=\prod_{i=1}^{N} (1-P_{i})\\
   \end{split}
\end{equation}

where $N$ is the number of trained labels or classes, $P_{i}$ is the performance of the $i_{th}$ class, and $N,i \in \mathbf{N}$.

The UQ using the rule of combination DSET is modeled using $b_{k}$ from Eq. (\ref{dempster__2}): 
\begin{equation} \label{uncertainty__quant__2}
   \begin{split}
   	UQ_{DS} = b_{k}\\
   \end{split}
\end{equation}
The UQ using the rule of combination YAGER is modeled using Eq. (\ref{yager__3}): 
\begin{equation} \label{uncertainty__quant__3}
   \begin{split}
   	UQ_{Y} = q(\phi)\\
   \end{split}
\end{equation}
The terms $b_{k}$ and $q(\phi)$ are calculated using the Eq. (\ref{dempster__2}):
$\sum_{A_{i}\bigcap B_{j}=\phi}^{}m_{1}(A_{i})\times m_{2}(B_{j})$. 

We quantify the uncertainty of the \textbf{individual classifiers} (e.g., SVM, alexnet) using $UQ_{P}$, $UQ_{DS}$, and $UQ_{Y}$. It is important to note that the performance metric is only used for individual classifiers because it relies on supervised classification. The latest means that the labels are known after training the classifier, so the performance metric can be applied.
$UQ_{DS}$ and $UQ_{Y}$ are calculated using a random validation batch. Each batch contains a number $M$ of samples of the validation data $D^{Va}$. The classifier is evaluated using a fixed number of iterations $i$, in which a new random validation batch is used. The predictions of each batch are combined using the rules of combination DSET and Yager to obtain $UQ_{DS}$ and $UQ_{Y}$, respectively. Furthermore, the predictions of each batch are used to calculate $UQ_{P}$.  

In the case of \textbf{ensemble classification}, we feed the data to the EC, but the ground truth remains unknown. For this reason, we quantify the uncertainty using the obtained $UQ_{DS}$ and $UQ_{Y}$ from the EC.

\subsection{Anomaly Detection using Uncertainty Quantification}\label{approach__AD}
The EC can predict the classes which are used during the training process. However, the EC has no learning capability once the training is consumed, which means that the EC excels in the classification task while the data corresponds to known classes. The constraint here lies in that the EC classifies the data of an unknown condition into the known classes which were used during the training phase. 
We propose an anomaly detection capability for the ECET, which is based on the uncertainty monitoring of $U_{D}$ and $U_{Y}$. This feature identifies anomalies while feeding data of unknown conditions; refer to Algorithm \ref{algorithm_approach__5}. This function, however, generates a new anomaly prediction $\hat{y}_{AN}$ and a new anomaly framework of discernment $\hat{\Theta}_{AN}$. In case of discovering a new unknown condition $K$, the frame of discernment $\hat{\Theta}_{AN}$ increases in one element: 
\begin{equation} \label{approach__AD__1}
   \begin{split}
   	\hat{\Theta}_{AN} &= \Theta + {A_{K}} \\
   	                  &= \{F_{1},...,F_{N}, A_{K}\} 
   \end{split}
\end{equation}
where $N, K \in \mathbf{N}$.

The EC provides the uncertainties $U_{D}$ and $U_{Y}$ while feeding a data sample. The procedure checks whether the uncertainties $U_{D}$ and $U_{Y}$ exceed the maximum thresholds $Tr_{D_{Mx}}$ and $Tr_{Y_{Mx}}$. If the case is affirmative, then the anomaly detection predictor $\hat{y}_{AN}$ receives a new class $A_{K}$, where $K \in \mathbf{N}$. In case the uncertainties lie below the thresholds, the anomaly prediction $\hat{y}_{AN}$ receives the value of the ensemble prediction $\hat{y}_{EN}$.    

\begin{algorithm}
\caption{Anomaly Detection using ECET}\label{algorithm_approach__5}
\begin{algorithmic}[3]
\Procedure{Anomaly Detection}{}
\State $N_{P}\gets calculating\_pool\_size()$
\If{$U_{D}<Tr_{D_{Mn}}$ \AND $U_{Y} < Tr_{Y_{Mn}}$ }
    \State $\hat{y_{AN}} \gets conv2Label(F_{D_{i}})$ \Comment by Eq.(\ref{argmax})
\ElsIf{$U_{D} > Tr_{D_{Mx}}$ \AND $U_{Y} > Tr_{Y_{Mx}}$ }
    \State $\hat{y_{AN}} \gets A_{K} $   
\Else
    \State $\hat{y_{AN}} \gets conv2Label(F_{D_{i}})$ \Comment by Eq.(\ref{argmax})   
\EndIf
\State  \textbf{return} $\hat{y_{AN}}$ %\textbf{return $S$, $w$} 
\EndProcedure
\end{algorithmic}
\end{algorithm}

\section{Use Case: Anomaly Detection using ECET on the Tennessee Eastman Dataset}\label{section__usecase}

This section presents the results of the classification performance and anomaly detection of ECET using the benchmark dataset Tennessee Eastman. For this purpose, we first present a description of the dataset and the considerations that are taken for the experiment design (e.g., data preparation, the pool of classifiers, and performance metrics). The subsection results provide the experiment's outcome for the uncertainty quantification, %optimized training using ALR-UQ, 
pool selection, classification performance, and anomaly detection. Two final subjects close the subsection: a comparison with literature and the discussion of the results.

\subsection{Description of the Dataset}
\begin{figure}[!ht]
	\centering
    \includegraphics[width=0.45\textwidth,keepaspectratio]{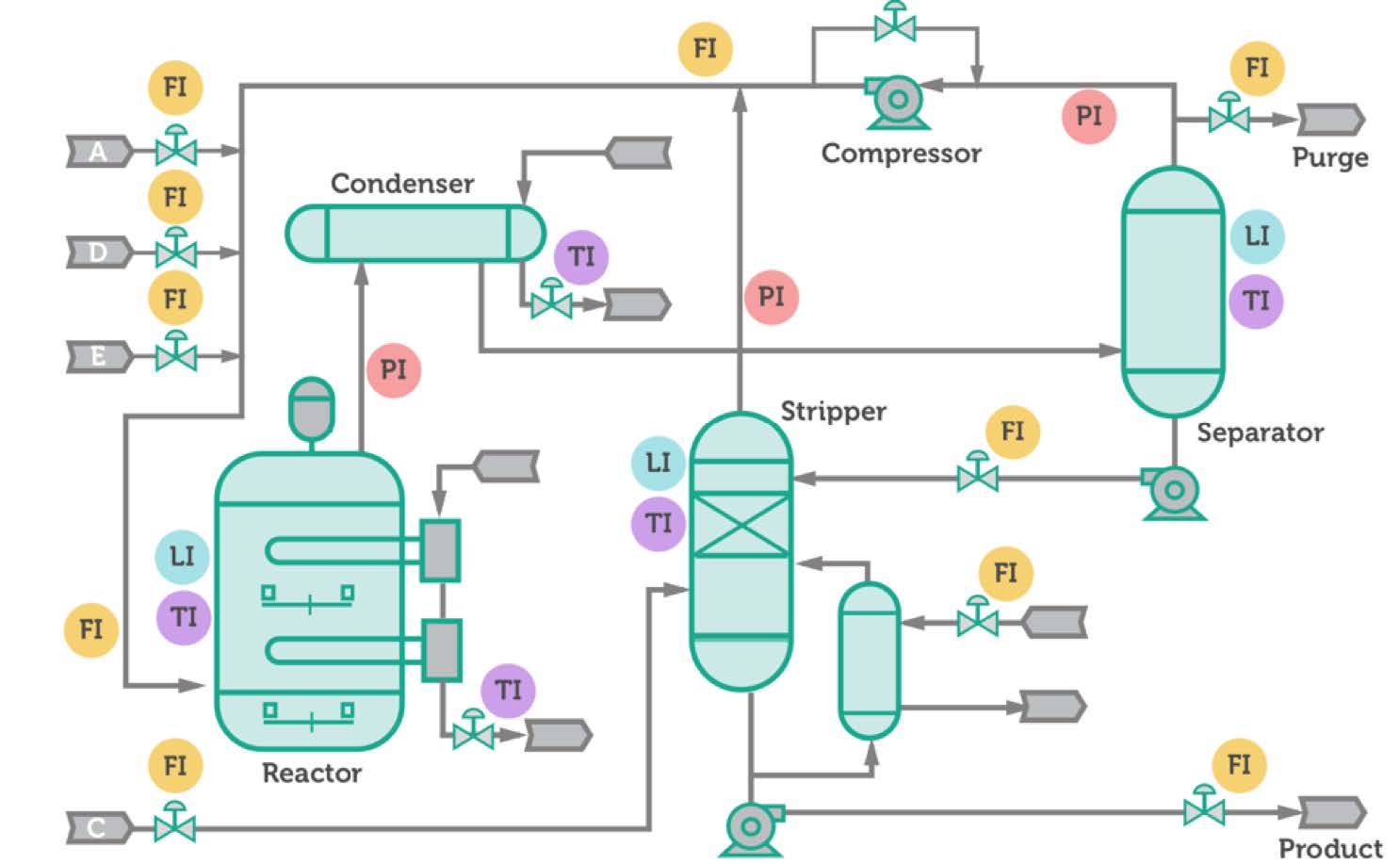}
    \caption{Tennessee Eastman plant diagram
    \cite{Downs1993}\cite{ChenData2019}.}\label{fig__TE_diagram}
\end{figure}

Down and Vogel created a simulation that replicated the process of the Tennessee Eastman chemical plant \cite{Downs1993}. The process consists of five principal process units: the reactor, condenser, recycle compressor, vapor-liquid separator, and product stripper. The plant produces two liquid products, G and H while using four gaseous inputs, A, D, E, and C. Fig. \ref{fig__TE_diagram} displays the piping and instrumentation diagram (P\&ID). The benchmark is a popular dataset in the research community because it provides a challenging use case for supervised classification and clustering approaches.   

The TE dataset consists of 21 class faults and a normal condition. The TE dataset consists of training and testing datasets for each case; see Table \ref{dataset__fault__cases}. 
A training dataset consists of 480 samples for each case. The testing dataset consists of 960 samples, where 160 correspond to the normal condition and 800 samples for the fault case. The dataset has 52 input variables.  
There are fault cases that are especially challenging while performing classification (e.g., fault cases 3, 9, 15, and 21). The fault cases have been grouped into three categories: easy (1, 2, 4, 5, 6, 7, 12, 14, and 18), medium (8, 10, 11, 13, 16, 17, 19, 20) and hard (3, 9, 15 and 21) \cite{ChadhaPanambilly2020}. For this reason, it is a usual practice to select the hard faults to test the robustness of the approaches. The results of subsections \ref{use_case__EC}, \ref{use_case__AD}, and \ref{comparison__literature} use the data of the hard faults in order to show the ECs' performance.

% \begin{figure}[!htbp]
% 	\centering
% 	\includegraphics[width=0.45\textwidth,height=12.0cm,keepaspectratio]{TE_diagram}
% 	\caption{Tennessee Eastman Plant diagram \textcolor{blue}{REPLACE}
%  \cite{Downs1993}}\label{fig__TE_diagram}
% \end{figure}

\begin{table}[!ht]
\centering
\caption{TE Dataset fault cases.}
\begin{tabular}{ccc}
\hline
\textbf{Fault} & \textbf{Description}                           & \textbf{Type}     \\ \hline
0                    & Normal Condition                               & -                 \\
1                    & A/C ratio, B composition cons. (Stream 4)   & Step              \\
2                    & 2 B composition, A/C ratio const. (Stream 4) & Step              \\
3                    & 3 D feed temp. (Stream 2)                & Step              \\
4                    & 4 Reactor cooling water supply temp.              & Step              \\
5                    & 5 Condenser cooling water supply temp.            & Step              \\
6                    & 6 A feed loss (Stream 1)                       & Step              \\
7                    & 7 C header pressure loss (Stream 4)            & Step              \\
8                    & 8 A, B, C feed composition (Stream 4)          & Random            \\
9                    & 9 D feed T° (Stream 2)                         & Random            \\
10                   & 10 C feed T° (Stream 4)                        & Random            \\
11                   & 11 Reactor cooling water supply temp.    & Random            \\
12                   & 12 Condenser cooling water supply temp.  & Random            \\
13                   & 13 Reaction Kinetics                           & Slow drift        \\
14                   & 14 Reactor cooling water valve                 & Sticking          \\
15                   & 15 Condenser cooling water valve               & Sticking          \\
16                   & Unknown                                        & -                 \\
17                   & Unknown                                        & -                 \\
18                   & Unknown                                        & -                 \\
19                   & Unknown                                        & -                 \\
20                   & Unknown                                        & -                 \\
21                   & 21 A, B, C feed valve (Stream 4)               & Const. pos.\\ \hline
\end{tabular}\label{dataset__fault__cases}
\end{table}

\subsection{Experiment Design}
We use the TE dataset to test the ECET capabilities: the classification performance and anomaly detection of the ensemble classifier (EC).
We trained ten classifiers in total: five NN-based models and five non-NN-based models (from now on, they will be referred to as Machine Learning or ML-based models). There are three primary ensemble classifiers (ECs), namely NN-based, ML-based, and Hybrid (a combination of the NN and ML-based models). 
The current approach is developed using python 3.7 under the IDE Spyder from Anaconda. The models were defined using the ML frameworks Scikit-learn (for ML models) and PyTorch (for NN-based models) \cite{PedregosaVaroquaux2011} \cite{PaszkeGross2019} \cite{Anaconda2016}. The experiments are performed using a CPU i7-7700 @3.60GHz x 8, 32GB RAM, a GPU NVIDIA GeForce GTX 1660 SUPER, and a Ubuntu 20.04.3 LTS environment.  

\subsubsection{Dataset preparation}
The considerations taken for the experiments are: 

\begin{itemize}
    \item The fault cases are grouped into main sets: (1,2,6,12) and (3,9,15,21). The first group contains a fraction of the easy faults, whereas the second group contains the hard faults. The data of (0,1,2,6,12) conform to an easy dataset, and a hard dataset contains the data of (0,3,9,15,21).   
    \item The binary ECs are trained using the normal condition (0) and one of the fault cases of the datasets (e.g., hard dataset (0,3,9,15,21)), whereas the multiclass ECs are trained using all the cases. 
    \item In the case of the anomaly detection experiments, the datasets are reduced to (0,1) and (0,3) for binary ECs, and (0,1,2,6,12) for multiclass ECs. Given the extent of all the possible combinations of data and ECs, the datasets were selected as representative to show the approach's performance. We use all the fault cases as unknown conditions.    
\end{itemize}

The easy and hard datasets result in a training dataset of 52 input variables and 2900 observations (500 samples for the normal condition + 480 samples per fault case * 5 fault cases), and a testing dataset of 52 input variables and 4800 observations (960 samples for normal condition + 960 samples per fault case * 4 fault cases). It is important to highlight that the first 160 samples of the testing data of a fault case correspond to the normal condition, giving a rest of 800 samples of faulty condition.
In the case of the binary datasets (e.g., (0,1)), the training dataset is composed by 52 input variables and 980 observations (500 samples for the normal condition + 480 samples for the fault case), and a testing dataset of 52 input variables and 1920 observations (960 samples for normal condition + 960 samples for the fault case).

The training dataset is split in a 70/30 ratio to have the training and validation datasets, respectively. The validation dataset is used to quantify the trained classifier's uncertainty and calculate the confidence weights for each class per classifier. The testing dataset is used to determine the classification performance of individual classifiers and ensemble classifiers.

The testing dataset for the anomaly detection experiments differs from the classification experiments in that an unknown fault case is added. As an example, in the case of a binary EC trained with the (training) dataset (0,1), the anomaly detection capability of the EC is tested using the (testing) dataset (0,1,2). For practical purposes, the label of (2) is changed to (-1), the reason of this lies in the fact that EC assigns the label (-1) to unknown conditions. 
The training dataset is scaled for a mean of $\mu=0$ and a standard variance of $\sigma=1$. The scaling parameters of the training data are applied to the testing dataset \cite{RaschkaMirjalili2019}.

\subsubsection{Pool of Classifiers}
\begin{figure}[!ht]
	\centering
	\includegraphics[width=0.45\textwidth,keepaspectratio]{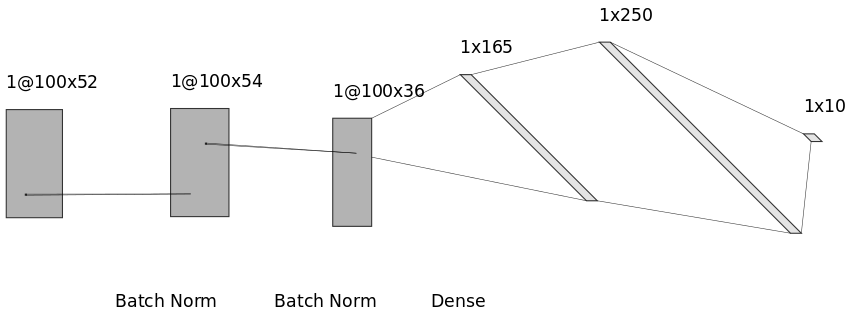}
	\caption{cmp network architecture.}\label{fig__CMP_network}
\end{figure}

\begin{figure}[!ht]
	\centering
	\includegraphics[width=0.15\textwidth,keepaspectratio]{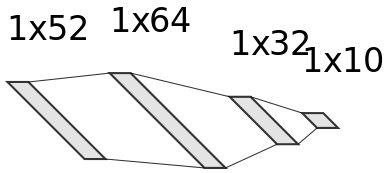}
	\caption{mlp network architecture.}\label{fig__MLP_network}
\end{figure}

The pool of classifiers considers two main groups: NN-based and ML-based models. The ML-based group consists of the classifiers: decision tree (DTR), support vector machine (SVM), K-Nearest-Neighbours (KNN), Naive Bayes (NBY), and AdaBoost (ADB). At the same time, the NN-based group consists of popular models such as alexnet (ale), lenet (len) and vgg. The second group is completed with two customized architectures a multilayer perceptron (mlp) and a deep neural network (cmp). 
%\subsubsection{Hyperparameters}
The hyperparameters for the ML-based models are obtained using the module \textit{gridsearch} from the ML framework \textit{scikit-learn}. 
The hyperparameters (HP) for the ML models are detailed as follows: DTR (criterion='gini', maximal depth=28), SVM(C=1000, gamma=0.1, kernel='rbf'), KNN (metric='manhattan', n-neighbors=3, weights=distance), NBY(no HP) and ADB(lr=0.01, number of estimators=50).  
The architecture and hyperparameters for the NN-based models ale, len, and vgg are documented in detail in \cite{Zhang2021}. However, the 2D convolutional layers are replaced by 1D convolutional layers since the dataset is 1D. The architecture and hyperparameters for mlp and cmp are detailed in Fig. \ref{fig__CMP_network} and Fig. \ref{fig__MLP_network}, respectively. The hyperparameters of cmp are obtained using the module \textit{optuna}. The learning rate is set to lr=0.001, and the number of epochs is set to 20. Experimental results did not show any improvement by increasing the number of epochs.

\subsubsection{Performance Metrics}
We choose the F1 score as the performance metric to monitor the learning capability of each of the experiments. The computation time is also tracked during each of the experiments. The fault detection rate (FDR) is used in section \ref{comparison__literature} to compare the approach's results with literature. 

\subsection{Results}
% \textcolor{red}{NOTE: ALL THE TEXT OF THIS SUBSECTION IS NEW (THIS INCLUDES THE TABLES AND PLOTS)}\\
This section presents the results of the experiments performed using ECET, specifically for the uncertainty quantification (individual classifiers and ECs), classification performance (individual classifiers and ECs), and anomaly detection (AD) (using ECs). 

\subsubsection{Uncertainty quantification of individual classifiers}
\begin{figure*}[!ht]
	\centering
	\begin{subfigure}[b]{0.3\textwidth}
    	\includegraphics[width=\textwidth,keepaspectratio]{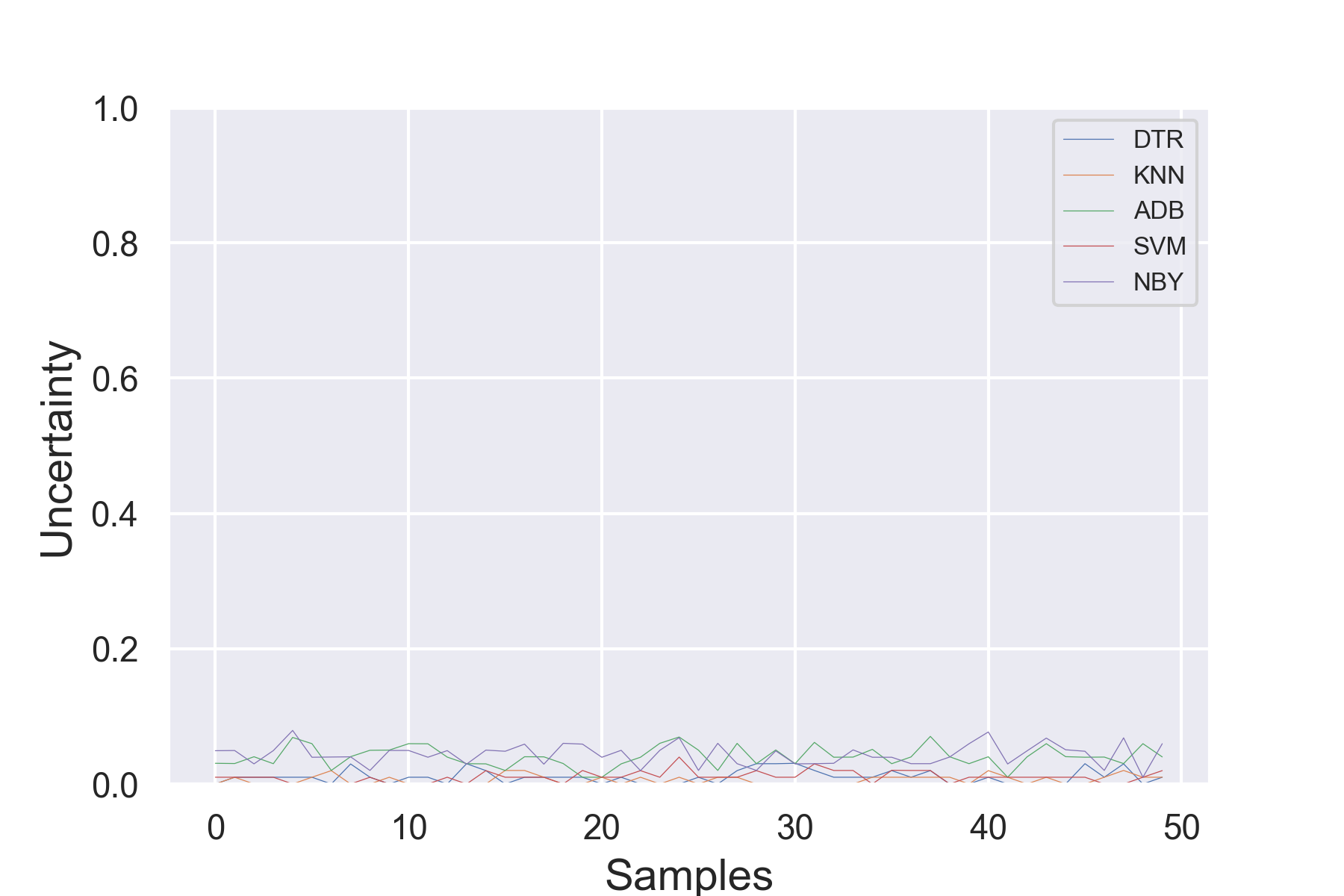}
    	\caption{F1 UQ for ML models}\label{fig__UQF1__ML}
	\end{subfigure}
	%~
	\begin{subfigure}[b]{0.3\textwidth}
    	\includegraphics[width=\textwidth,keepaspectratio]{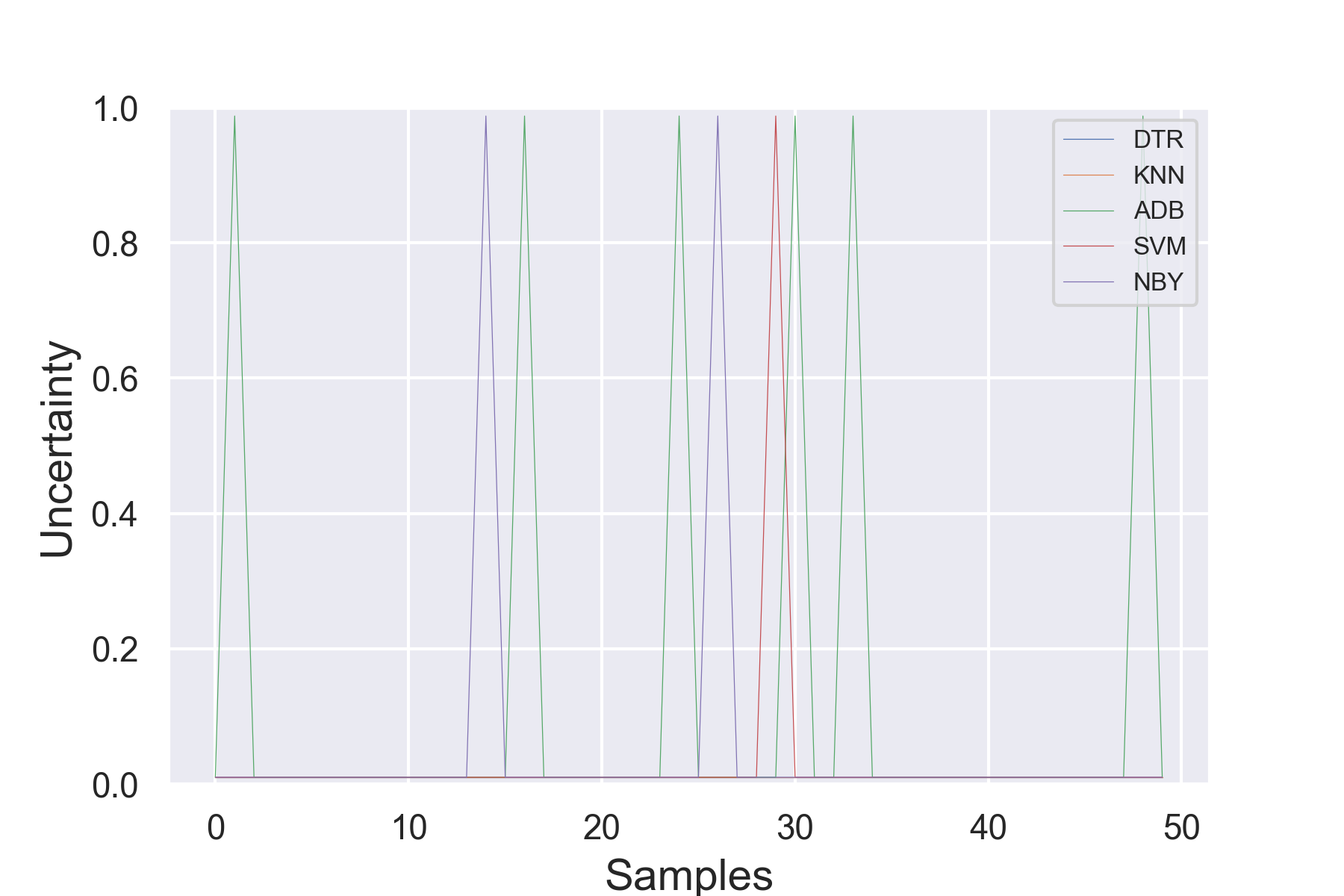}
    	\caption{DSET UQ for ML models}\label{fig__UQDSET__ML}
	\end{subfigure}
	%~
	\begin{subfigure}[b]{0.3\textwidth}
    	\centering
    	\includegraphics[width=\textwidth,keepaspectratio]{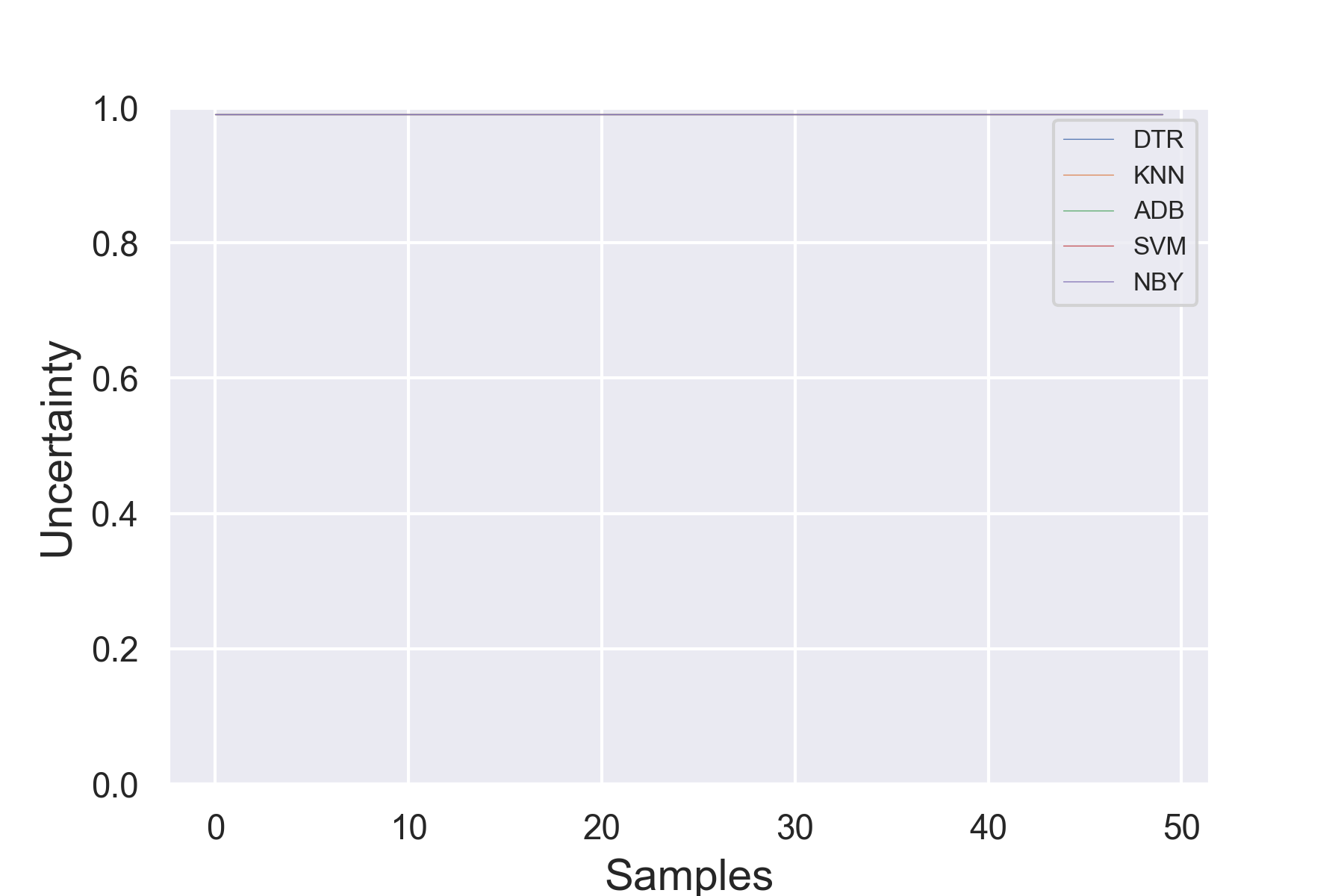}
    	\caption{YAGER UQ for ML models}\label{fig__UQYAGER__ML}
	\end{subfigure}
			~
	\begin{subfigure}[b]{0.3\textwidth}
    	\includegraphics[width=\textwidth,keepaspectratio]{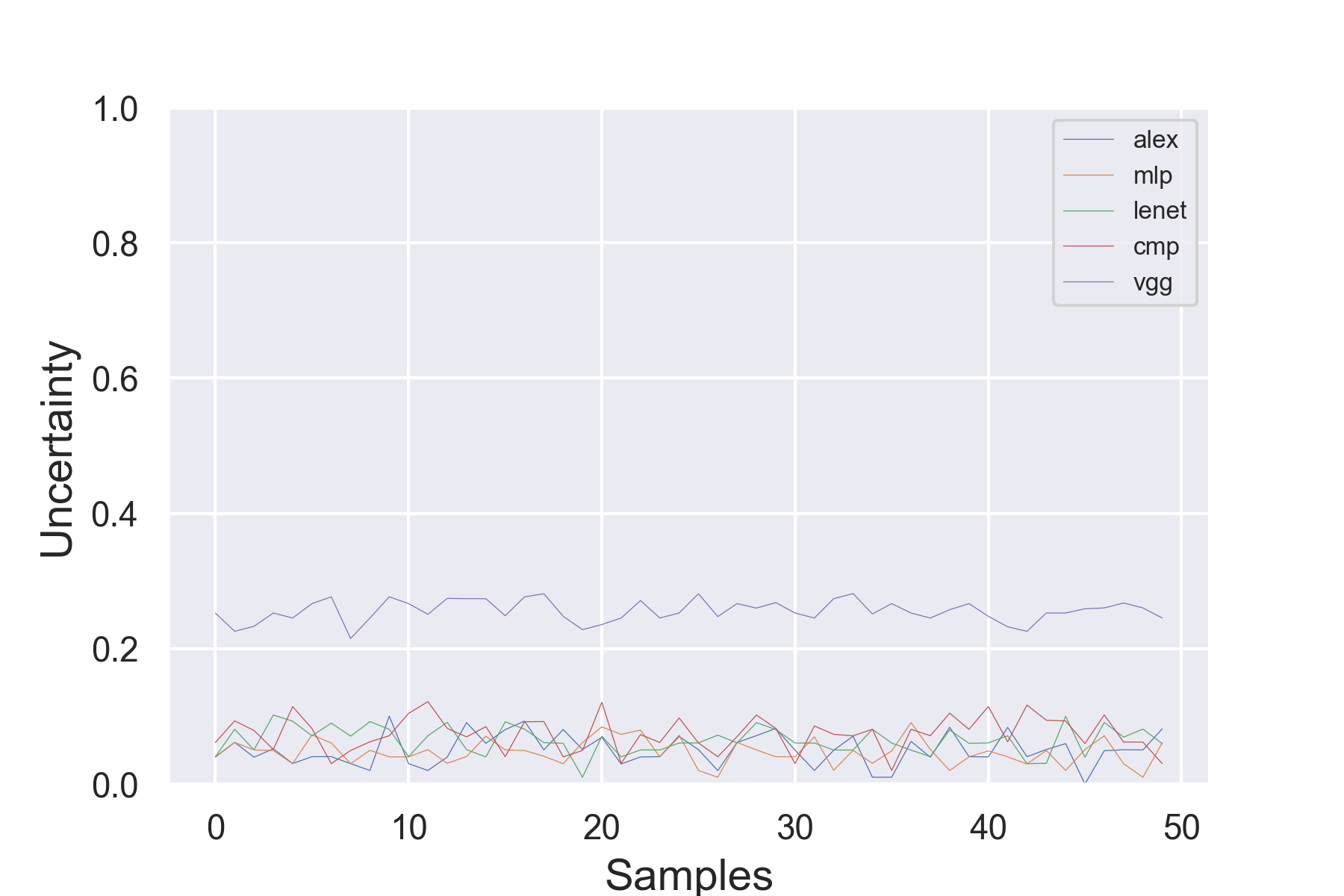}
    	\caption{F1 UQ for NN-based models}\label{fig__UQF1__NN}
	\end{subfigure}
	%~
	\begin{subfigure}[b]{0.3\textwidth}
    	\includegraphics[width=\textwidth,keepaspectratio]{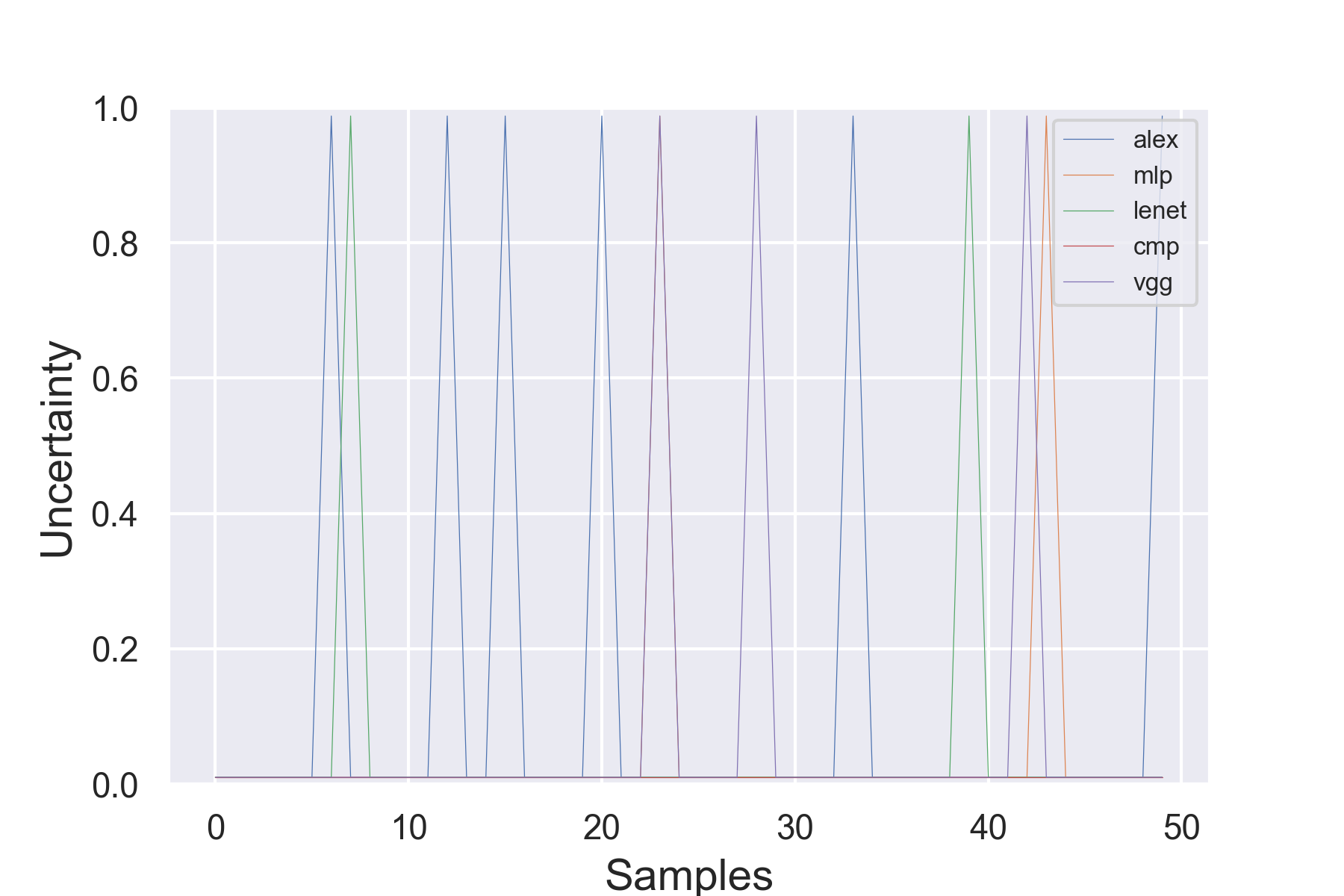}
    	\caption{DSET UQ for NN-based models}\label{fig__UQFDSET__NN}
	\end{subfigure}
	%~
	\begin{subfigure}[b]{0.3\textwidth}
    	\centering
    	\includegraphics[width=\textwidth,keepaspectratio]{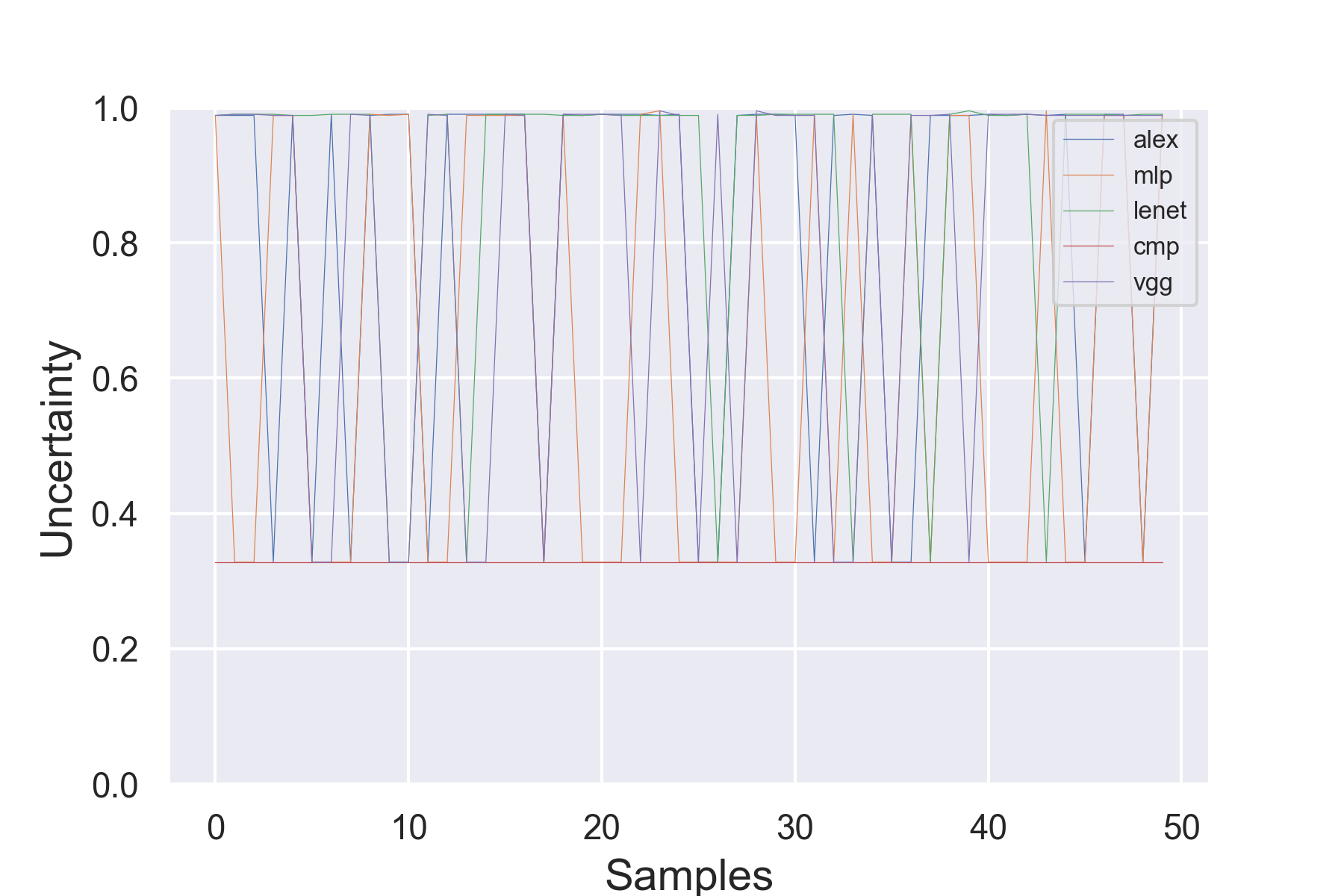}
    	\caption{YAGER UQ for NN-based models}\label{fig__UQYAGER__NN}
	\end{subfigure}
	\caption{Uncertainty quantification of multiclass classifiers using the cases (0, 1, 2, 6, and 12)}: ML-based classifiers (a)-(c), and NN-based classifiers (d)-(f).\label{figure__UQ__classifiers__ML}
\end{figure*}

We quantify the uncertainty after the training of the models using $UQ_{P}$, $UQ_{DS}$, and $UQ_{Y}$ as detailed in section \ref{approach__uncert_quantif}. The uncertainty quantification (UQ) is represented as plots for the ML-based and NN-based classifiers. For illustration purposes, we present only the plots for the multiclass classifiers using the easy dataset.   
Figures \ref{fig__UQF1__ML}, \ref{fig__UQDSET__ML}, and \ref{fig__UQYAGER__ML} present the UQ for the ML-based classifiers, namely KNN, SVM, NBY, DTR, and ADB. It is important to note that the desired value for $UQ_{P}$, $UQ_{DS}$, and $UQ_{Y}$ is zero, which implies consistent and accurate predictions. In the case of $UQ_{DS}$ and $UQ_{Y}$, zero value means no conflicting evidence.
A random validation batch of 20 samples is used to quantify the uncertainty. These samples are extracted from the validation data. This operation is performed 50 times. 
The ML-based classifiers show comparable results using $UQ_{DS}$ and $UQ_{F1}$ with values tending to zero (with the exception of some samples in the case of $UQ_{DS}$ that take the value of one). 
In contrast, $UQ_{Y}$ has a stable approximated value of one, which implies constant conflicting evidence in the predictions of the random batch.

Figures \ref{fig__UQF1__NN}, \ref{fig__UQFDSET__NN}, and \ref{fig__UQYAGER__NN} present the UQ for the NN-based classifiers, namely ale, len, vgg, mlp, and cmp.
Likewise, the NN-based ones show comparable results using $UQ_{DS}$ and $UQ_{F1}$ with values tending to zero for the classifiers alex, mlp, lenet, and cmp (except for some samples in the case of $UQ_{DS}$ that takes the value of one). In the case of vgg, the $UQ_{F1}$ displays values closer to 0.2. 
In contrast, $UQ_{Y}$ shows mixed results for the NN-based classifiers. The classifier cmp has values closer to 0.3. Whereas, the rest of the classifiers have fluctuating values in the range of 0.3 and 1. This fluctuating behavior suggests constant conflicting evidence, which implies changes in the predictions while applying the random batch. 

\subsubsection{Pool Selection}

We first construct a baseline with the performance of the individual classifiers. In this manner, it is possible to compare the performance of the new ECs. As stated in the experiment design, we use two datasets: easy (0,1,2,6,12) and hard (0,3,9,15,21). In addition, we create binary and multiclass classifiers for the two datasets. For illustration purposes, we present the performance of each (binary and multiclass) classifier (ML-based and NN-based) using the easy dataset. Each table summarizes the performance (F1-score) per fault case, an average F1-score, and the computation time for training the (individual) classifiers of the pool. For illustration purposes, we present only the results of the ECs using the easy dataset. 

Table \ref{table__results__classification__MC__individual__ML__NN} shows the performance (F1-score) and computation time for the individual multiclass (ML-based and NN-based) classifiers using the fault cases (0,1,2,6,12)). KNN presents the highest average F1-score with a value of 0.94 from the ML classifiers. In contrast, ale and mlp present the better average F1-score with values of 0.95 and 0.93 from the NN-based classifiers, respectively.        
\begin{table*}[!ht]
\centering
\caption{Classification results of multiclass classifiers using the cases (0,1,2,6,12), and F1-score.}
\begin{tabular}{c|ccccc|ccccc}
\hline
\multirow{2}{*}{\textbf{Fault}} & \multicolumn{5}{c|}{\textbf{ML Classifiers}} & \multicolumn{5}{c}{\textbf{NN-based Classifiers}} \\ \cline{2-11} 
 & \textbf{NBY} & \textbf{ADB} & \textbf{DTR} & \textbf{SVM} & \textbf{KNN} & \textbf{ale} & \textbf{len} & \textbf{vgg} & \textbf{cmp} & \textbf{mlp} \\ \hline
0 & 0.00 & 0.70 & 0.75 & 0.94 & 0.92 & 0.95 & 0.84 & 0.81 & 0.67 & 0.88 \\
1 & 0.99 & 0.97 & 0.92 & 0.99 & 0.98 & 0.98 & 0.95 & 0.96 & 0.99 & 0.99 \\
2 & 0.98 & 0.98 & 0.92 & 0.99 & 0.98 & 0.96 & 0.98 & 0.99 & 0.97 & 0.97 \\
6 & 0.97 & 0.99 & 1.00 & 0.55 & 0.99 & 0.99 & 0.97 & 0.98 & 0.85 & 0.98 \\
12 & 0.48 & 0.57 & 0.73 & 0.71 & 0.84 & 0.85 & 0.72 & 0.09 & 0.74 & 0.82 \\ \hline
\multicolumn{1}{r|}{Avg F1-score} & 0.68 & 0.84 & 0.86 & 0.83 & 0.94 & 0.95 & 0.89 & 0.77 & 0.84 & 0.93 \\
\multicolumn{1}{r|}{Relative time {[}\%{]}} & 16.8 & 100 & 11.9 & 14.2 & 21.6 & 71.3 & 22.1 & 84.9 & 24.1 & 17.9\\ \hline
\end{tabular}
\label{table__results__classification__MC__individual__ML__NN}
\end{table*}
Table \ref{table__results__classification__BIN__individual__ML__NN} shows the performance and computation time for the individual binary classifiers using the normal condition (0) and one of the fault cases (1,2,6,12). For illustration purposes, the normal condition of the table is the average of the results of the normal condition of all the binary classifiers. The best average F1-score corresponds to NBY and SVM with a value of 0.98 for the ML-based classifiers, whereas ale and cmp have the highest average F1-score with values of 0.98 and 0.97 for the NN-based models, respectively. 

\begin{table*}[!ht]
\centering
\caption{Classification results of individual binary classifiers using the normal condition (0) and the faults (1,2,6,12), and F1-score.}
\begin{tabular}{c|ccccc|ccccc}
\hline
\multirow{2}{*}{\textbf{Fault}} & \multicolumn{5}{c|}{\textbf{ML Classifiers}} & \multicolumn{5}{c}{\textbf{NN-based Classifiers}} \\ \cline{2-11} 
 & \textbf{NBY} & \textbf{ADB} & \textbf{DTR} & \textbf{SVM} & \textbf{KNN} & \textbf{ale} & \textbf{len} & \textbf{vgg} & \textbf{cmp} & \textbf{mlp} \\ \hline
0 & 0.99 & 0.93 & 0.91 & 0.98 & 0.98 & 0.99 & 0.96 & 0.90 & 0.98 & 0.97 \\
1 & 0.96 & 0.90 & 0.93 & 0.99 & 0.99 & 0.99 & 0.99 & 0.99 & 1.00 & 0.99 \\
2 & 0.99 & 0.93 & 0.84 & 0.95 & 0.98 & 0.98 & 0.95 & 0.98 & 0.97 & 0.94 \\
6 & 1.00 & 1.00 & 1.00 & 1.00 & 0.99 & 1.00 & 1.00 & 0.99 & 0.99 & 1.00 \\
12 & 0.98 & 0.82 & 0.82 & 0.96 & 0.87 & 0.94 & 0.83 & 0.73 & 0.93 & 0.90 \\ \hline
\multicolumn{1}{r|}{Avg F1-score} & 0.98 & 0.91 & 0.90 & 0.98 & 0.96 & 0.98 & 0.94 & 0.92 & 0.97 & 0.96 \\
%\multicolumn{1}{r|}{End time {[}s{]}} & N/A & N/A & N/A & N/A & N/A & N/A & N/A & N/A & N/A & N/A \\ 
\hline
\end{tabular}
\label{table__results__classification__BIN__individual__ML__NN}
\end{table*}

The next step is the creation of the ECs: ML-based, NN-based, and Hybrid. After applying the procedure described in section \ref{approach__pool_selection}, we obtain 38 ECs. As seen in Table \ref{table__results__pool_selection__selection__classification}, the parameters are expert (Exp), diversity (Div), diversity version (Ver), and pre-cut (P-C). It is important to note that we divide the ECs into three groups: ML (M2..M5), DL (D2..D5), and Hybrid (H2-1..H10-1). Each row of the table is an EC (e.g., M2 EC consists of the ML models KNN and SVM), in which the first letter denotes the nature of the classifier (e.g., M for ML-based, D for NN-based, and H for Hybrid), the first number is the ensemble size. The last number is the consecutive number for this EC.
The ECs H5-1, H6-1, and H6-4 present the highest average F1 score with a value of 0.97 for each EC. The training time shows mixed results varying from 8s for M2 and 1242s for H9-3. The ECs H5-1, H6-1, and H6-4 have a training time of 557s, 476s, and 611s, respectively. The training time is represented as a percentage using a \textit{relative time}. For this purpose, we consider the highest training time of the EC H9-3 with a duration of 1242 seconds (relative time of 100\%) as reference. In contrast, one of the ECs with the smallest training time M2 presents a duration of 8 seconds (relative time of 1\%).
\begin{table*}[!ht]
\centering
\caption{Selected ensemble multiclass classifiers using pool selection and the cases (0,1,2,6,12) and F1-score for classification performance.}
\begin{tabular}{c|cccc|c|ccc|l}
\hline
\multirow{2}{*}{\textbf{E-S}} & \multirow{2}{*}{\textbf{Exp}} & \multirow{2}{*}{\textbf{Div}} & \multirow{2}{*}{\textbf{Ver}} & \multirow{2}{*}{\textbf{P-C}} & \multirow{2}{*}{\textbf{Exper.}} & \multirow{2}{*}{\textbf{F1}} & \multirow{2}{*}{\textbf{Tr. time {[}s{]}}} & \multirow{2}{*}{\textbf{Rel. time}} & \multicolumn{1}{c}{\multirow{2}{*}{\textbf{Pool}}} \\
 &  &  &  &  &  &  &  &  & \multicolumn{1}{c}{} \\ \hline
2 & False & False & False & False & M2 & 0.86 & 8 & 1\% & DTR-KNN \\
3 & False & False & False & False & M3 & 0.86 & 34 & 3\% & DTR-KNN-NBY \\
4 & False & False & False & False & M4 & 0.90 & 35 & 3\% & DTR-KNN-NBY-SVM \\
5 & False & False & False & False & M5 & 0.91 & 39 & 3\% & DTR-KNN-NBY-SVM-ADB \\ \hline
2 & False & False & False & False & D2 & 0.84 & 512 & 41\% & cmp-vgg \\
3 & False & False & False & False & D3 & 0.86 & 518 & 42\% & lenet-cmp-vgg \\
4 & False & False & False & False & D4 & 0.92 & 651 & 52\% & alex-lenet-cmp-vgg \\
5 & False & False & False & False & D5 & 0.96 & 1221 & 98\% & mlp-alex-lenet-cmp-vgg \\ \hline
2 & False & False & False & False & H2-1 & 0.86 & 11 & 1\% & DTR-KNN \\
2 & False & True & False & False & H2-2 & 0.84 & 670 & 54\% & cmp-vgg \\
2 & False & True & False & True & H2-3 & 0.86 & 9 & 1\% & DTR-cmp \\ \hline
3 & False & False & False & False & H3-1 & 0.86 & 35 & 3\% & DTR-KNN-NBY \\
3 & False & True & False & False & H3-2 & 0.86 & 676 & 54\% & lenet-cmp-vgg \\
3 & True & False & False & False & H3-3 & 0.94 & 17 & 1\% & KNN-NBY-lenet \\
3 & True & True & True & True & H3-4 & 0.86 & 41 & 3\% & DTR-NBY-lenet \\ \hline
4 & False & False & False & False & H4-1 & 0.90 & 36 & 3\% & DTR-KNN-NBY-SVM \\
4 & False & True & False & False & H4-2 & 0.92 & 680 & 55\% & alex-lenet-cmp-vgg \\
4 & True & False & False & False & H4-3 & 0.88 & 45 & 4\% & DTR-KNN-NBY-lenet \\
4 & True & True & True & True & H4-4 & 0.88 & 44 & 4\% & DTR-NBY-lenet-vgg \\ \hline
5 & False & False & False & False & H5-1 & \textbf{0.97} & 557 & 45\% & DTR-KNN-NBY-SVM-mlp \\
5 & False & True & False & False & H5-2 & 0.93 & 1126 & 91\% & NBY-alex-ADB-cmp-vgg \\
5 & True & False & False & False & H5-3 & 0.95 & 561 & 45\% & DTR-KNN-NBY-vgg-lenet \\ \hline
6 & False & False & False & False & H6-1 & \textbf{0.97} & 476 & 38\% & DTR-KNN-NBY-SVM-mlp-alex \\
6 & False & True & False & False & H6-2 & 0.91 & 683 & 55\% & NBY-alex-ADB-lenet-cmp-vgg \\
6 & False & True & False & True & H6-3 & 0.95 & 477 & 38\% & DTR-KNN-NBY-mlp-alex-ADB \\
6 & True & False & False & False & H6-4 & \textbf{0.97} & 611 & 49\% & DTR-KNN-NBY-vgg-lenet-alex \\ \hline
7 & False & False & False & False & H7-1 & 0.96 & 477 & 38\% & DTR-KNN-NBY-SVM-mlp-alex-ADB \\
7 & False & True & False & False & H7-2 & 0.95 & 1114 & 90\% & NBY-mlp-alex-ADB-lenet-cmp-vgg \\
7 & False & True & False & True & H7-3 & 0.95 & 483 & 39\% & DTR-KNN-NBY-mlp-alex-ADB-lenet \\
7 & True & False & False & False & H7-4 & 0.96 & 614 & 49\% & DTR-KNN-NBY-vgg-lenet-alex-mlp \\ \hline
8 & False & False & False & False & H8-1 & 0.97 & 483 & 39\% & DTR-KNN-NBY-SVM-mlp-alex-ADB-lenet \\
8 & False & True & False & False & H8-2 & 0.93 & 1115 & 90\% & DTR-NBY-mlp-alex-ADB-lenet-cmp-vgg \\
8 & False & True & False & True & H8-3 & 0.96 & 624 & 50\% & DTR-KNN-NBY-mlp-alex-ADB-lenet-cmp \\
8 & True & False & False & False & H8-4 & 0.97 & 1102 & 89\% & DTR-KNN-NBY-SVM-vgg-lenet-alex-mlp \\ \hline
9 & False & False & False & False & H9-1 & 0.97 & 623 & 50\% & DTR-KNN-NBY-SVM-mlp-alex-ADB-lenet-cmp \\
9 & False & True & False & False & H9-2 & 0.96 & 1115 & 90\% & DTR-KNN-NBY-mlp-alex-ADB-lenet-cmp-vgg \\
9 & True & False & False & False & H9-3 & 0.97 & 1242 & 100\% & DTR-KNN-NBY-SVM-ADB-vgg-lenet-alex-mlp \\ \hline
10 & False & False & False & False & H10-1 & 0.97 & 1114 & 90\% & DTR-KNN-NBY-SVM-mlp-alex-ADB-lenet-cmp-vgg \\ \hline
\end{tabular}
\label{table__results__pool_selection__selection__classification}
\end{table*}

\subsubsection{Ensemble Classification Performance}\label{use_case__EC}
This subsection presents the classification performance of the ECs (ML-based, NN-based, and Hybrid) in the form of tables and plots. The plots represent the performance of selected ECs, which can be visualized through a confusion matrix, a plot of prediction versus ground-truth, a plot of the DSET uncertainty, and a plot of the Yager uncertainty of an EC. Likewise the previous section, the tables summarize the individual F1-score per fault case, an average F1-score, and the relative time for each EC. The inference time is the time required to process all the testing samples, afterwards the relative time is calculated.

Table \ref{table__results__inference__classification__MC__1} presents the multiclass ECs (H5-1, H6-1, and H6-4) and the individual multiclass classifiers (KNN, ale, and mlp) with the best performance from Table \ref{table__results__pool_selection__selection__classification} while using the cases (0,1,2,6,12). 
The ECs H5-1, H6-1, and H6-4 present comparable results with an average F1-score of 0.97, whereas the individual classifiers KNN, ale, and mlp have values of 0.95, 0.94, and 0.94, respectively. It is important to highlight the relative time difference between H6-1 with 38.3\% and KNN with 0.5\%.%, which can also be expressed as 0.0992 sample/s and 0.0012 sample/s, respectively.}

\begin{table}[!ht]
\centering
\caption{Inference results of selected ensemble multiclass classifiers using the cases (0,1,2,6,12), and F1-score.}
\begin{tabular}{c|ccc|ccc}
\hline
\multirow{2}{*}{\textbf{Fault}} & \multicolumn{3}{c|}{\textbf{MC   EC}} & \multicolumn{3}{c}{\textbf{INDIV}} \\ \cline{2-7} 
 & \textbf{H5-1} & \textbf{H6-1} & \textbf{H6-4} & \textbf{KNN} & \textbf{ale} & \textbf{mlp} \\ \hline
0 & 0.96 & \textbf{0.96} & 0.96 & 0.92 & 0.95 & 0.88 \\
1 & 0.99 & \textbf{0.99} & 0.99 & 0.98 & 0.98 & 0.99 \\
2 & 0.98 & \textbf{0.99} & 0.99 & \textbf{0.98} & 0.96 & 0.97 \\
6 & 0.99 & 0.99 & 0.99 & 0.99 & 0.99 & 0.98 \\
12 & 0.92 & \textbf{0.93} & 0.92 & \textbf{0.84} & 0.85 & 0.82 \\ \hline
\multicolumn{1}{r|}{Avg F1-score} & 0.97 & \textbf{0.97} & 0.97 & 0.95 & 0.94 & 0.94 \\
\multicolumn{1}{r|}{Rel. time {[}\%{]}} & 44.9 & \textbf{38.3} & 49.3 & 0.5 & 1.6 & 0.4 \\ \hline
\end{tabular}
\label{table__results__inference__classification__MC__1}
\end{table}

Table \ref{table__results__inference__classification__BIN__1} presents the binary ECs (M2, M5, and H6-2) and the individual binary classifiers (NBY, SVM, and ale) while using the normal condition (0) and the fault cases (1,2,6,12). Similarly to Table \ref{table__results__inference__classification__MC__1}, the rows represent the performance of a binary EC, and the normal condition is the average of all the binary classifiers of the EC. 
The ECs M2, M5, and H6-2 present comparable results with an average F1-score of 0.98, 0.99, and 0.99, respectively, whereas the individual classifiers NBY, SVM, and ale have a value of 0.99 each. Moreover, the results of the binary ECs are comparable to those of the multiclass ECs of Table \ref{table__results__inference__classification__MC__1}. Every binary EC of a fault case has its correspondent computing time; therefore, it is inadequate for comparison purposes. 
\begin{table}[!ht]
\centering
\caption{Inference results of selected ensemble binary classifiers using the normal condition (0) and the cases (1,2,6,12), and F1-score.}
\begin{tabular}{c|ccc|ccc}
\hline
\multirow{2}{*}{\textbf{Fault}} & \multicolumn{3}{c|}{\textbf{BIN   EC}} & \multicolumn{3}{c}{\textbf{INDIV}} \\ \cline{2-7} 
 & \textbf{M2} & \textbf{M5} & \textbf{H6-2} & \textbf{NBY} & \textbf{SVM} & \textbf{ale} \\ \hline
0 & 0.98 & 0.99 & 0.99 & 0.99 & 0.98 & 0.99 \\
1 & 0.99 & \textbf{1.00} & 1.00 & 0.96 & 0.99 & 0.99 \\
2 & 0.98 & 0.98 & 0.98 & \textbf{0.99} & 0.95 & 0.98 \\
6 & 1.00 & 1.00 & 1.00 & 1.00 & 1.00 & 1.00 \\
12 & 0.94 & 0.96 & 0.97 & \textbf{0.98} & 0.96 & 0.94 \\ \hline
\multicolumn{1}{r|}{Avg F1-score} & 0.98 & \textbf{0.99} & 0.99 & 0.98 & 0.98 & 0.98 \\
%\multicolumn{1}{r|}{End time {[}s{]}} & N/A & N/A & N/A & N/A & N/A & N/A \\ 
\hline
\end{tabular}
\label{table__results__inference__classification__BIN__1}
\end{table}

Table \ref{table__results__inference__classification__MC__2} presents the multiclass ECs (M5, H4-1, and H4-3) and the individual multiclass classifiers (NBY, SVM, and KNN) with the best performance from Table \ref{table__results__pool_selection__selection__classification} while using the cases (0,3,9,15,21). 
The ECs M5, H4-1, and H4-3 present comparable results with an average F1-score of 0.22, 0.23, and 0.23, respectively, whereas the individual classifiers NBY, SVM, and KNN have values of 0.5, 0.58, and 0.58, respectively. It is important to highlight the relative time difference between H4-1 with 1.1\% and KNN with 0.3\%.%, which can also be expressed as 0.0025 sample/s and 0.000625 sample/s, respectively.} 
\begin{table}[!ht]
\centering
\caption{Inference results of selected ensemble multiclass classifiers using the cases (0,3,9,15,21), and F1-score.}
\begin{tabular}{c|ccc|ccc}
\hline
\multirow{2}{*}{\textbf{Fault}} & \multicolumn{3}{c|}{\textbf{MC EC}} & \multicolumn{3}{c}{\textbf{INDIV}} \\ \cline{2-7} 
 & \textbf{M5} & \textbf{H4-1} & \textbf{H4-3} & \textbf{NBY} & \textbf{SVM} & \textbf{KNN} \\ \hline
0 & 0.30 & \textbf{0.39} & 0.39 & 0.66 & \textbf{0.70} & 0.67 \\
3 & 0.19 & \textbf{0.17} & 0.17 & \textbf{0.52} & 0.44 & 0.48 \\
9 & 0.17 & \textbf{0.18} & 0.18 & \textbf{0.47} & 0.45 & 0.43 \\
15 & 0.23 & 0.20 & \textbf{0.20} & \textbf{0.49} & 0.46 & 0.44 \\
21 & 0.23 & \textbf{0.20} & 0.20 & \textbf{0.52} & \textbf{0.98} & 0.98 \\ \hline
\multicolumn{1}{r|}{Avg F1-score} & 0.22 & \textbf{0.23} & \textbf{0.23} & 0.50 & \textbf{0.58} & 0.58 \\
\multicolumn{1}{r|}{Rel. time {[}\%{]}} & 2.1 & \textbf{1.1} & \textbf{1.0} & 0.2 & 2 & 0.3 \\ 
\hline
\end{tabular}
\label{table__results__inference__classification__MC__2}
\end{table}

Table \ref{table__results__inference__classification__BIN__2} presents the binary ECs (M5, H3-3, and H5-1) and the individual binary classifiers (NBY, ADB, and DTR) while using the normal condition (0) and the fault cases (3,9,15,21).  
The ECs M5, H3-3, and H5-1 present comparable results with an average F1-score of 0.63, 0.62, and 0.63, respectively, whereas the individual classifiers NBY, ADB, and DTR have values of 0.5, 0.58, and 0.58, respectively. Moreover, the results of the binary ECs are superior with respect to the results of the multiclass ECs of Table \ref{table__results__inference__classification__MC__2} (e.g., the average F1-score of H4-1 is 0.23, whereas the H5-1 has a value of 0.63). 
\begin{table}[!ht]
\centering
\caption{Inference results of selected ensemble binary classifiers using the normal condition (0) and the cases (3,9,15,21), and F1-score.}
\begin{tabular}{c|ccc|ccc}
\hline
\multirow{2}{*}{\textbf{Fault}} & \multicolumn{3}{c|}{\textbf{BIN   EC}} & \multicolumn{3}{c}{\textbf{INDIV}} \\ \cline{2-7} 
 & \textbf{M5} & \textbf{H3-3} & \textbf{H5-1} & \textbf{NBY} & \textbf{ADB} & \textbf{DTR} \\ \hline
0 & 0.70 & \textbf{0.67} & 0.68 & 0.66 & \textbf{0.70} & 0.67 \\
3 & 0.49 & \textbf{0.49} & 0.50 & \textbf{0.52} & 0.44 & 0.48 \\
9 & 0.47 & \textbf{0.47} & 0.47 & \textbf{0.47} & 0.45 & 0.43 \\
15 & 0.50 & 0.49 & \textbf{0.53} & \textbf{0.49} & 0.46 & 0.44 \\
21 & 0.98 & \textbf{0.98} & 0.98 & \textbf{0.52} & \textbf{0.98} & 0.98 \\ \hline
\multicolumn{1}{r|}{Avg F1-score} & 0.63 & \textbf{0.62} & \textbf{0.63} & 0.50 & \textbf{0.58} & 0.58 \\
%\multicolumn{1}{r|}{End time {[}s{]}} & N/A & N/A & N/A & N/A & N/A & N/A \\ 
\hline
\end{tabular}
\label{table__results__inference__classification__BIN__2}
\end{table}

\begin{figure*}[!ht]
	\centering
	\begin{subfigure}[b]{0.3\textwidth}
	\includegraphics[width=\textwidth,keepaspectratio]{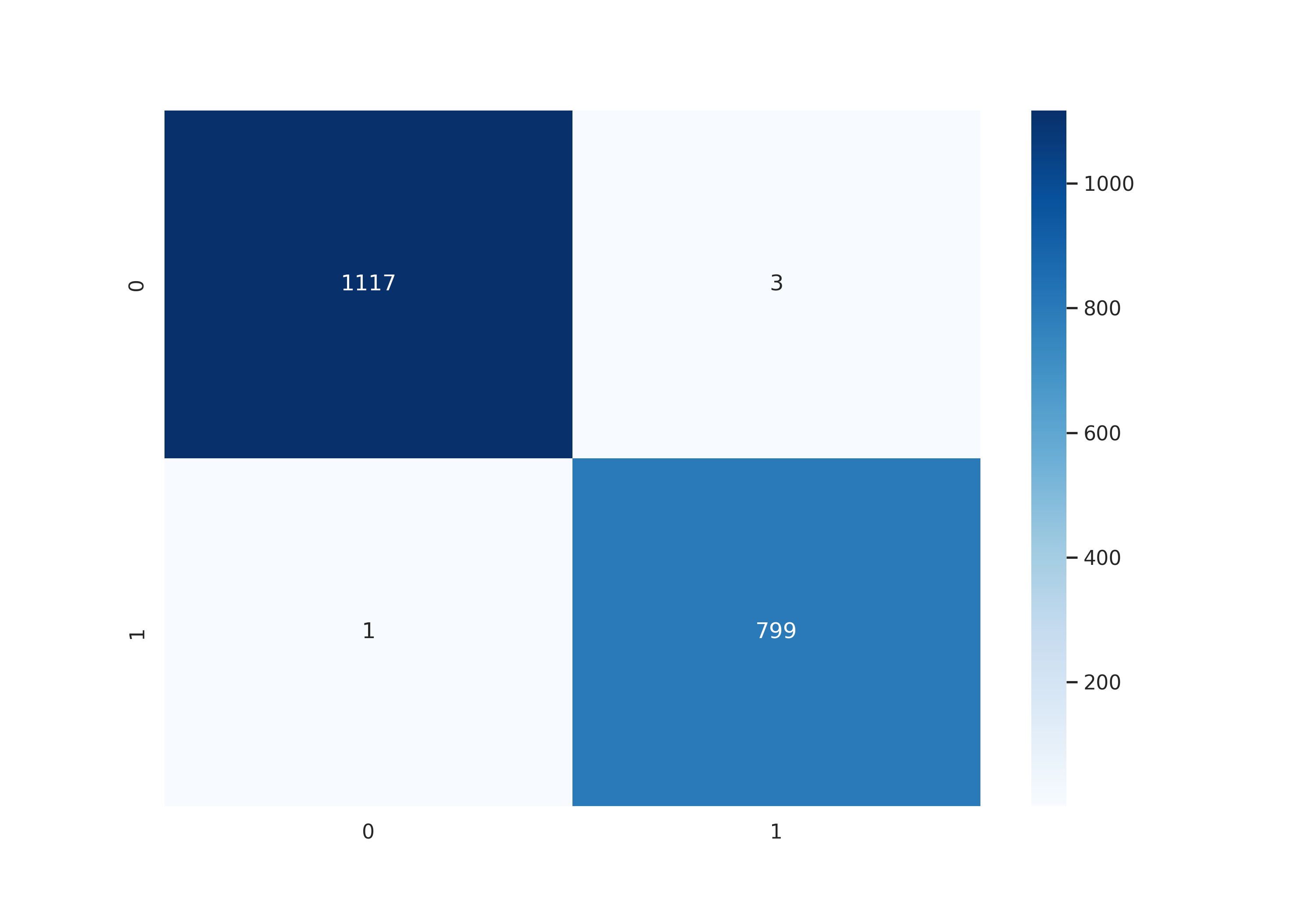}
	\caption{CM for M5}\label{figure__use_cm__M5}
	\end{subfigure}
	\begin{subfigure}[b]{0.3\textwidth}
	\includegraphics[width=\textwidth,keepaspectratio]{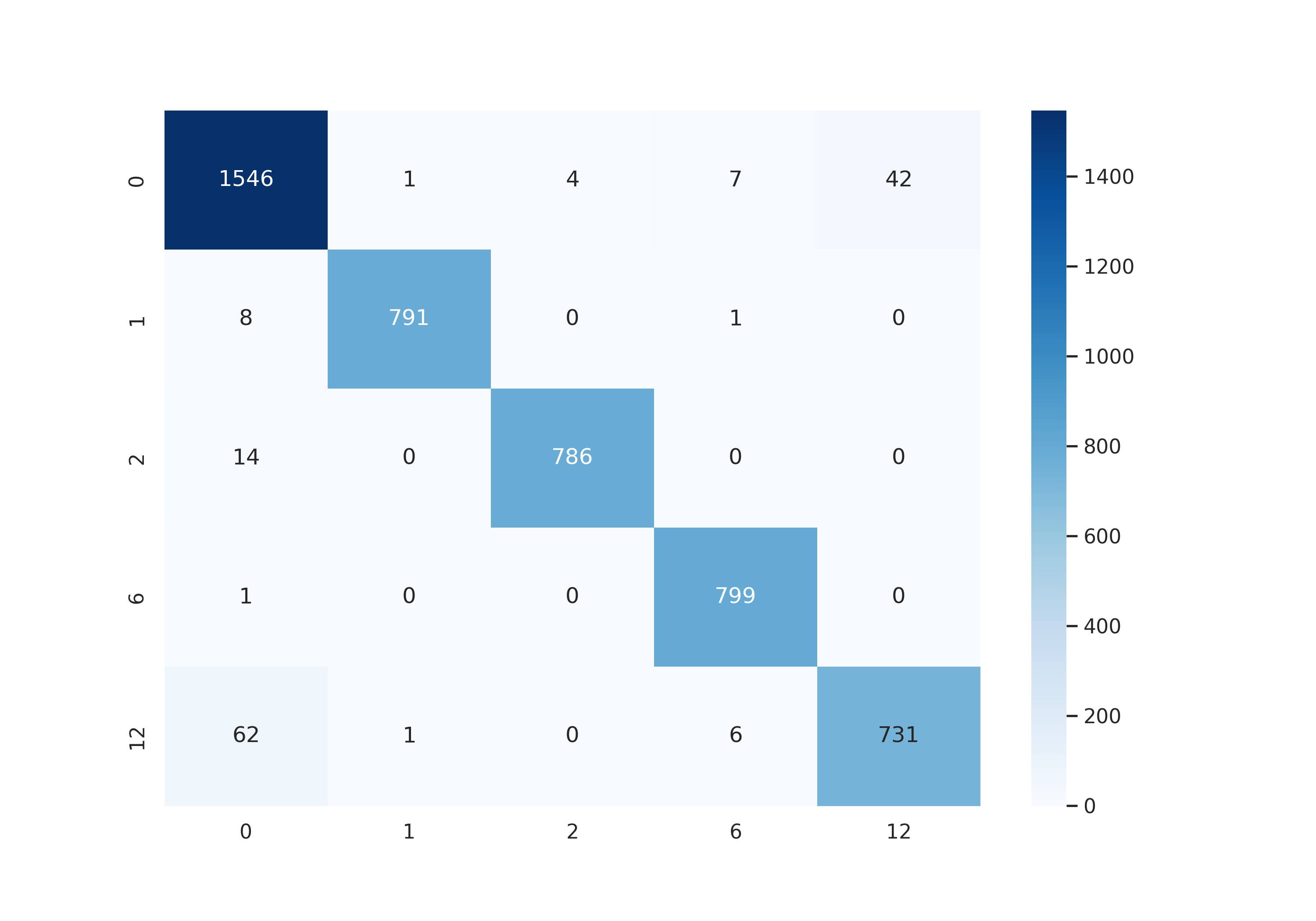}
	\caption{CM for H6-1}\label{figure__use_cm__H6_1}
	\end{subfigure}
	\begin{subfigure}[b]{0.3\textwidth}
	\includegraphics[width=\textwidth,keepaspectratio]{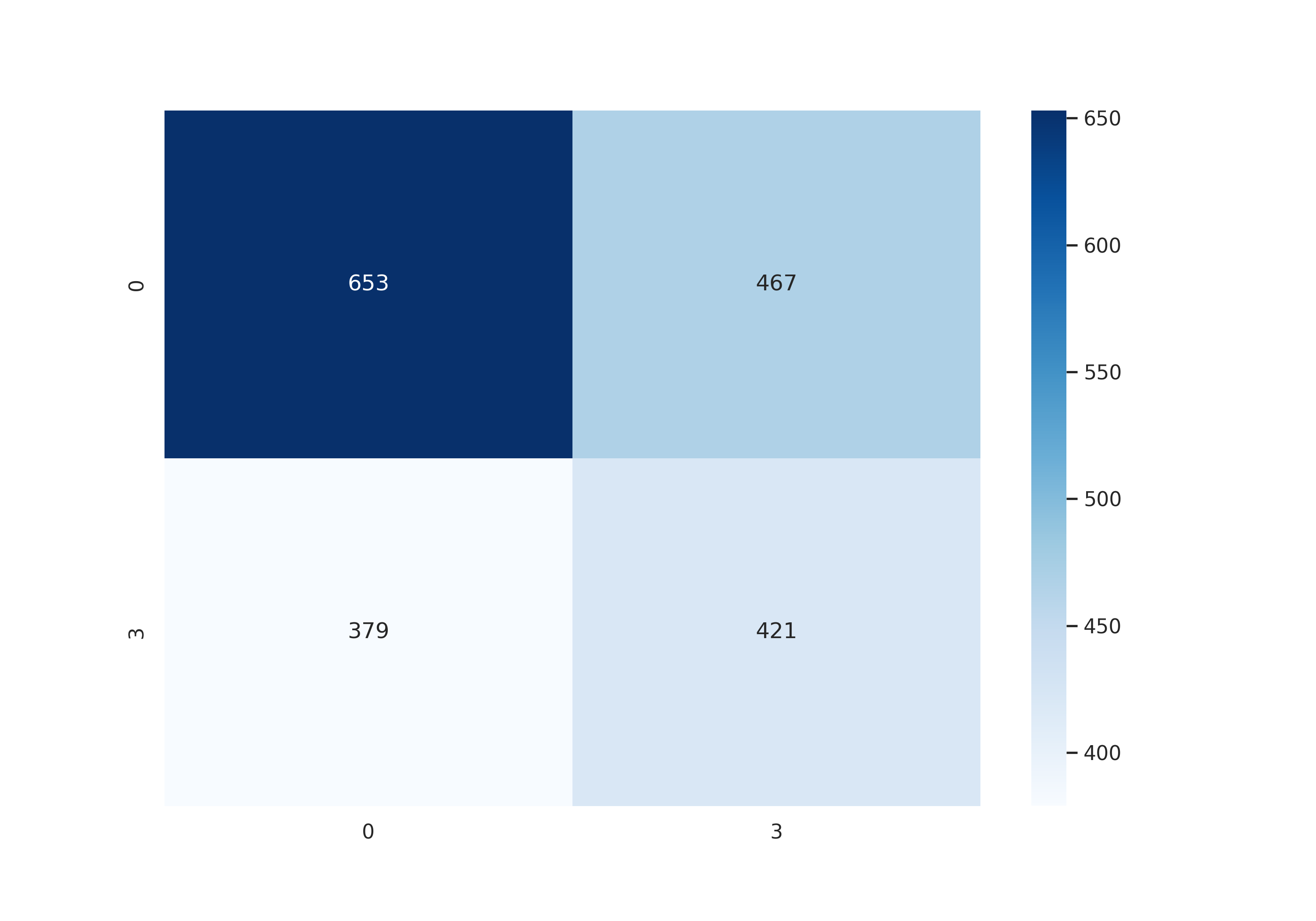}
	\caption{CM for H5-1}\label{figure__use_cm__H5_1}
	\end{subfigure}
	~
	\begin{subfigure}[b]{0.3\textwidth}
	\includegraphics[width=\textwidth,keepaspectratio]{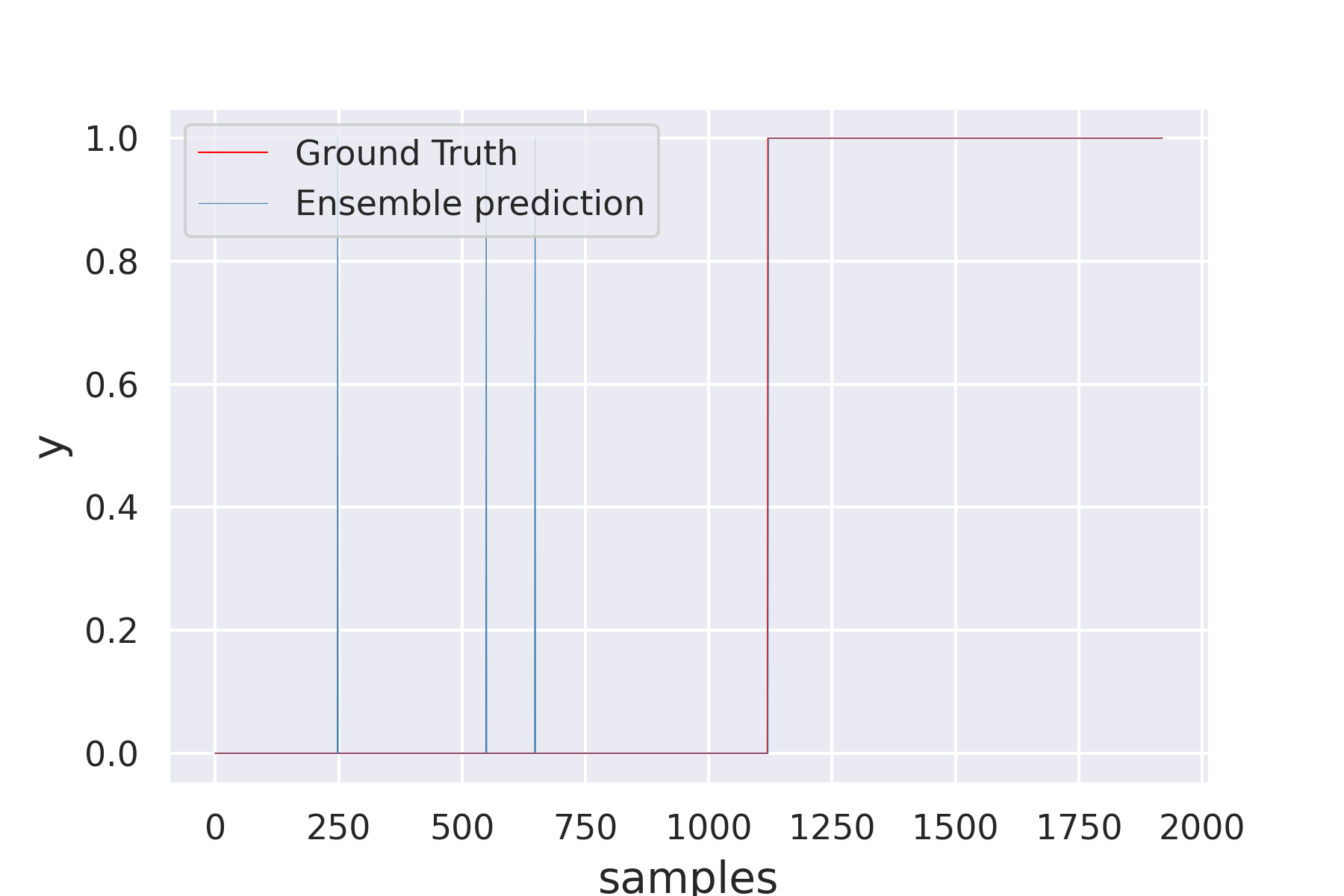}
	\caption{Predictions for M5}\label{figure__use_classification__M5}
	\end{subfigure}
	\begin{subfigure}[b]{0.3\textwidth}
	\includegraphics[width=\textwidth,keepaspectratio]{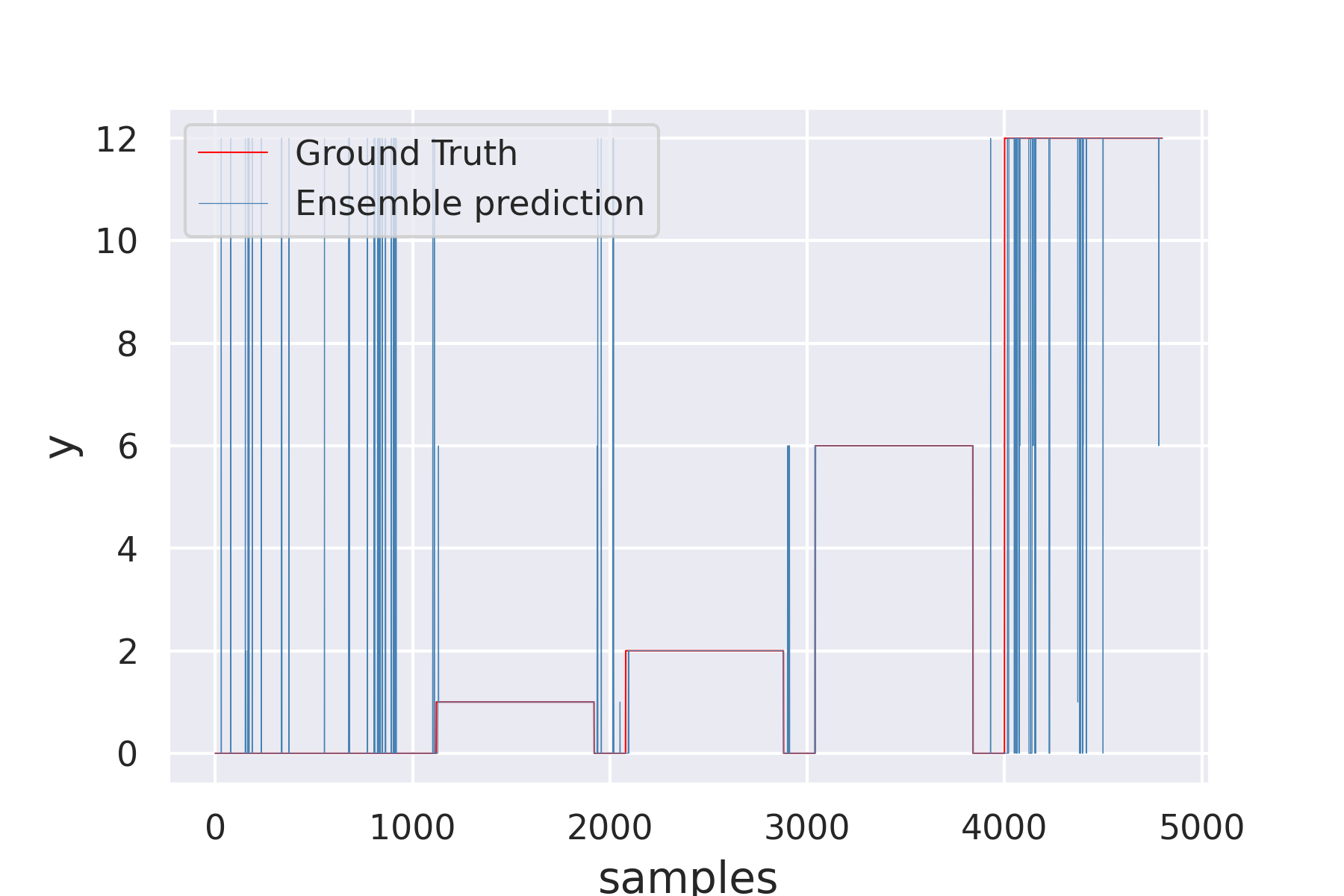}
	\caption{Predictions for H6-1}\label{figure__use_classification__H6_1}
	\end{subfigure}
	\begin{subfigure}[b]{0.3\textwidth}
	\includegraphics[width=\textwidth,keepaspectratio]{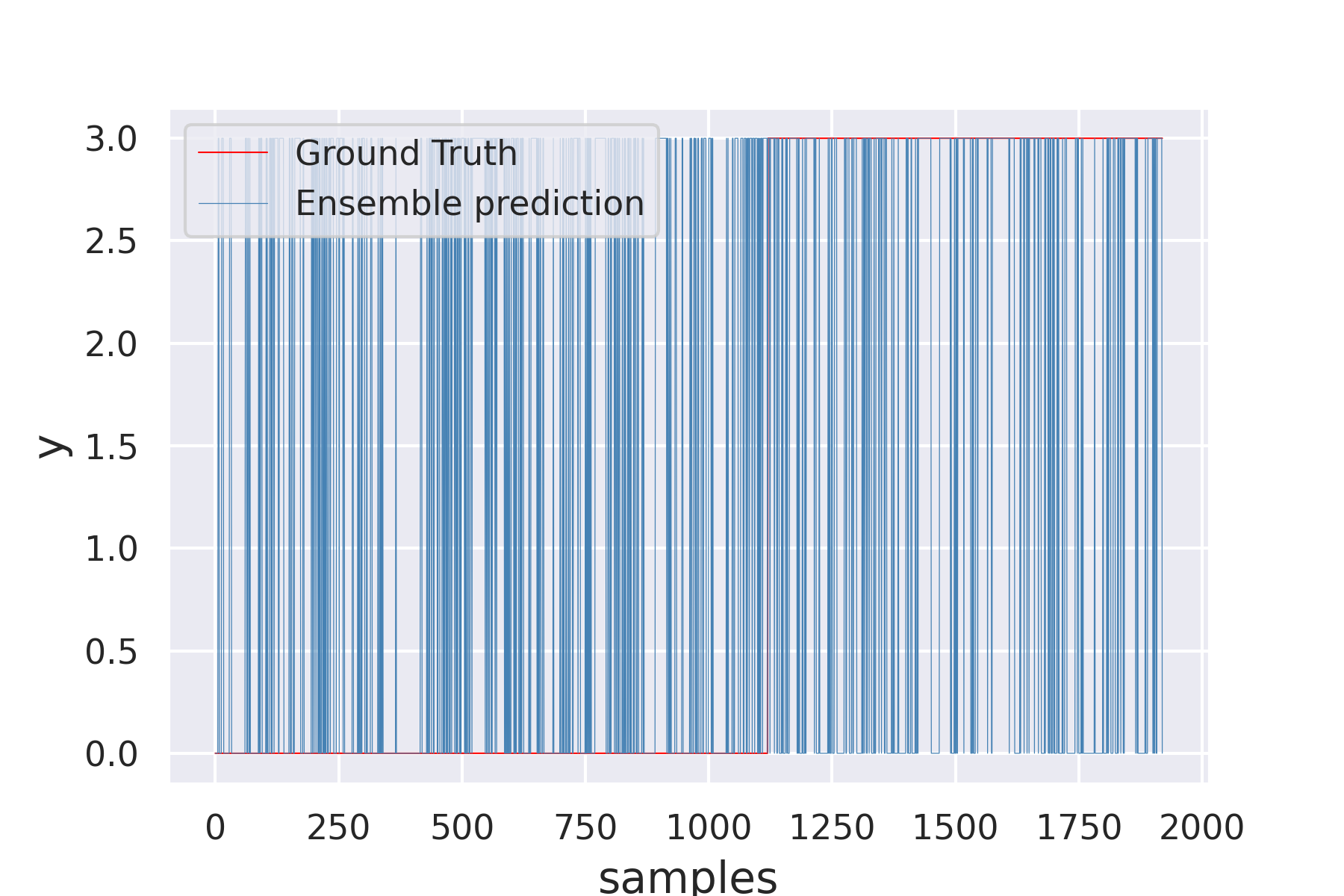}
	\caption{Predictions for H5-1}\label{figure__use_classification__H5_1}
	\end{subfigure}
	~
	\begin{subfigure}[b]{0.3\textwidth}
	\includegraphics[width=\textwidth,keepaspectratio]{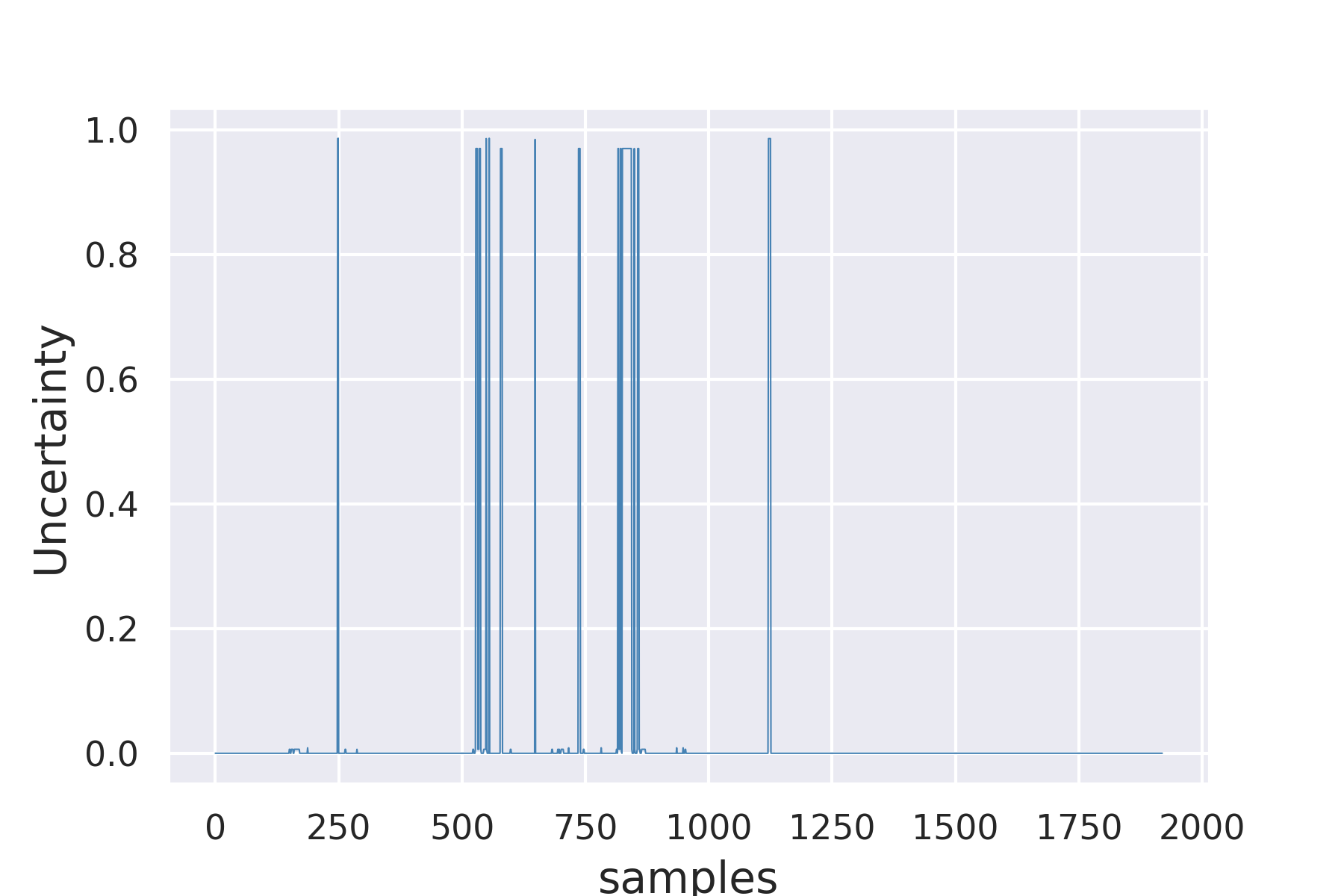}
	\caption{DSET UQ for M5}\label{figure__use_UQDS__M5}
	\end{subfigure}
	\begin{subfigure}[b]{0.3\textwidth}
	\includegraphics[width=\textwidth,keepaspectratio]{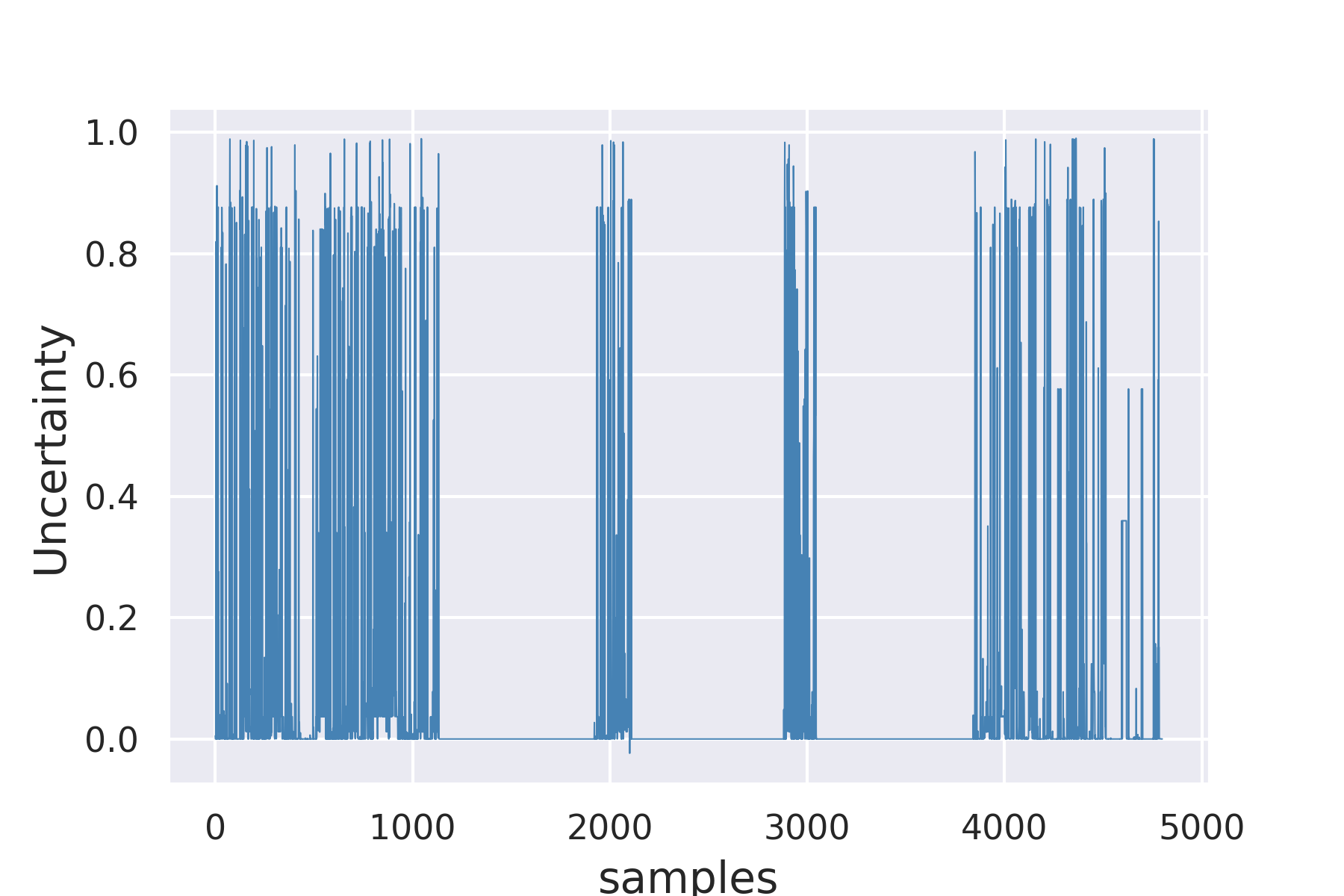}
	\caption{DSET UQ for H6-1}\label{figure__use_UQDS__H6_1}
	\end{subfigure}
	\begin{subfigure}[b]{0.3\textwidth}
	\includegraphics[width=\textwidth,keepaspectratio]{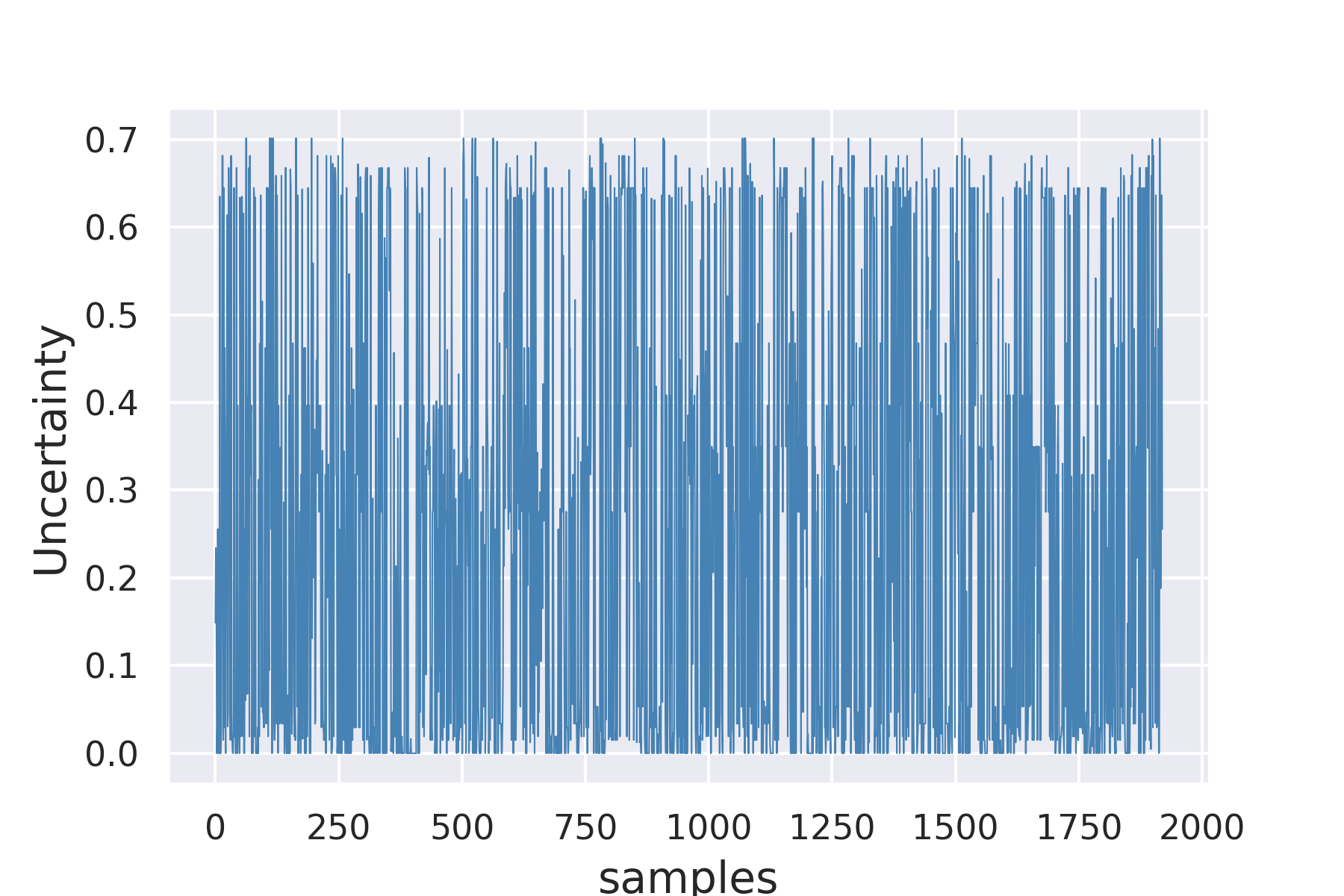}
	\caption{DSET UQ for H5-1}\label{figure__use_UQDS__H5_1}
	\end{subfigure}
    ~
    \begin{subfigure}[b]{0.3\textwidth}
	\includegraphics[width=\textwidth,keepaspectratio]{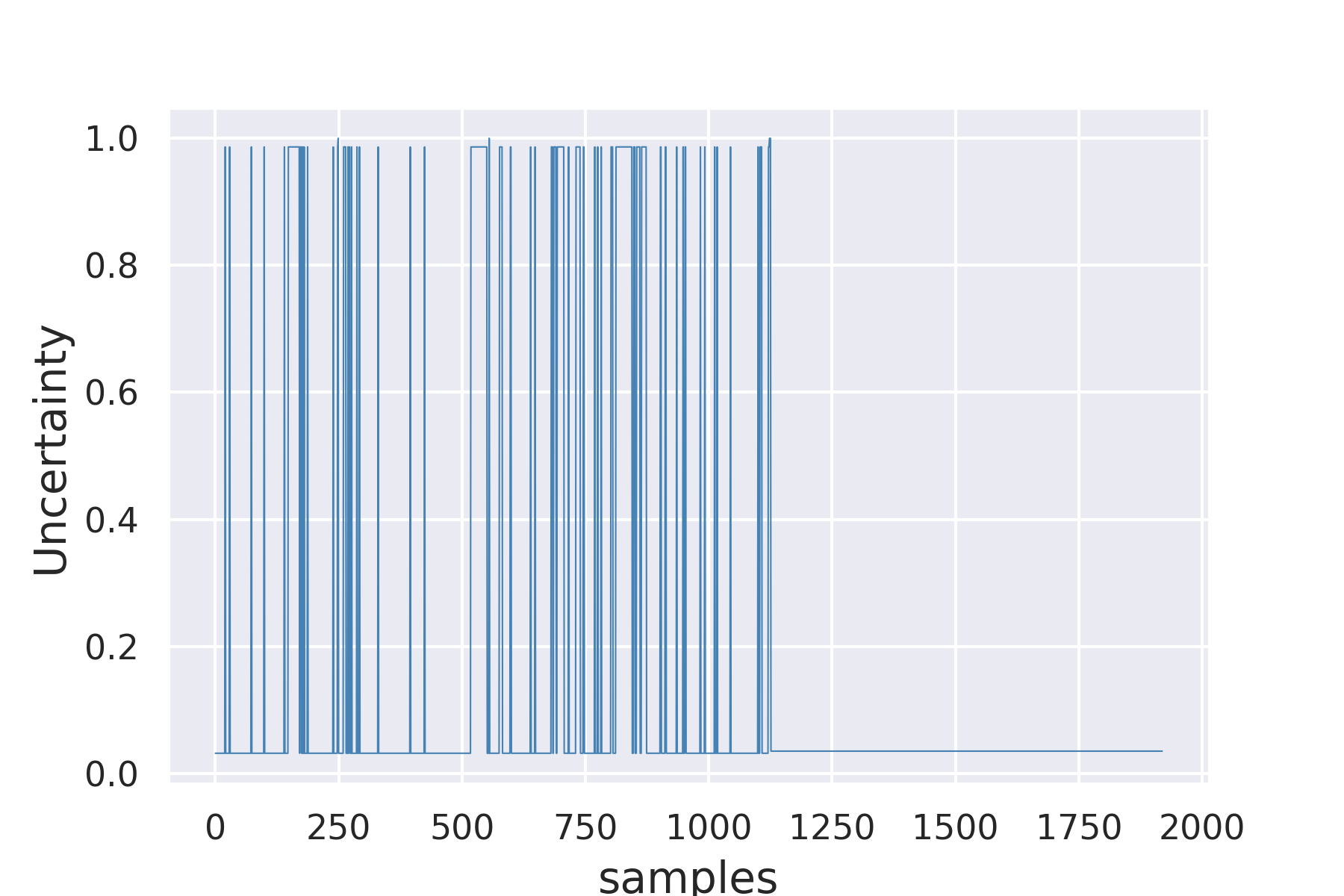}
	\caption{YAGER UQ for M5}\label{figure__use_UQY__M5}
	\end{subfigure}
	\begin{subfigure}[b]{0.3\textwidth}
	\includegraphics[width=\textwidth,keepaspectratio]{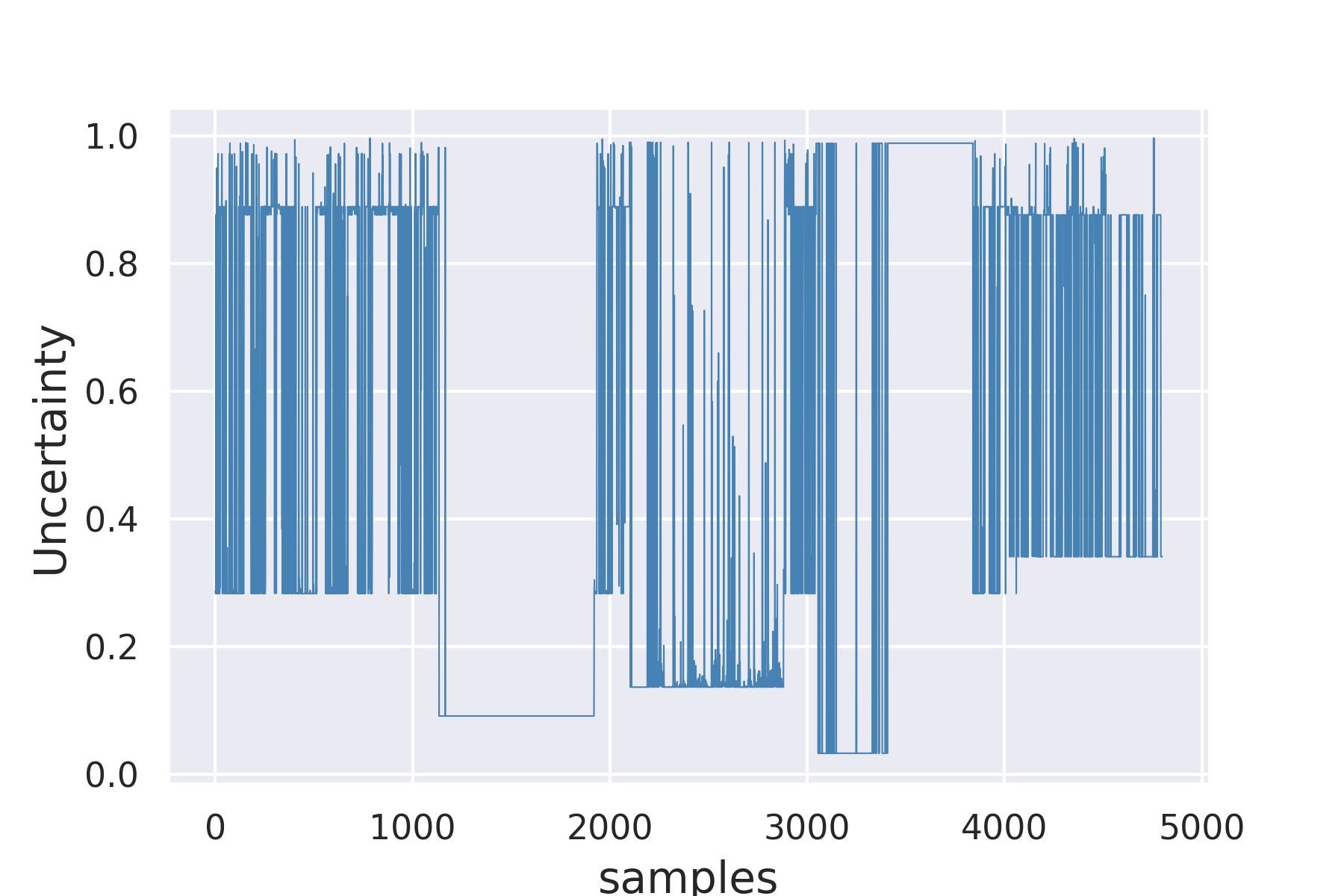}
	\caption{YAGER UQ for H6-1}\label{figure__use_UQY__H6_1}
	\end{subfigure}
	\begin{subfigure}[b]{0.3\textwidth}
	\includegraphics[width=\textwidth,keepaspectratio]{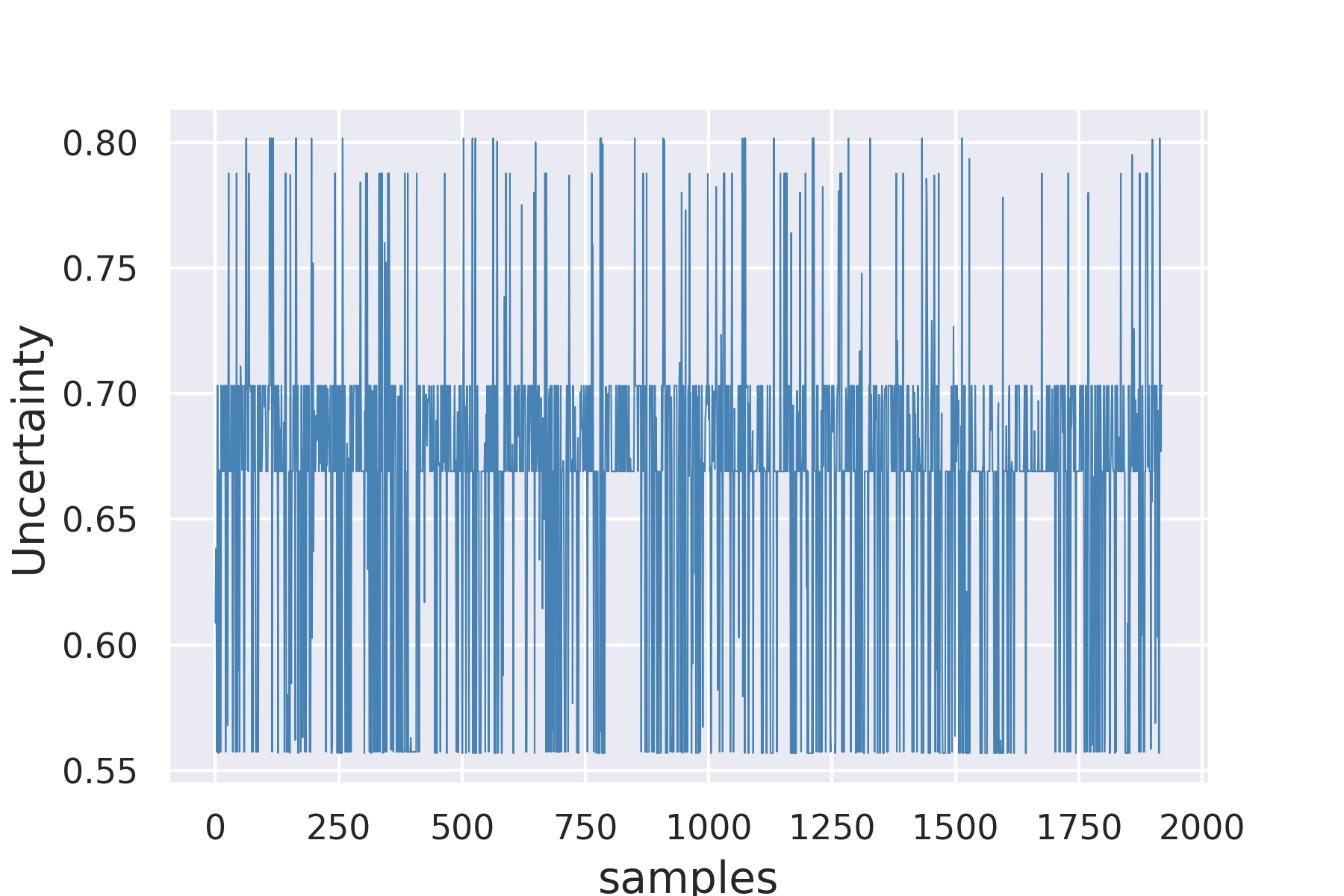}
	\caption{YAGER UQ for H5-1}\label{figure__use_UQY__H5_1}
	\end{subfigure}
	\caption{Classification and UQ results for ensemble classifiers. M5 is a BIN EC trained with (0,1), H6-1 is a MC EC trained with (0,1,2,6,12), and H5-1 is a MC EC trained with (0,3). Confusion matrices (a)-(c), Classification results (d)-(f), DSET UQ (g)-(i), and YAGER UQ (j)-(l).}\label{figure__results__classification}
\end{figure*}

Fig. \ref{figure__results__classification} presents the plots of selected ECs: binary M5 trained with (0,1), multiclass H6-1 trained with (0,1,2,6,12), and binary H5-1 trained with (0,3).

Figures \ref{figure__use_cm__M5}, \ref{figure__use_cm__H6_1} and \ref{figure__use_cm__H5_1} show the confusion matrices for the ECs M5, H6-1 and H5-1, respectively. As it can be noted, the confusion matrix (CM) with the best results (results mostly in the diagonal, which means that the predictions are equal to the ground truth) corresponds to the EC M5. The EC H6-1 presents a CM with mostly correct predictions, except for the cases (2,12), which present misclassifications. In contrast, the EC H5-1 presents the highest number of misclassifications, which implies the inability of the EC to distinguish between the cases (inadequate training).    
Figures \ref{figure__use_classification__M5}, \ref{figure__use_classification__H6_1} and \ref{figure__use_classification__H5_1} display the predictions (blue) compared to the ground truth (red) for the ECs M5, H6-1, and H5-1, respectively. The EC M5 shows the clearest plot, in which the predictions correspond mainly to the ground truth. Whereas the EC 6-1 presents good results for the fault cases (1,2,6). In contrast, the EC H5-1 presents a noisy plot, which, similar to its confusion matrix, implies poor performance.
Figures \ref{figure__use_UQDS__M5}, \ref{figure__use_UQDS__H6_1} and \ref{figure__use_UQDS__H5_1} present the DSET UQ for the ECs M5, H6-1 and H5-1, respectively. The uncertainty of the EC M5 presents mostly values near zero, implying reduced conflicting evidence. In the case of H6-1, the uncertainty has notable fluctuations for the cases (0,12). The EC H5-1 presents a plot full of fluctuations (with values ranging from zero to one), implying the inability to properly classify the faults and resulting in conflicting evidence during the fusion.
Figures \ref{figure__use_UQY__M5}, \ref{figure__use_UQY__H6_1} and \ref{figure__use_UQY__H5_1} show the YAGER UQ for the ECs M5, H6-1 and H5-1, respectively. The uncertainty of the EC M5 presents a notable amount of fluctuations, which implies the presence of conflicting evidence during the fusion. In the case of H6-1, the uncertainty has notable fluctuations for the cases (0,2,6,12). It is also noticeable that the uncertainty has a constant value under 0.4. The EC H5-1 presents a plot full of fluctuations, implying conflicting evidence during the fusion. It is visible that the uncertainty has a constant value of 0.7 in most of the cases.
    
\subsubsection{Anomaly Detection}\label{use_case__AD}
This subsection presents the anomaly detection performance of the ECs in the form of tables and plots.
Table \ref{table__results__pool_selection__selection__anomaly} presents a full series of experiments for multiclass ECs trained with the cases (0,1,2,6,12). 
The ECs H5-2, H6-2, and H7-2 present the highest average F1-score with a value of 0.63 for each EC. The inference time shows mixed results varying from 250s for M2 and 31611s for H9-2, which correspond to the relative time of 1\% and 100\%, respectively.  
The ECs H5-2, H6-2, and H7-2 have a training time of 29364s, 31101s, and 31598s, respectively. In addition, the relative time of the ECs H5-2, H6-2, and H7-2 have values of 93\%, 98\% and 100\%, respectively.
\begin{table}[!ht]
\centering
\caption{Selected ensemble multiclass classifiers using pool selection and the cases (0,1,2,6,12), and F1-score for anomaly detection performance.}
\begin{tabular}{c|c|ccc}
\hline
\multirow{2}{*}{\textbf{E-S}} & \multirow{2}{*}{\textbf{Exper.}} & \multirow{2}{*}{\textbf{F1}} & \multirow{2}{*}{\textbf{Infer. time {[}s{]}}} & \multirow{2}{*}{\textbf{Rel. time {[}\%{]}}} \\
 &  &  &  &  \\ \hline
2 & M2 & 0.45 & 250 & 1\% \\
3 & M3 & 0.40 & 256 & 1\% \\
4 & M4 & 0.37 & 289 & 1\% \\
5 & M5 & 0.38 & 890 & 3\% \\ \hline
2 & D2 & 0.47 & 15882 & 50\% \\
3 & D3 & 0.49 & 16036 & 51\% \\
4 & D4 & 0.61 & 29798 & 94\% \\
5 & D5 & 0.62 & 29869 & 94\% \\ \hline
2 & H2-1 & 0.45 & 557 & 2\% \\
2 & H2-2 & 0.47 & 15716 & 50\% \\
2 & H2-3 & 0.44 & 471 & 1\% \\ \hline
3 & H3-1 & 0.40 & 265 & 1\% \\
3 & H3-2 & 0.49 & 15987 & 51\% \\
3 & H3-3 & 0.52 & 439 & 1\% \\
3 & H3-4 & 0.49 & 378 & 1\% \\ \hline
4 & H4-1 & 0.37 & 298 & 1\% \\
4 & H4-2 & 0.61 & 28014 & 89\% \\
4 & H4-3 & 0.47 & 465 & 1\% \\
4 & H4-4 & 0.53 & 13308 & 42\% \\ \hline
5 & H5-1 & 0.38 & 419 & 1\% \\
5 & H5-2 & \textbf{0.63} & 29364 & 93\% \\
5 & H5-3 & 0.46 & 15977 & 51\% \\ \hline
6 & H6-1 & 0.55 & 13794 & 44\% \\
6 & H6-2 & \textbf{0.63} & 31101 & 98\% \\
6 & H6-3 & 0.46 & 15101 & 48\% \\
6 & H6-4 & 0.50 & 30559 & 97\% \\ \hline
7 & H7-1 & 0.46 & 15416 & 49\% \\
7 & H7-2 & \textbf{0.63} & 31598 & 100\% \\
7 & H7-3 & 0.51 & 15733 & 50\% \\
7 & H7-4 & 0.58 & 30669 & 97\% \\ \hline
8 & H8-1 & 0.51 & 13350 & 42\% \\
8 & H8-2 & 0.61 & 30586 & 97\% \\
8 & H8-3 & 0.34 & 15944 & 50\% \\
8 & H8-4 & 0.55 & 30725 & 97\% \\ \hline
9 & H9-1 & 0.32 & 16023 & 51\% \\
9 & H9-2 & 0.61 & 31611 & 100\% \\
9 & H9-3 & 0.55 & 31283 & 99\% \\ \hline
10 & H10-1 & 0.62 & 31084 & 98\% \\ \hline
\end{tabular}
\label{table__results__pool_selection__selection__anomaly}
\end{table}

As stated in the experiment design, the fault cases are injected as anomalies one at a time. Table \ref{table__results__anomaly__1} presents the binary ECs (M2, M5, and H6-2) trained with the cases (0,1), the binary ECs (M2, H3-3, and H5-1) trained with the cases (0,3), and the multiclass ECs (H5-2, H3-3, and H5-1) trained with the cases (0,1,2,6,12). 
The binary ECs M2, M5, and H6-2 trained with the cases (0,1) present mixed results, with an average F1-score of 0.45, 0.62, and 0.58. In contrast, the binary ECs M2, H3-3, and H5-1 trained with the cases (0,3) have an average F1-score of 0.3, 0.2, and 0.03. The multiclass ECs H5-2, H3-3, and H5-1 present comparable results with an average F1-score of 0.63 each.
The ECs M5 trained with cases (0,1), H5-2, H6-2, and H7-3 present comparable results with an average F1-score of 0.62, 0.63, 0.63, and 0.63, respectively. It is important to highlight the relative time difference between the ECs, specifically while observing M5 and H5-2 with values of 1.9\% and 92.9\%, respectively. %The last values can also be expressed as 0.0102 sample/s and 0.4855 sample/s, respectively.}

\begin{table*}[!ht]
\centering
\caption{Anomaly detection results of selected ensemble binary and multiclass classifiers.}
\begin{tabular}{c|ccc|ccc|ccc}
\hline
\multirow{2}{*}{\textbf{Fault}} & \multicolumn{3}{c|}{\textbf{Ensemble   BIN cases (0,1)}} & \multicolumn{3}{l|}{\textbf{Ensemble   BIN cases (0,3)}} & \multicolumn{3}{l}{\textbf{Ensemble MC (0,1,2,6,12)}} \\ \cline{2-10} 
 & \textbf{M2} & \textbf{M5} & \textbf{H6-2} & \textbf{M2} & \textbf{H3-3} & \textbf{H5-1} & \textbf{H5-2} & \textbf{H6-2} & \textbf{H7-2} \\ \hline
1 & 0.52 & \textbf{0.61} & 0.57 & 0.34 & 0.21 & 0.02 & \textbf{0.61} & \textbf{0.61} & 0.60 \\
2 & 0.42 & \textbf{0.63} & 0.55 & 0.36 & 0.22 & 0.05 & 0.56 & 0.54 & 0.56 \\
3 & 0.48 & 0.62 & 0.57 & 0.34 & 0.25 & 0.03 & \textbf{0.64} & 0.62 & 0.63 \\
4 & 0.46 & \textbf{0.60} & 0.58 & 0.35 & 0.23 & 0.05 & \textbf{0.60} & \textbf{0.60} & 0.59 \\
5 & 0.48 & \textbf{0.65} & 0.62 & 0.30 & 0.16 & 0.01 & 0.61 & 0.61 & 0.61 \\
6 & 0.18 & \textbf{0.91} & \textbf{0.91} & 0.06 & 0.12 & 0.00 & 0.42 & 0.45 & 0.45 \\
7 & 0.46 & \textbf{0.67} & 0.66 & 0.25 & 0.15 & 0.00 & 0.60 & 0.61 & 0.60 \\
8 & 0.39 & 0.44 & 0.44 & 0.32 & 0.21 & 0.03 & \textbf{0.71} & 0.69 & 0.68 \\
9 & 0.48 & 0.58 & 0.56 & 0.36 & 0.23 & 0.03 & 0.63 & \textbf{0.64} & \textbf{0.64} \\
10 & 0.44 & 0.55 & 0.51 & 0.32 & 0.23 & 0.03 & 0.64 & 0.65 & \textbf{0.66} \\
11 & 0.46 & \textbf{0.65} & 0.61 & 0.35 & 0.24 & 0.02 & 0.60 & 0.59 & 0.58 \\
12 & 0.48 & 0.57 & 0.55 & 0.26 & 0.17 & 0.01 & 0.62 & \textbf{0.63} & 0.62 \\
13 & 0.55 & 0.52 & 0.35 & 0.30 & 0.23 & 0.02 & \textbf{0.71} & 0.69 & 0.68 \\
14 & 0.45 & \textbf{0.63} & 0.58 & 0.34 & 0.24 & 0.04 & 0.62 & 0.62 & \textbf{0.63} \\
15 & 0.47 & 0.59 & 0.53 & 0.34 & 0.22 & 0.05 & \textbf{0.64} & 0.63 & \textbf{0.64} \\
16 & 0.51 & 0.57 & 0.51 & 0.36 & 0.24 & 0.01 & 0.61 & \textbf{0.63} & \textbf{0.63} \\
17 & 0.47 & 0.61 & 0.56 & 0.36 & 0.20 & 0.03 & \textbf{0.62} & \textbf{0.62} & \textbf{0.62} \\
18 & 0.28 & 0.80 & 0.82 & 0.06 & 0.02 & 0.00 & 0.85 & 0.89 & \textbf{0.90} \\
19 & 0.45 & 0.62 & 0.57 & 0.36 & 0.24 & 0.04 & 0.62 & 0.61 & \textbf{0.63} \\
20 & 0.49 & \textbf{0.61} & 0.56 & 0.35 & 0.24 & 0.01 & 0.60 & \textbf{0.61} & \textbf{0.61} \\
21 & 0.43 & \textbf{0.63} & 0.57 & 0.26 & 0.25 & 0.08 & \textbf{0.63} & \textbf{0.63} & \textbf{0.63} \\ \hline
Avg. F1-score & 0.45 & 0.62 & 0.58 & 0.30 & 0.20 & 0.03 & \textbf{0.63} & \textbf{0.63} & \textbf{0.63} \\
Rel. time {[}\%{]} & 0.4 & 1.9 & 33.5 & 0.9 & 0.5 & 21.3 & 92.9 & 98.4 & 100.0 \\ \hline
\end{tabular}
\label{table__results__anomaly__1}
\end{table*}

\begin{figure*}[!ht]
	\centering
	\begin{subfigure}[b]{0.3\textwidth}
	\includegraphics[width=\textwidth,keepaspectratio]{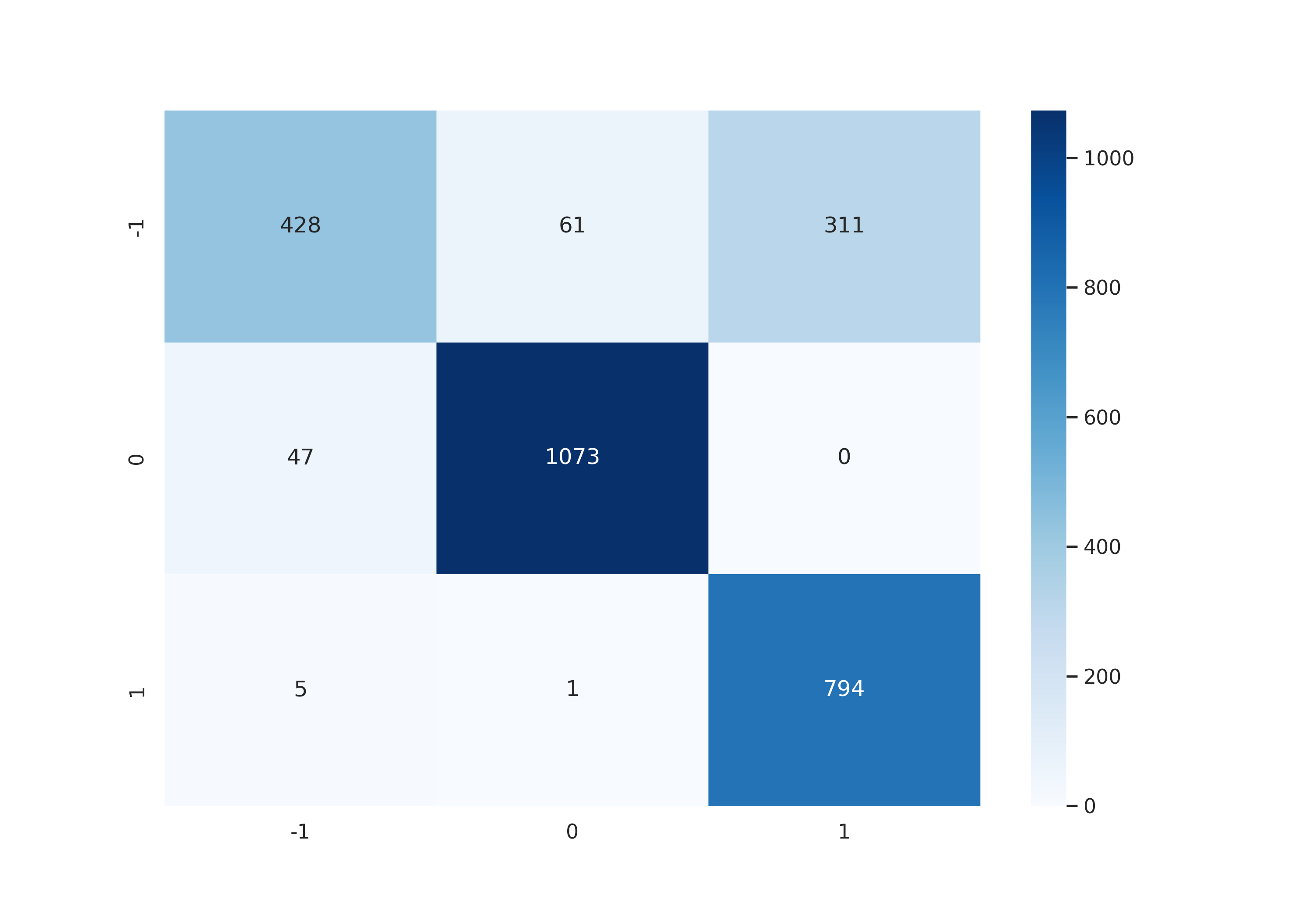} 
	\caption{CM for M5}\label{figure__use_cm__AD__M5}
	\end{subfigure}
	\begin{subfigure}[b]{0.3\textwidth}
	\includegraphics[width=\textwidth,keepaspectratio]{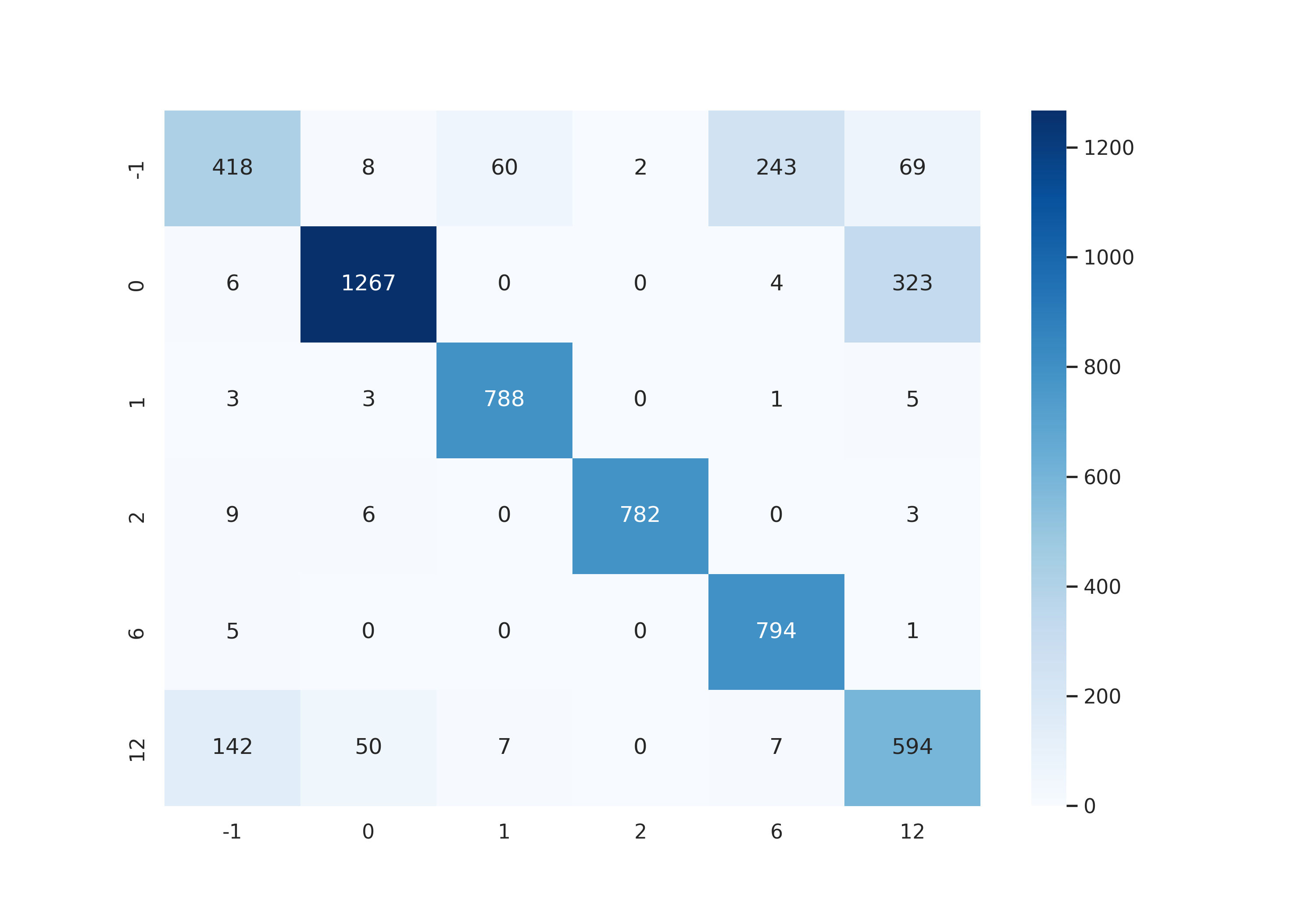} 
	\caption{CM for H5-2}\label{figure__use_cm__AD__H5_2}
	\end{subfigure}
	\begin{subfigure}[b]{0.3\textwidth}
	\includegraphics[width=\textwidth,keepaspectratio]{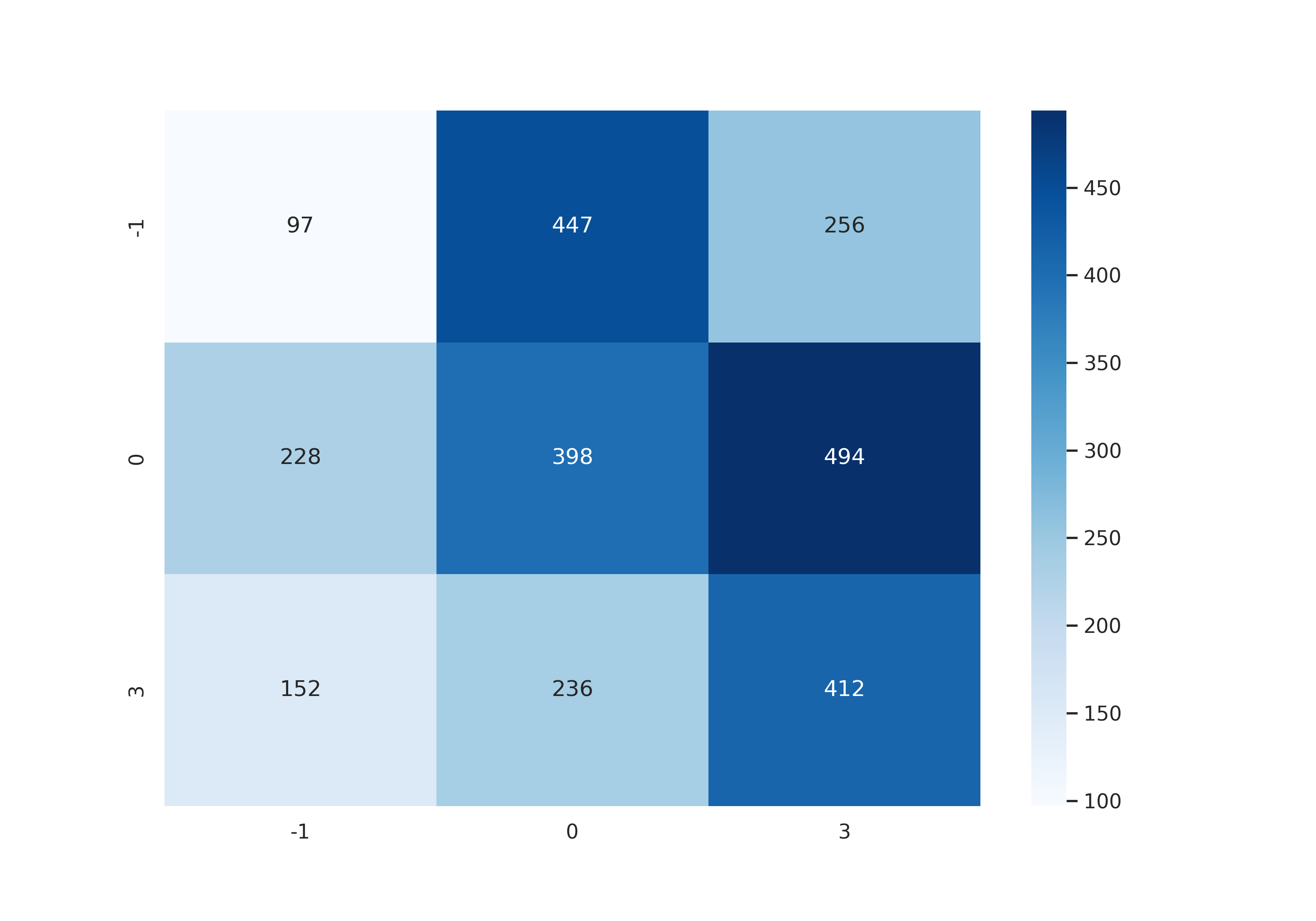} 
	\caption{CM for H3-3}\label{figure__use_cm__AD__H3_3}
	\end{subfigure}
	~
	\begin{subfigure}[b]{0.3\textwidth}
	\includegraphics[width=\textwidth,keepaspectratio]{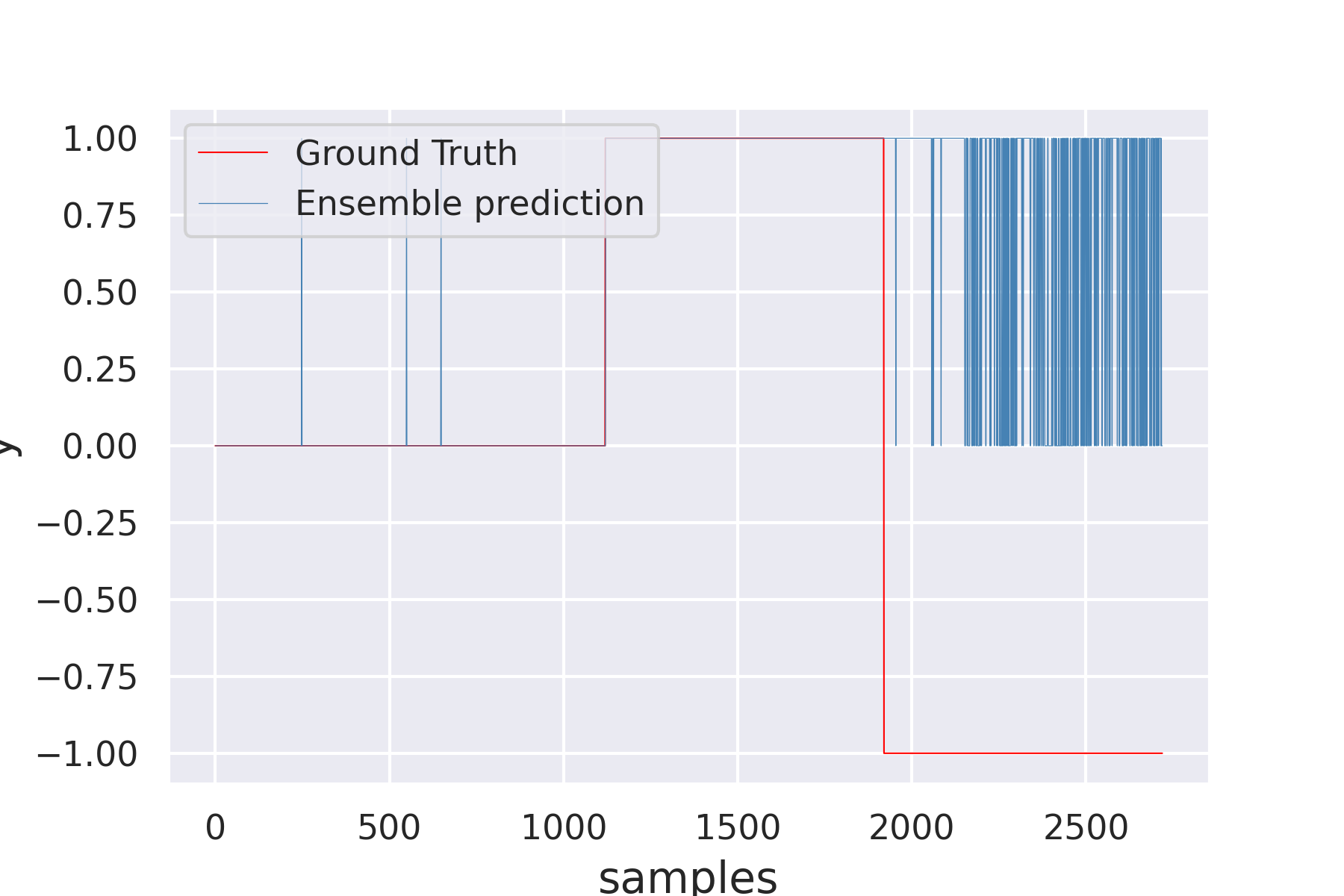} 
	\caption{Predictions for M5}\label{figure__use_classification__AD__M5}
	\end{subfigure}
	\begin{subfigure}[b]{0.3\textwidth}
	\includegraphics[width=\textwidth,keepaspectratio]{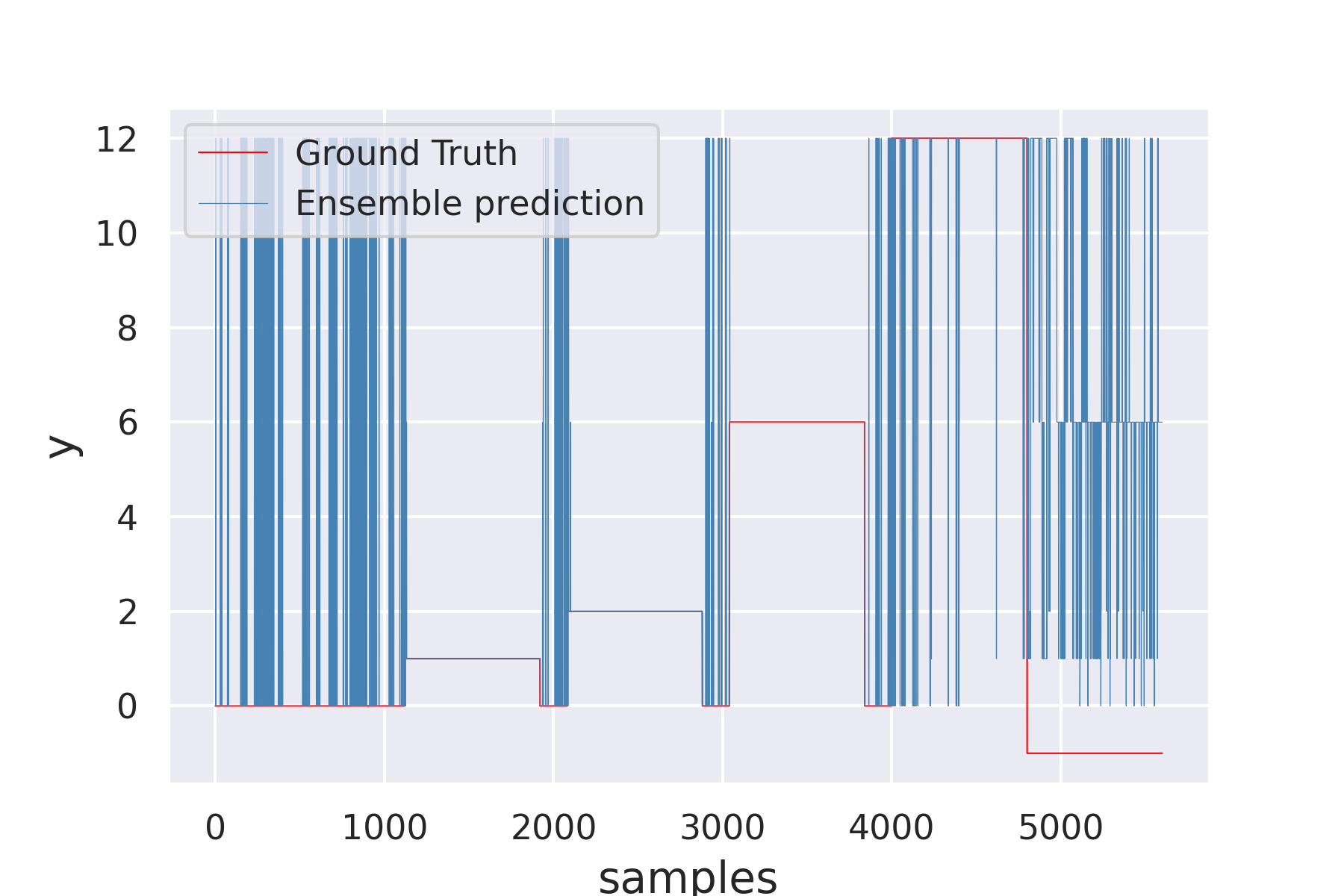} 
	\caption{Predictions for H5-2}\label{figure__use_classification__AD__H5_2}
	\end{subfigure}
	\begin{subfigure}[b]{0.3\textwidth}
	\includegraphics[width=\textwidth,keepaspectratio]{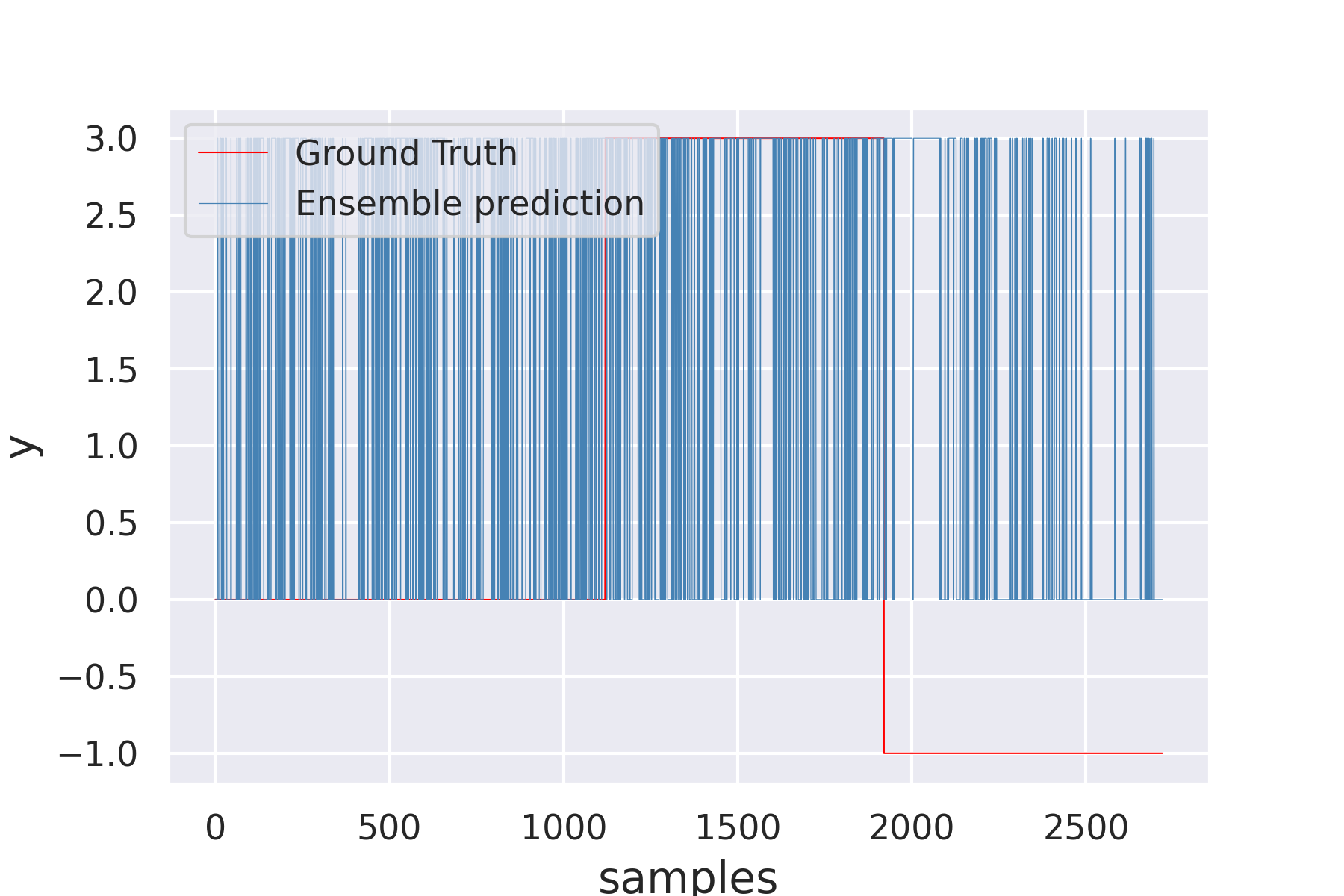} 
	\caption{Predictions for H3-3}\label{figure__use_classification__AD__H3_3}
	\end{subfigure}
	~
	\begin{subfigure}[b]{0.3\textwidth}
	\includegraphics[width=\textwidth,keepaspectratio]{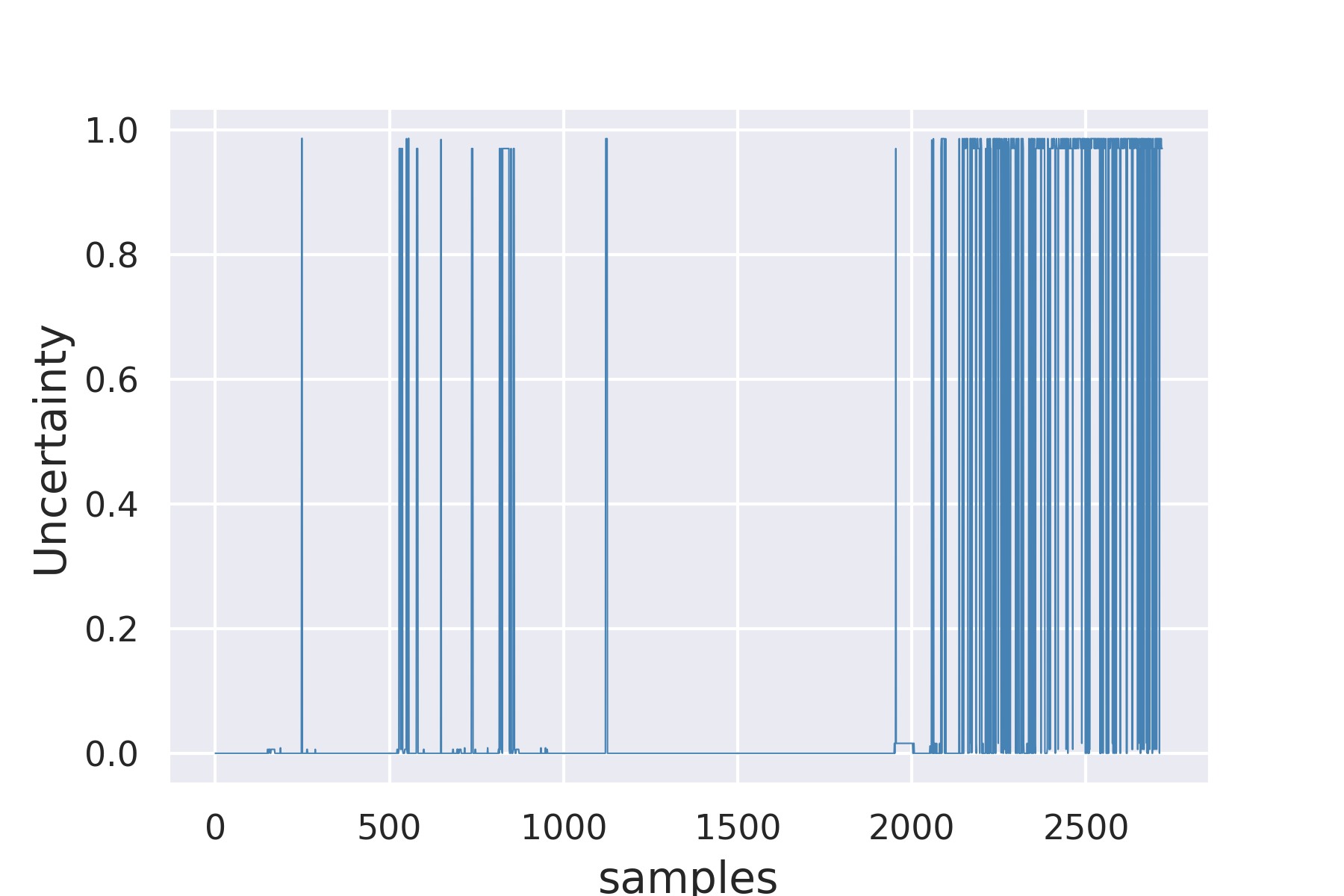}
	\caption{DSET UQ for M5}\label{figure__use_UQDS__AD__M5}
	\end{subfigure}
		\begin{subfigure}[b]{0.3\textwidth}
	\includegraphics[width=\textwidth,keepaspectratio]{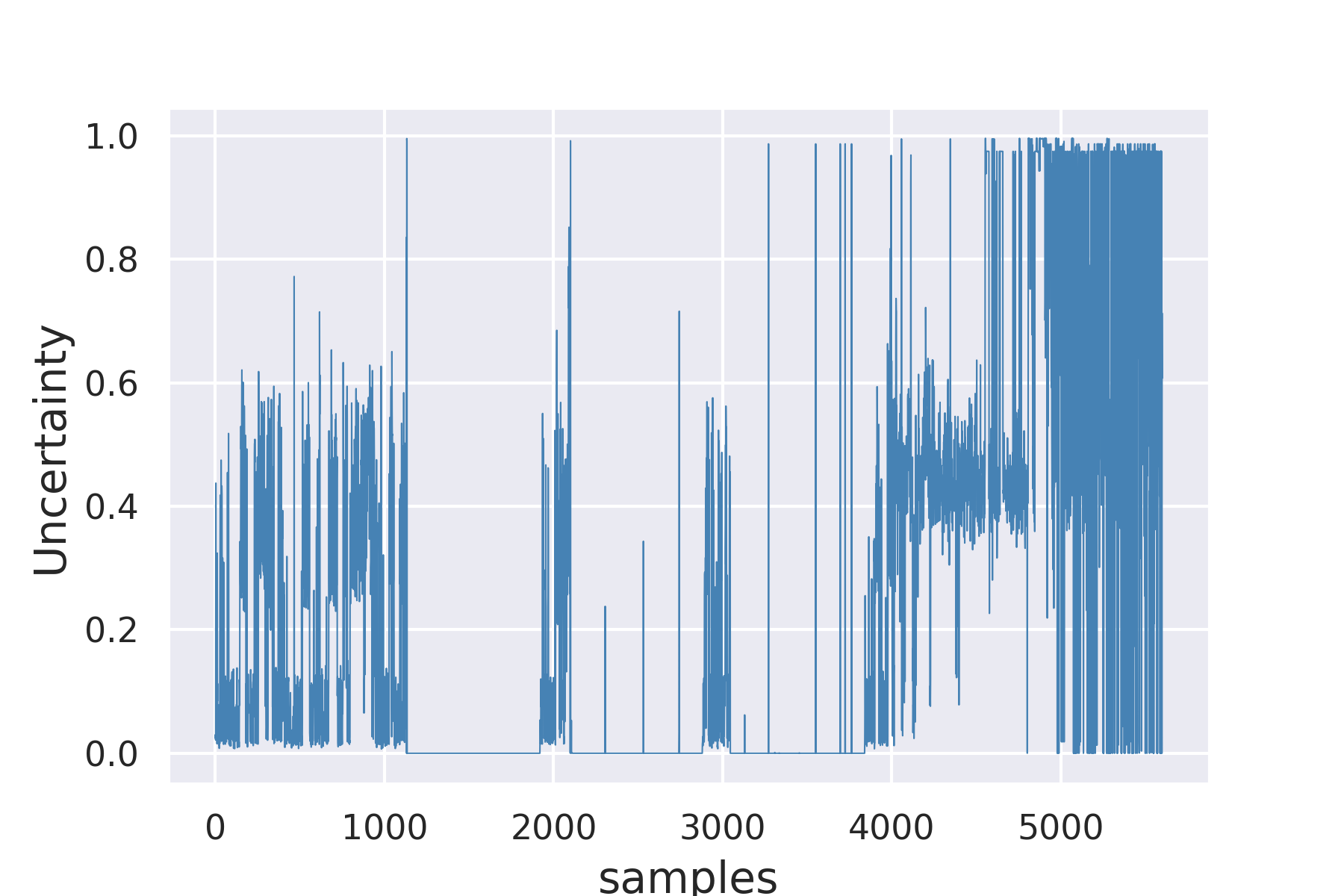}
	\caption{DSET UQ for H5-2}\label{figure__use_UQDS__AD__H5_2}
	\end{subfigure}
	\begin{subfigure}[b]{0.3\textwidth}
	\includegraphics[width=\textwidth,keepaspectratio]{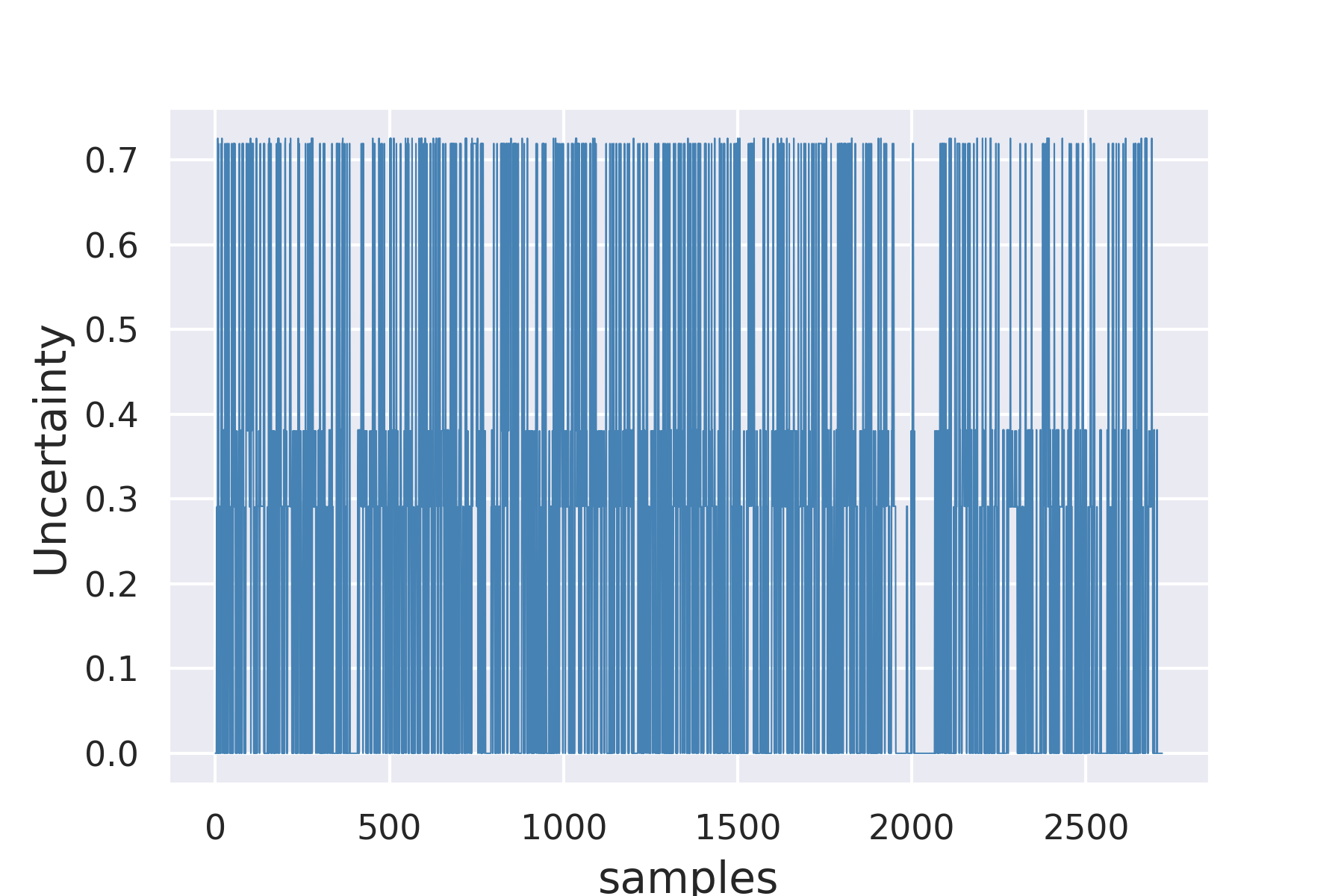}
	\caption{DSET UQ for H3-3}\label{figure__use_UQDS__AD__H3_3}
	\end{subfigure}
		~
	\begin{subfigure}[b]{0.3\textwidth}
	\includegraphics[width=\textwidth,keepaspectratio]{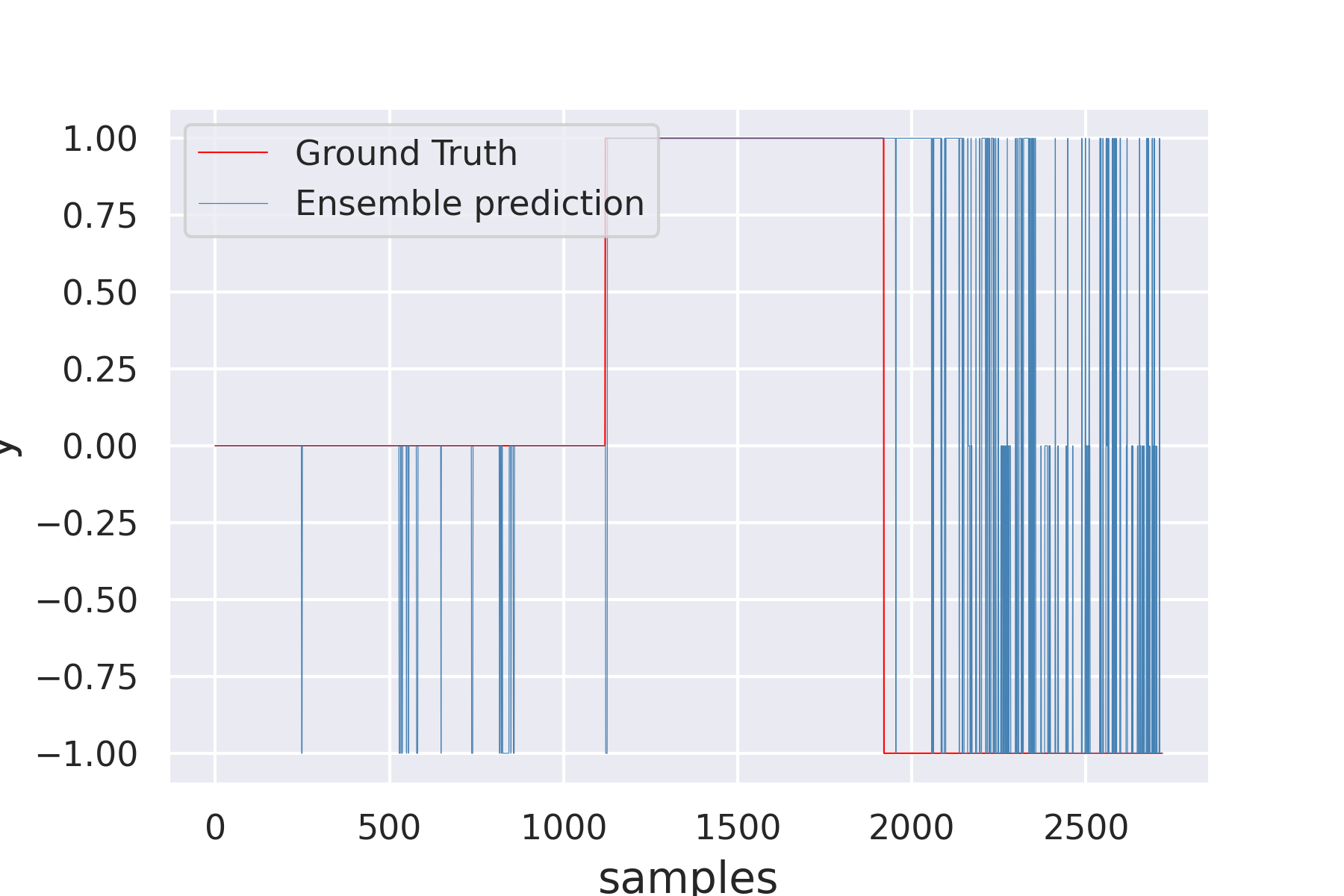} 
	\caption{AD for M5}\label{figure__use_Anomaly__AD__M5}
	\end{subfigure}
	\begin{subfigure}[b]{0.3\textwidth}
	\includegraphics[width=\textwidth,keepaspectratio]{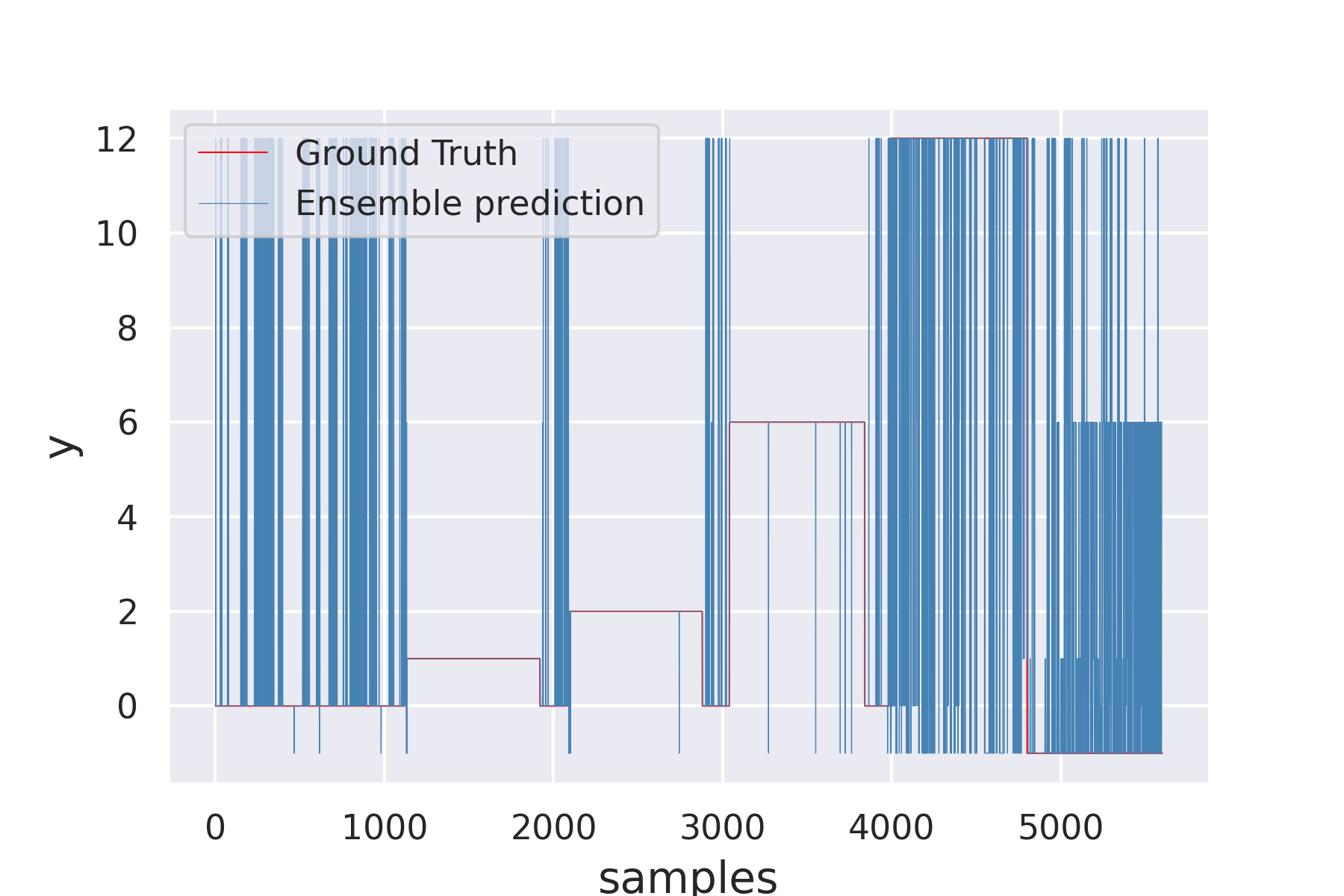} 
	\caption{AD for H5-2}\label{figure__use_Anomaly__AD__H5_2}
	\end{subfigure}
	\begin{subfigure}[b]{0.3\textwidth}
	\includegraphics[width=\textwidth,keepaspectratio]{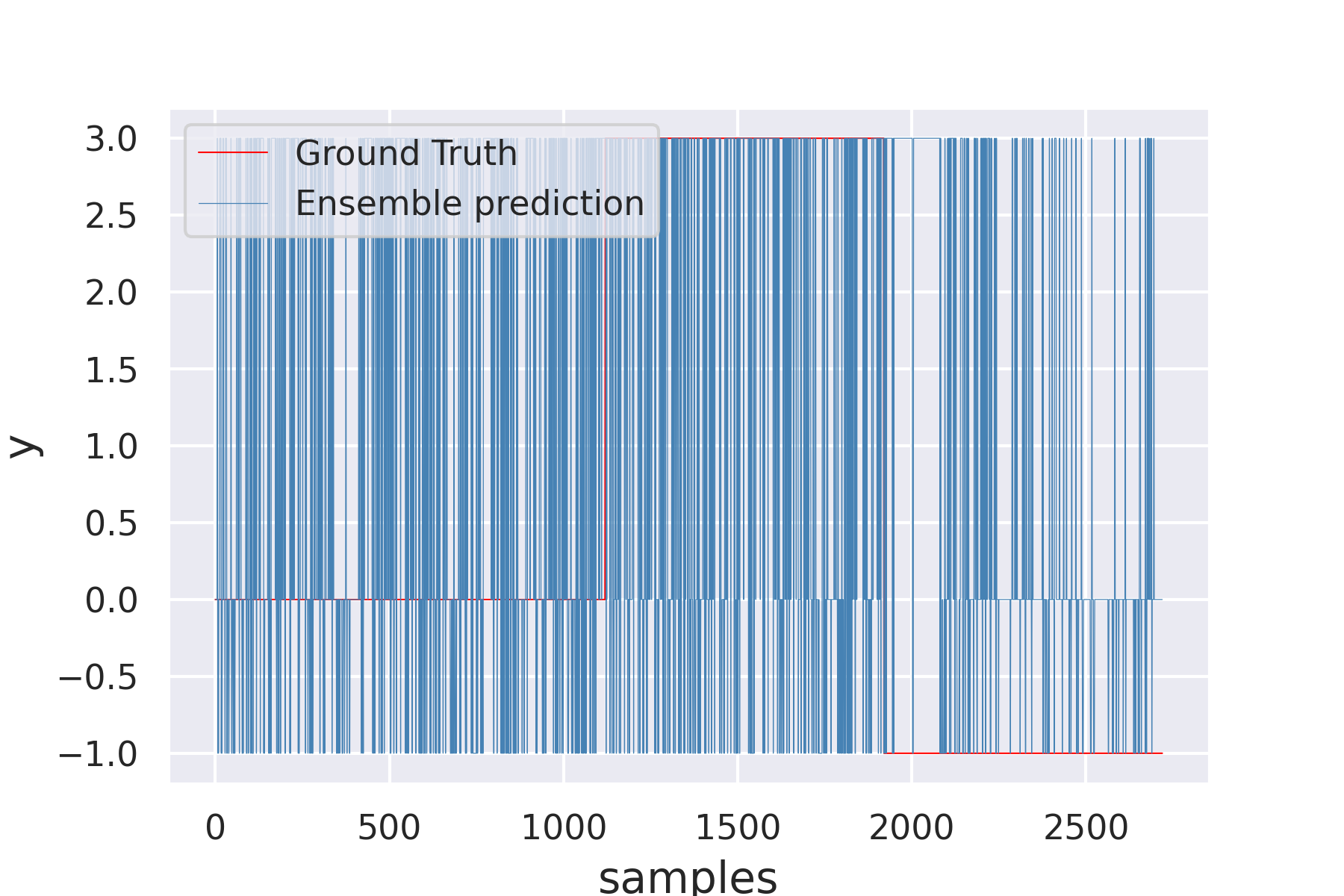} 
	\caption{AD for H3-3}\label{figure__use_Anomaly__AD__H3_3}
	\end{subfigure}
	\caption{Anomaly Detection and UQ results for ensemble classifiers. M5 is a BIN EC trained with cases (0,1), H5-2 is a MC EC trained with cases (0,1,2,6,12), and H3-3 is a MC EC trained with cases (0,3). Classification results (a)-(c), DSET UQ (d)-(f), YAGER UQ (g)-(i), and anomaly detection results (j)-(l) while injecting anomaly 7.
  }\label{figure__results__anomaly}
\end{figure*}

Fig. \ref{figure__results__anomaly} presents the plots of selected ECs: binary M5 trained with (0,1), multiclass H5-2 trained with (0,1,2,6,12), and binary H3-3 trained with (0,3) while injecting an anomaly (fault case 7).
Figures \ref{figure__use_cm__AD__M5}, \ref{figure__use_cm__AD__H5_2} and \ref{figure__use_cm__AD__H3_3} show the confusion matrices for the ECs M5, H5-2 and H3-3, respectively. As it can be noted, the confusion matrix (CM) with the best results corresponds to the EC M5, and most of the samples of the unknown condition are detected as an anomaly by the EC.
The EC H6-1 presents a CM with mostly correct predictions, except for the cases (2,12), which present misclassifications. The EC presents a good anomaly detection capability. 
In contrast, the EC H5-1 presents the highest number of misclassifications, as well as an inability to detect the unknown condition samples.    
Figures \ref{figure__use_classification__AD__M5}, \ref{figure__use_classification__AD__H5_2} and \ref{figure__use_classification__AD__H3_3} display the predictions (blue) compared to the ground truth (red) for the ECs M5, H5-2 and H3-3, respectively. The EC M5 shows the most precise plot, in which the predictions correspond mostly to the ground-truth. The EC 6-1 presents good results for the fault cases (1,2,6). In contrast, the EC H5-1 presents a noisy plot, which, similar to its confusion matrix, implies poor performance. It is important to remark that the EC, without 
the anomaly detection tracking results in an EC classifying the unknown data as the known cases and subsequently in a high fluctuation area.
Figures \ref{figure__use_UQDS__AD__M5}, \ref{figure__use_UQDS__AD__H5_2} and \ref{figure__use_UQDS__AD__H3_3} present the DSET UQ for the ECs M5, H5-2 and H3-3, respectively. The uncertainty of the EC M5 presents mostly values near zero, except for the unknown case (anomaly), which implies increasing conflicting evidence during the fusion. In the case of H3-3, the uncertainty has notable fluctuations for the cases (0,12), especially in the anomaly samples. The EC H5-1 presents a plot full of fluctuations, which implies a notable amount of conflicting evidence during the fusion.
Figures \ref{figure__use_Anomaly__AD__M5}, \ref{figure__use_Anomaly__AD__H5_2} and \ref{figure__use_Anomaly__AD__H3_3} show the anomaly detection (AD) tracking for the ECs M5, H5-2, and H3-3, respectively. The anomaly detection of the EC M5 identifies most of the anomalous samples. In the case of H5-2, the anomaly detection has notable fluctuations while injecting the anomalous samples. The EC H3-3 has the lowest performance in terms of anomaly detection, which is reflected in the high amount of fluctuations while injecting the anomaly.

\subsection{Comparison with Literature}\label{comparison__literature}
This subsection compares the ECET with literature, specifically while addressing classification and anomaly detection performance. For this purpose, we selected BIN and MC ECs, which will be compared with other literature approaches using F1-score and FDR (e.g., depending on the selected metric by the paper's author).  

\subsubsection{Comparison of classification results}

Table \ref{table__comparison__classification__1} displays the classification results of the binary ECs M5, H3-3, and H5-1. The binary ECs are trained using the normal condition and one fault case (e.g., the first row of Fault 1 represents the classification results of the binary EC trained with data of the normal condition and fault 1). The experiments address the 21 fault cases, which means that we have 21 binary ECs.
The overall classification results are comparable with the literature approaches such as support vector machines (SVM) and modified partial least squares(MPLS). The average FDR of EC H5-1 is 83.57\%, while SVM has a score of 81.49\% and MPLS has a score of 83.93\%. At first sight, the approaches yield similar results while looking into the detailed fault cases, specifically the easy faults (1,2,4,5,6,7,12,14 and 18). The rest of the fault cases show mixed (inferior or superior) results with respect to literature (e.g., medium and hard faults). 

\begin{table}[!ht]
\centering
\caption{Classification results comparison of binary ensemble classifiers respect to literature using FDR.}
\begin{tabular}{c|ccc|cc}
\hline
\multirow{2}{*}{\textbf{Fault}} & \multicolumn{3}{c|}{\textbf{BIN Ensemble}} & \multirow{2}{*}{\textbf{SVM}\cite{Assis2014}} & \multirow{2}{*}{\textbf{MPLS}\cite{Assis2014}} \\ \cline{2-4}
 & \textbf{M5} & \textbf{H3-3} & \textbf{H5-1} &  &  \\ \hline
1 & \textbf{99.75} & 99.38 & 99.32 & \textbf{97.93} & \textbf{100.00} \\
2 & \textbf{97.89} & 92.29 & 95.49 & 97.50 & \textbf{98.88} \\
\textit{3} & 48.57 & 48.85 & 49.88 & \textbf{59.29} & 18.75 \\
4 & \textbf{99.75} & 99.75 & 100.00 & \textbf{100.00} & 100.00 \\
5 & \textbf{100.00} & 99.88 & 100.00 & 100.00 & 100.00 \\
6 & \textbf{100.00} & 100.00 & 100.00 & 100.00 & 100.00 \\
7 & \textbf{99.88} & 99.50 & 100.00 & 100.00 & 100.00 \\
8 & 88.34 & 81.81 & 89.44 & \textbf{87.71} & \textbf{98.63} \\
9 & 46.94 & 47.38 & 47.45 & \textbf{51.58} & 12.13 \\
10 & 70.78 & 71.09 & 72.90 & \textbf{68.70} & \textbf{91.13} \\
11 & \textbf{87.70} & \textbf{91.04} & \textbf{89.79} & 68.28 & 83.25 \\
12 & 96.15 & 85.16 & 96.45 & 96.58 & \textbf{99.88} \\
13 & 40.65 & 41.11 & 43.10 & \textbf{65.75} & \textbf{95.50} \\
14 & \textbf{98.89} & 100.00 & 100.00 & \textbf{95.95} & 100.00 \\
15 & 49.91 & 48.68 & \textbf{52.87} & \textbf{52.41} & \textbf{23.25} \\
16 & 73.83 & 70.81 & 74.81 & \textbf{70.59} & \textbf{94.28} \\
17 & 86.02 & 79.10 & 86.02 & \textbf{94.68} & \textbf{97.13} \\
18 & 87.49 & 95.89 & \textbf{97.97} & \textbf{91.54} & 91.25 \\
19 & 75.54 & 83.02 & 85.75 & \textbf{82.83} & \textbf{94.25} \\
20 & \textbf{77.53} & 75.79 & 75.51 & \textbf{88.41} & \textbf{91.50} \\
21 & \textbf{98.16} & 98.16 & \textbf{98.16} & \textbf{41.58} & 72.75 \\ \hline
\multicolumn{1}{l|}{Avg. FDR} & 82.08 & 81.37 & \textbf{83.57} & \textbf{81.49} & \textbf{83.93} \\
%\multicolumn{1}{r|}{Time {[}s{]}} & N/A & N/A & N/A & N/A & N/A  \\ 
\hline
\end{tabular}
\label{table__comparison__classification__1}
\end{table}

Table \ref{table__comparison__classification__2} showcases the classification results of the multiclass ECs trained with the fault cases (1,2,6 and 12). The F1-score results are compared with the literature approaches multilayer perceptron (MLP), (ACGAN-MLP) and (MDAC). The ECs M2, M5, and H5-3 are selected to be compared with the literature approaches multilayer perceptron (MLP), auxiliary classifier generative adversarial network (ACGAN-MLP), and multigenerator data augmentation classifier (MDAC). 
The overall classification results of the ECs are superior with respect to the literature approaches. The average F1-score of M2, M5, and H5-3 are 98\%, 99\%, and 99\%, respectively, whereas the results of MLP, ACGAN-MLP, and MDAC are 70\%, 83.25\%, and 88.25\% respectively.
The EC H5-3 has superior results for fault cases 1 and 12, with F1-scores of 99\% and 98\%, while MDAC results are 83\% and 71\% for fault cases 1 and 12. The results for fault cases 2 and 6 are comparable, showing slight differences.

\begin{table*}[!ht]
\centering
\caption{Multilabel classification comparison of ensemble classifiers respect to literature using the hard faults of TE and F1-score.}
\begin{tabular}{c|ccc|ccc}
\hline
\multirow{2}{*}{\textbf{Fault}} & \multicolumn{3}{c|}{\textbf{Ensemble MC (1,2,6,12)}} & \multirow{2}{*}{\textbf{MLP} \cite{JiangGe2021}} & \multirow{2}{*}{\textbf{ACGAN-MLP} \cite{JiangGe2021}} & \multirow{2}{*}{\textbf{MDAC} \cite{JiangGe2021}} \\
 & \textbf{M2} & \textbf{M5} & \textbf{H5-3} &  &  &  \\ \hline
1 & 97.00 & 99.00 & \textbf{99.00} & 79.00 & \textbf{94.00} & \textbf{83.00} \\
2 & 99.00 & 99.00 & 99.00 & 91.00 & 99.00 & \textbf{100.00} \\
6 & 100.00 & 100.00 & \textbf{100.00} & 92.00 & 82.00 & \textbf{99.00} \\
12 & 96.00 & 98.00 & \textbf{98.00} & 18.00 & 58.00 & \textbf{71.00} \\ \hline
\multicolumn{1}{r|}{Avg. F1-score} & 98.00 & 99.00 & \textbf{99.00} & 70.00 & 83.25 & \textbf{88.25} \\
\multicolumn{1}{r|}{Rel. time {[}\%{]}} & 0.9 & 1.6 & \textbf{2.6} & N/A & N/A & N/A \\ \hline
\end{tabular}
\label{table__comparison__classification__2}
\end{table*}

Table \ref{table__comparison__classification__3} presents the results of the same ECs and literature approaches of Table \ref{table__comparison__classification__2} but using the metric FDR instead. 
The overall classification results are comparable with respect to the literature approaches. The average FDR of M2, M5, and H5-3 are 97.5\%, 99.25\%, and 99\%, respectively, whereas the results of MLP, ACGAN-MLP, and MDAC are 77.5\%, 85\%, and 88.75\% respectively.
In this case, the results show slight differences while considering the fault cases 1, 2, and 6 for the ECs and the literature approaches. However, the FDR of the ECs M2, M5, and H5-3 (e.g., with FDR of 97.5\%, 99.25\%, and 99\% respectively) are remarkably superior to the literature approaches (e.g., the higher score corresponds to MDAC with an FDR of 88.75\%).

\begin{table*}[!ht]
\centering
\caption{Multilabel classification comparison of ensemble classifiers respect to literature using the hard faults of TE and FDR.}
\begin{tabular}{c|ccc|ccc}
\hline
\multirow{2}{*}{\textbf{Fault}} & \multicolumn{3}{c|}{\textbf{Ensemble MC (1,2,6,12)}} & \multirow{2}{*}{\textbf{MLP} \cite{JiangGe2021}} & \multirow{2}{*}{\textbf{ACGAN-MLP} \cite{JiangGe2021}} & \multirow{2}{*}{\textbf{MDAC} \cite{JiangGe2021}} \\
 & \textbf{M2} & \textbf{M5} & \textbf{H5-3} &  &  &  \\ \hline
1 & 99.00 & 99.00 & \textbf{99.00} & 100.00 & 99.00 & \textbf{100.00} \\
2 & 98.00 & 98.00 & 98.00 & 100.00 & 100.00 & \textbf{100.00} \\
6 & 99.00 & 100.00 & \textbf{100.00} & 100.00 & 100.00 & \textbf{100.00} \\
12 & 94.00 & 100.00 & \textbf{99.00} & 10.00 & 41.00 & 55.00 \\ \hline
\multicolumn{1}{r|}{Avg. F1-score} & 97.50 & 99.25 & \textbf{99.00} & 77.50 & 85.00 & \textbf{88.75} \\
\multicolumn{1}{r|}{Rel. time {[}\%{]}} & 0.9 & 1.6 & \textbf{2.6} & N/A & N/A & N/A \\ \hline
\end{tabular}
\label{table__comparison__classification__3}
\end{table*}

\subsubsection{Comparison of anomaly detection results}
It is essential to highlight that the usual candidates for anomaly detection are approaches based on unsupervised classification. Therefore, the ECET provides a different approach to identifying unknown conditions.
Table \ref{table__comparison__anomaly__1} showcases the anomaly detection results of the binary and multiclass ECs. The binary ECs M2, M5, and H6-2 are trained with the cases (0,1), whereas the binary ECs M2, H3-3, and H5-1 are trained with the cases (0,3). The multiclass ECs H5-2, H6-2, and H7-2 are trained with the cases (0,1,2,6,12). The testing data, used as unknown or anomaly data, is composed of cases (3, 9, 15, and 21).
We use the F1-score to compare the ECs with respect to the literature, specifically to the unsupervised method (Top-K DCCA).
The results of the ECs are mixed, in which the binary ECs M5 and H6-2 (trained with the cases (0,1) ) show superior results with scores of 60.21\% and 55.93\% respectively, compared to the Top-K DCCA score of 50.04\%. Likewise, the MC ECs H5-2, H6-2, and H7-2 show superior results with average F1-scores of 63.69\%, 63.17\%, and 63.45\%. 
The best results of an EC are achieved by H5-2, which has F1-scores of 64.3\%, 63.01\%, 64.35\%, and 63.10\% for the fault cases (3,9,15,21), respectively. These results are better than Top-K DCCA with the scores 53.82\%, 52,31\%, 43.98\%, and 50.05\% for the fault cases (3,9,15,21), respectively.

\begin{table*}[!ht]
\centering
\caption{Anomaly detection comparison of ensemble classifiers respect to literature using the hard faults of TE and F1-score.}
\begin{tabular}{c|ccc|ccc|ccc|c}
\hline
\multirow{2}{*}{\textbf{Fault}} & \multicolumn{3}{c|}{\textbf{Ensemble   BIN (0,1)}} & \multicolumn{3}{c|}{\textbf{Ensemble BIN (0,3)}} & \multicolumn{3}{c|}{\textbf{Ensemble MC (0,1,2,6,12)}} & \multirow{2}{*}{\textbf{Top-K DCCA} \cite{ChadhaIslam2021}} \\ \cline{2-10}
 & \textbf{M2} & \textbf{M5} & \textbf{H6-2} & \textbf{M2} & \textbf{H3-3} & \textbf{H5-1} & \textbf{H5-2} & \textbf{H6-2} & \textbf{H7-2} &  \\ \hline
3 & 48.49 & \textbf{61.58} & 56.97 & 0.34 & 24.93 & 2.60 & \textbf{64.30} & 62.20 & 63.14 & 53.82 \\
9 & 47.63 & \textbf{57.76} & 56.46 & 0.36 & 23.35 & 3.07 & 63.01 & \textbf{64.00} & 63.66 & 52.31 \\
15 & 46.75 & \textbf{58.59} & 52.81 & 0.34 & 22.16 & 4.68 & \textbf{64.35} & 63.47 & 63.52 & 43.98 \\
21 & 43.13 & 62.91 & \textbf{57.47} & 0.26 & 24.54 & 8.49 & 63.10 & 63.01 & \textbf{63.47} & 50.05 \\ \hline
\multicolumn{1}{r|}{Avg. F1-score} & 46.50 & \textbf{60.21} & 55.93 & 0.33 & 23.74 & 4.71 & \textbf{63.69} & 63.17 & 63.45 & 50.04 \\
\multicolumn{1}{r|}{Rel. time {[}\%{]}} & 0.3 & 1.9 & 33.5 & 1.4 & 0.5 & 21.3 & 92.9 & 98.4 & 100 & N/A  \\ \hline
\end{tabular}
\label{table__comparison__anomaly__1}
\end{table*}

Table \ref{table__comparison__anomaly__2} presents the anomaly detection results of the %binary and 
multiclass ECs. 
%The binary ECs M5 and H6-2 are trained with cases (0,1), whereas the binary EC H3-3 is trained with cases (0,3). 
The multiclass ECs M3 and H3-4 are trained with the cases (0,1,2,6,12). The testing data, used as unknown or anomaly data, comprises all the fault cases (1,2,...,20, 21). We use the FDR to compare the ECs with respect to the literature, specifically the methods DPCA-DR, principal component analysis (PCA), autoencoder (AE), adversarial AE (AAE), and modified partial least squares (MOD-PLS).
%The binary ECs present a poor performance with an average FDR of 48.78\%, 42.35\%, and 16.95\% for the ECs M5, H6-2, and H3-3, respectively. 
The multiclass EC H3-4 presents better results with an average FDR of 73.76\%, in some cases comparable to literature, specifically using PCA 76.68\%, AE 76.56\%, and AAE 78.55\%. The multiclass EC M3 presents the best average FDR with a score of 87.97\%, which are superior to DPCA-DR and MOD-PLS with scores of 83.51\% and 83.83\%, respectively.
The results of M3 are mixed while considering all the fault cases. However, the ability of the EC to identify the hard faults is remarkable. The M3 results achieved an FDR of 91.88\%, 90.75\%, 91.25\%, and 94.13\% for fault cases 3, 9, 15, and 21, respectively. The best result for fault case 3 corresponds to AAE with an FDR of 34.88\%, whereas the highest result for fault case 9 is achieved by AAE with an FDR of 33.62\%. The best result for fault case 15 corresponds to DPCA-DR with an FDR of 38.5\%, whereas the highest result for fault case 21 is achieved by MOD-PLS with an FDR of 72.66\%.  

\begin{table*}[!ht]
\centering
\caption{Anomaly detection comparison of ensemble classifiers respect to literature using all faults of TE and FDR.}
\begin{tabular}{c|cc|c|ccc|c}
\hline
\multirow{2}{*}{\textbf{Fault}} & \multicolumn{2}{c|}{\textbf{Ensemble MC (0,1,2,6,12)}} & \multirow{2}{*}{\textbf{DPCA-DR} \cite{RatoReis2013}} & \multirow{2}{*}{\textbf{PCA} \cite{JangHong2022}} & \multirow{2}{*}{\textbf{AE} \cite{JangHong2022}} & \multirow{2}{*}{\textbf{AAE} \cite{JangHong2022}} & \multirow{2}{*}{\textbf{MOD-PLS} \cite{YinDing2011}} \\ \cline{2-3}
 & \textbf{M3} & \textbf{H3-4} &  &  &  &  &  \\ \hline
1 & 93.50 & 82.13 & 99.60 & \textbf{100.00} & 99.75 & \textbf{100.00} & 99.88 \\
2 & 94.75 & 83.63 & 98.50 & 98.75 & \textbf{99.00} & \textbf{99.00} & 98.75 \\
3 & \textbf{91.88} & 75.00 & 2.10 & 25.88 & 25.88 & 34.88 & 18.73 \\
4 & 89.75 & 76.50 & 99.80 & 98.75 & 99.38 & 98.62 & \textbf{99.88} \\
5 & 90.63 & 75.25 & \textbf{99.90} & 51.12 & 49.75 & 55.00 & 99.88 \\
6 & 97.50 & 64.75 & 99.90 & \textbf{100.00} & \textbf{100.00} & \textbf{100.00} & 99.88 \\
7 & 88.25 & 77.50 & 99.90 & \textbf{100.00} & \textbf{100.00} & \textbf{100.00} & 99.88 \\
8 & 87.00 & 76.38 & 98.50 & 98.75 & \textbf{99.00} & 97.88 & 98.50 \\
9 & \textbf{90.75} & 74.88 & 2.00 & 22.62 & 25.50 & 33.62 & 12.11 \\
10 & 88.13 & 75.13 & \textbf{95.60} & 69.50 & 67.88 & 74.00 & 91.01 \\
11 & 92.50 & 73.63 & \textbf{96.50} & 82.25 & 84.38 & 82.00 & 83.15 \\
12 & 81.50 & 80.50 & \textbf{99.80} & 99.00 & 99.38 & 99.75 & 99.75 \\
13 & 84.38 & 71.38 & 95.80 & 95.88 & \textbf{96.38} & 96.25 & 95.38 \\
14 & 91.00 & 71.75 & 99.80 & \textbf{100.00} & \textbf{100.00} & \textbf{100.00} & 99.88 \\
15 & \textbf{91.25} & 66.25 & 38.50 & 27.50 & 31.50 & 31.25 & 23.22 \\
16 & 90.25 & 76.25 & \textbf{97.60} & 64.50 & 63.62 & 64.75 & 94.26 \\
17 & 91.25 & 75.75 & 97.60 & \textbf{99.50} & 95.50 & 96.00 & 97.00 \\
18 & 37.63 & 56.00 & 90.50 & 92.88 & 92.88 & \textbf{95.00} & 91.14 \\
19 & 92.00 & 74.50 & \textbf{97.10} & 50.50 & 47.75 & 53.87 & 94.13 \\
20 & 89.38 & 72.63 & 90.80 & 73.12 & 69.25 & 78.62 & 91.26 \\
21 & \textbf{94.13} & 69.25 & 53.90 & 59.75 & 60.88 & 59.00 & 72.66 \\ \hline
Avg. F1-score & \textbf{87.97} & 73.76 & 83.51 & 76.68 & 76.56 & 78.55 & 83.83 \\
Rel. time {[}\%{]} & 0.8 & 1.2 & N/A & N/A & N/A & N/A & N/A \\ \hline
\end{tabular}
\label{table__comparison__anomaly__2}
\end{table*}

\subsection{Discussion}
%UQ
The \textit{uncertainty quantification} (UQ) assessed the training capability of the individual classifiers (e.g., SVM with low uncertainty and high performance, while ale with high uncertainty and low performance). In the case of the EC, the UQ is successfully used to detect anomalies (e.g., high uncertainty reflects the likeliness of an unknown condition).   

%pool selection
For the \textit{ensemble classification}, a pool of ML-based models and NN-based models were trained. Because the ensemble classification relies on the heterogeneity of its sources, a combined expert-diversity strategy was used during the \textit{pool selection}. The proposed strategy systematically reduces the number of possible combinations while providing diverse ensemble classifiers (EC). The performance of the ECs was comparable, and in some cases superior, with respect to the individual classifiers.  
% EC
In addition, the experiments consider the impact of varying the ensemble size and different classifiers architectures (e.g., ML-based, NN-based, and Hybrid EC), as well as binary and multiclass ECs. The ensemble classification uses the evidence theory to perform the fusion of the classifiers at the decision level. To this end, the rules of combination DSET and YAGER were applied. The first one was mainly used to combine the classifiers, whereas both were used to detect uncertainty changes. 
Some ECs provided better results such as the case of the multiclass EC H6-1 (trained with the cases (0,1,2,6,12)) with an average F1-score of 0.97. In the case of the multiclass EC H4-1 (trained with the cases (0,3,9,15,21)), the EC had a poor performance with an average F1-score of 0.23. The poor performance can be attributed to the challenging fault cases and to a (poor) training of the classifiers that conform the EC.
The training time showed notable differences, ranging from 476s for the EC H6-1 and 12s for the EC H4-1. The latest represented with the relative time with values of 38\% and 3\%, respectively. 
The EC's constraints rely on the classifiers' static behavior since the classifiers were trained once with specific data of the input space. The reliability of a model is related directly to how the training data represents the input space or how representative the training samples are with respect to the overall data.
Another constraint relies on the primary assumption of the mutual exclusion between the faults. This situation can occur in real applications, although a combination of faults is a common scenario.
% AD through UQ
The \textit{anomaly detection} relies on the ensemble classification performance. This situation is visible while comparing the average F1-score of binary EC M5 (trained with the cases (0,1)) and the binary EC H3-3 (trained with the cases (0,3)), which have values of 0.62 and 0.20, respectively. As it was noted in the classification performance subsection, the ECs trained with the cases (0,3) showed poor classification performance. Consequently, the anomaly detection performance of the ECs trained with the cases (0,3) is also poor. Important to note is that classifiers with poor performance add noise during the fusion.
This situation is clear while visualizing the anomaly detection results in the plot prediction versus ground truth.

\section{Conclusion}\label{section__conclusions}
We presented a novel approach for binary and multiclass supervised classification and anomaly detection of unknown conditions using ensemble classification and evidence theory (ECET).  
%UQ
The \textit{uncertainty quantification (UQ)} during the training assessed the learning capability of the trained models. Whereas the UQ, during the inference of the ensemble classifier (EC), tracked the uncertainty changes while feeding testing data. This last feature is used to detect anomalies because a high uncertainty would correspond to a high likeliness of an anomaly, as demonstrated in the results.  

%pool selection
The \textit{pool selection} strategy plays a vital role in selecting the classifiers that form the EC. The combination complexity is reduced, and the resulting pool is derived from a selection of an optimal EC with respect to individual classifiers (e.g., NN-based and ML-based classifiers).   
%EC
We test different binary and multiclass \textit{ensemble classifiers} using the Tennessee Eastman benchmark, obtaining favorable results (e.g., depending on the EC). Selected ECs present comparable results, and in some cases superior results, with respect to Literature.  
%AD
We propose a hybrid strategy based on DSET and Yager UQ for the \textit{detection of anomalies}. The strategy was successful because the EC could detect anomalies while feeding unknown data. The latest was also validated while comparing to other literature approaches.  

%future
Future research includes using the anomaly detection approach for an automatic update of the ensemble classifiers. Different pool selection strategies will remain in focus since they proved to have a significant impact on ensemble selection. Further experiments will include training different classifiers using different feature reduction strategies with the same model.

\bibliographystyle{IEEEtran}
% argument is your BibTeX string definitions and bibliography database(s)
\bibliography{IEEEabrv,ECET}

%\textcolor{white}{\lipsum[3-10]}

\begin{IEEEbiography}[{\includegraphics[width=1in,height=1.25in,clip,keepaspectratio]{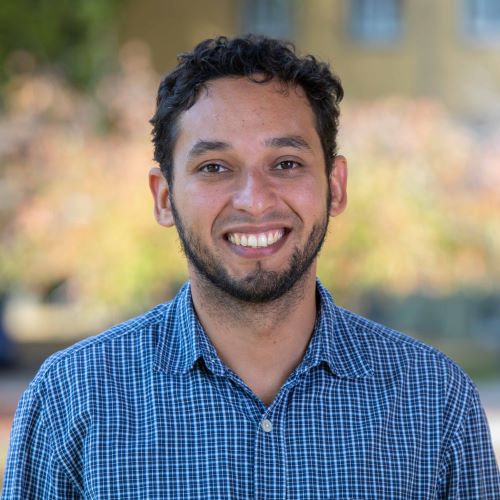}}]{Fernando Ar\'{e}valo} (M'16--SM'22) %--SM'81--F'87) 
%(Member,IEEE)
received the Engineer degree in electrical engineering from the Universidad de El Salvador, San Salvador, El Salvador, in 2005, and the M.Sc. degree in systems engineering and engineering management from the South Westphalia University of Applied Sciences, Soest, Germany, in 2012. He is currently pursuing the Ph.D. degree with the Ruhr-Universität Bochum, Bochum, Germany. From 2008 to 2015, he worked as a Project Engineer with Kimberly Clark, and MTU Friedrichshafen, among others. 
From 2016 to 2021, he worked as a Research Assistant with the Automation Technology Department, South Westphalia University of Applied Sciences. Since 2022, he has been working as a Data Science Consultant IoT with the company adesso. His research interests include data-driven fault diagnosis, information fusion, uncertainty quantification, knowledge extraction, the IoT/IIoT, and augmented reality with application in the process industry and manufacturing. 
\end{IEEEbiography}

\begin{IEEEbiography}[{\includegraphics[width=1in,height=1.25in,clip,keepaspectratio]{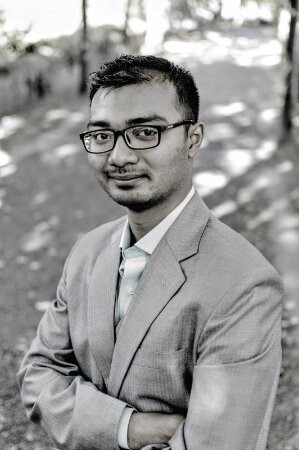}}]{M. Tahasanul Ibrahim}  (M'20) %--SM'81--F'87) 
%(Member,IEEE)
received the bachelor's degree in electrical and electronics engineering from American International University, Bangladesh, in 2014, and the M.Sc. degree in systems engineering and engineering management from the South Westphalia University of Applied Sciences, Soest, Germany, in 2020. Since 2021, he has been working as a Research Assistant with the Automation Technology and Learning Systems Department, South Westphalia University of Applied Sciences. His research interests include information fusion based on uncertainty, rule-based systems, deep neural networks in computer vision, classified maneuver prediction and knowledge attenuation techniques, and back-end development in container systems.
\end{IEEEbiography}

\begin{IEEEbiography}[{\includegraphics[width=1in,height=1.25in,clip,keepaspectratio]{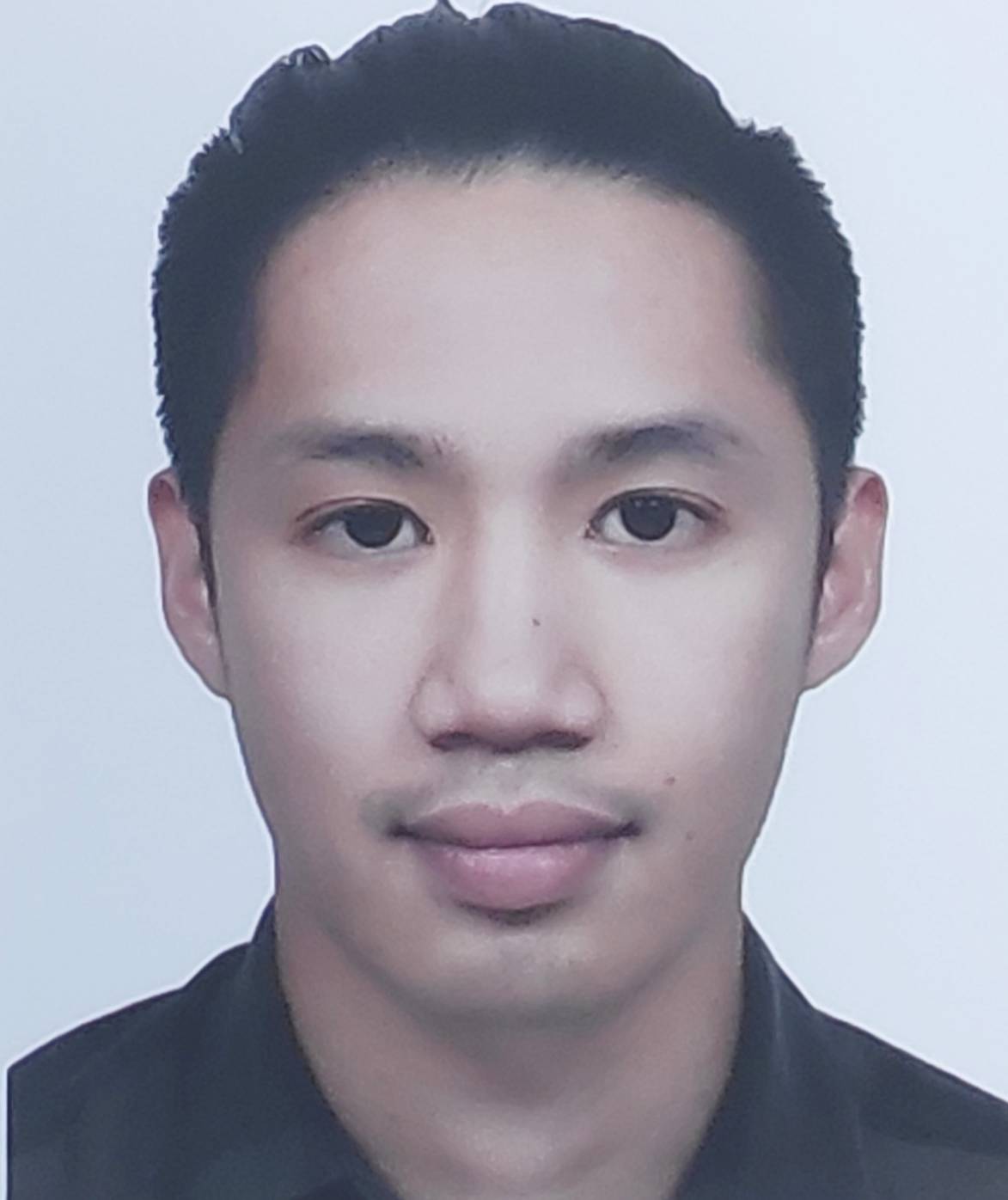}}]{Christian Alison M. P.} %(M'76--SM'81--F'87)
received the dual bachelor's degree in industrial engineering from Swiss German University, Indonesia, and the South Westphalia University of Applied Sciences, Soest, Germany, in 2020, and the M.Sc. degree in systems engineering and engineering management from the South Westphalia University of Applied Sciences, in 2022. Since 2022, he has been working as an XR Consultant with the company PACE Aerospace \& IT. His research interests include the future applications of augmented reality and virtual reality for process improvement and knowledge internalization.
\end{IEEEbiography}

\begin{IEEEbiography}[{\includegraphics[width=1in,height=1.25in,clip,keepaspectratio]{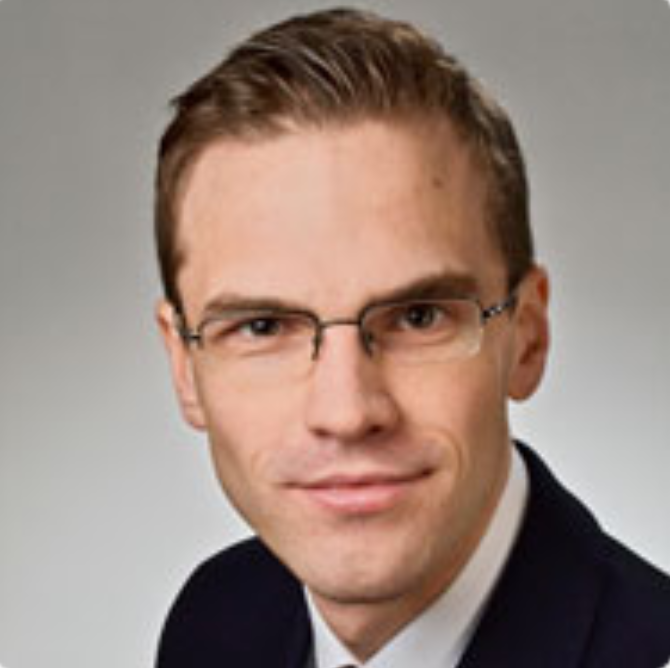}}]{Andreas Schwung} %(M'76--SM'81--F'87) 
received the Ph.D. degree in electrical engineering from the Technische Universit\"{a}t Darmstadt, Darmstadt, Germany in 2011.
From 2011 to 2015, he was an R\&D Engineeer with MAN Diesel \& Turbo SE, Oberhausen, Germany. Since 2015, he has been a Professor of automation technology at the South Westphalia University of Applied Sciences, Soest, Germany. 
His research interests include model-based control, networked automation systems, and intelligent data analytics with application in manufacturing, process industry, and electromobility.
\end{IEEEbiography}

% \EOD

\end{document}